\documentclass[runningheads]{llncs}

\usepackage{eccv}

\usepackage{eccvabbrv}
\usepackage[symbol]{footmisc}
\usepackage{graphicx}
\usepackage{booktabs}
\usepackage{multirow}
\usepackage{amsmath}
\usepackage{bm}
\usepackage{algorithm}
\usepackage{algpseudocode}
\usepackage{float}
\definecolor{bluenice}{HTML}{DAE8FC}
\DeclareMathOperator*{\argmin}{arg\,min}

\usepackage[accsupp]{axessibility}  
\usepackage{wrapfig,lipsum}

\usepackage{hyperref}

\usepackage{orcidlink}

\begin{document}

\title{Data-to-Model Distillation: Data-Efficient Learning Framework} 


\author{
Ahmad Sajedi\textsuperscript{1\thanks{Equal contribution \\ \textit{We include some relevant evaluation protocol material in \ref{sec:eval}}. }}\orcidlink{0009-0000-0618-519822} \and
Samir Khaki\textsuperscript{1*}\orcidlink{0009-0006-7278-8652}\and
Lucy Z. Liu\textsuperscript{2} \and Ehsan Amjadian \textsuperscript{2}\orcidlink{0000-0002-4400-6717}\and Yuri A. Lawryshyn\textsuperscript{1}\and Konstantinos N. Plataniotis \textsuperscript{1}\orcidlink{0000-0003-3647-5473}}
\authorrunning{Ahmad Sajedi\textsuperscript{*}, Samir Khaki\textsuperscript{*} et al.}

\institute{University of Toronto, Toronto, Canada \and
Royal Bank of Canada (RBC), Toronto, Canada\\
\url{https://datadistillation.github.io/D2M/} \\
\email{\{ahmad.sajedi,samir.khaki\}@mail.utoronto.ca}}

\maketitle

\begin{abstract}
Dataset distillation aims to distill the knowledge of a large-scale real dataset into small yet informative synthetic data such that a model trained on it performs as well as a model trained on the full dataset. Despite recent progress, existing dataset distillation methods often struggle with computational efficiency, scalability to complex high-resolution datasets, and generalizability to deep architectures. These approaches typically require retraining when the distillation ratio changes, as knowledge is embedded in raw pixels. In this paper, we propose a novel framework called Data-to-Model Distillation (D2M) to distill the real dataset’s knowledge into the learnable parameters of a pre-trained generative model by aligning rich representations extracted from real and generated images. The learned generative model can then produce informative training images for different distillation ratios and deep architectures. Extensive experiments on 15 datasets of varying resolutions show D2M's superior performance, re-distillation efficiency, and cross-architecture generalizability. Our method effectively scales up to high-resolution 128$\times$128 ImageNet-1K. Furthermore, we verify D2M's practical benefits for downstream applications in neural architecture search. 

\end{abstract}

\section{Introduction} \label{sec:intro}
\begin{figure} [t]
    \centering
    \includegraphics[width=\textwidth]{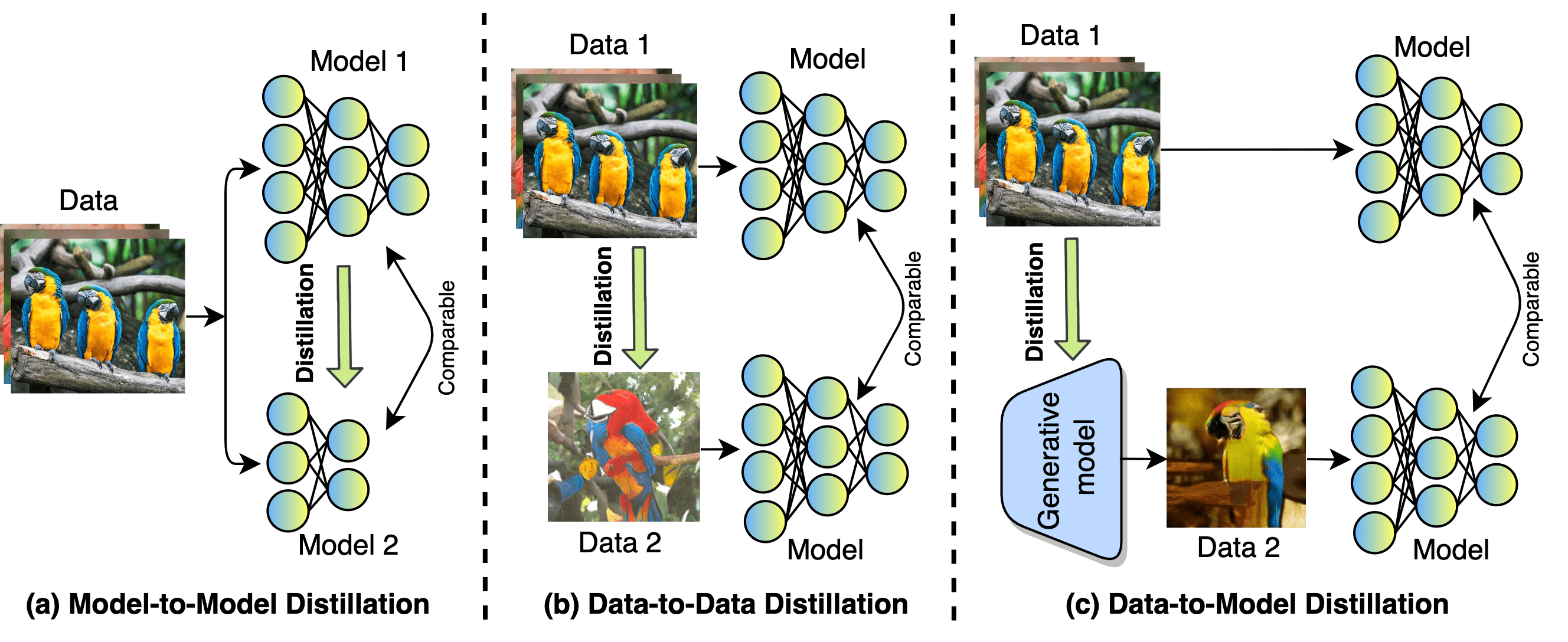}
    \caption{Different distillation frameworks for efficient learning.}
    \label{fig:distill}
\end{figure}

Deep learning has been successful in various domains, including computer vision and natural language processing \cite{he2016deep, tan2019efficientnet, brown2020language, kenton2019bert, sajedi2023end, sajedi2024probmcl, amer2021high, 9679989}; however, it often relies on deep neural networks (DNNs) and large-scale datasets. These requirements necessitate significant investments in training time, data storage, and electricity consumption, making training impractical for those with limited computational resources. Techniques such as pruning \cite{lipruning, molchanov2019pruning, khaki2024need, khaki2023cfdp}, quantization \cite{zhou2023dataset, wu2016quantized}, and model distillation \cite{hinton2015distilling, li2022knowledge} show promise in reducing these computational expenses while preserving performance. \textit{Model distillation} \cite{hinton2015distilling, romero2014fitnets, zagoruyko2016paying, furlanello2018born, park2019relational, sajedi2021efficiency, sajedi2022subclass, he2023yoco}, in particular, transfers the informative knowledge from a large teacher model to a smaller student one to facilitate model compression (\cref{fig:distill}\textcolor{black}{(a)}). Recently, \textit{dataset distillation} \cite{wang2018dataset, zhao2021datasetDC, cazenavette2022dataset, zhao2023dataset, sajedi2023datadam, zhoudataset, cazenavette2023generalizing, zhang2023accelerating} has emerged as a promising data-efficient learning approach that distills knowledge from a large training dataset into a small set of synthetic images. These images enable models to attain test performance comparable to those trained on the original dataset (\cref{fig:distill}\textcolor{black}{(b)}). Dataset distillation is widely applied in computer vision for applications like neural architecture search \cite{ho2016generative, such2020generative}, continual learning \cite{zhao2021datasetDSA, sajedi2023datadam, chen2023data, gu2024ssd, yang2024efficient}, federated learning \cite{xiong2023feddm, liu2022meta, liu2023slimmable}, and privacy-preserving \cite{dong2022privacy, chen2022private, loo2023understanding}.

Traditional data-centric methods, known as \textit{coreset selection}, reduce training costs by selecting a subset of the original dataset based on specific metrics \cite{tonevaempirical, zhou2023dataset, rebuffi2017icarl, belouadah2020scail, seneractive}. However, these approaches have a deficiency in representation capability and coverage of chosen samples, leading to suboptimal results for classification tasks \cite{tonevaempirical, rebuffi2017icarl, zhou2023dataset}. State-of-the-art dataset distillation algorithms overcome these limitations by synthesizing a small yet informative dataset that aligns characteristics between synthetic and real images. Drawing on seminal work \cite{wang2018dataset}, researchers have developed various methods, including gradient matching \cite{zhao2021datasetDC, zhao2021datasetDSA, lee2022dataset, du2024sequential}, distribution matching \cite{zhao2023dataset, wang2022cafe, zhao2023improved, sajedi2023datadam, khaki2024atom}, trajectory matching \cite{cazenavette2022dataset, du2023minimizing, cui2023scaling, guo2024towards}, and kernel-inducing points \cite{loo2022efficient, nguyen2021dataset, zhoudataset}. These works initialize a small number of learnable images and update their raw pixel values using different matching strategies during distillation.

Existing dataset distillation studies have shown remarkable performance but suffer from three major drawbacks. First, they require complete retraining of distillation algorithms when the distillation ratio (or the number of images per class) changes, which leads to a computationally demanding re-distillation process. Second, these methods generally struggle with high-resolution, large-scale datasets (\eg 128$\times$128 ImageNet-1K \cite{deng2009imagenet}) and tend to distill visually noisy images \cite{cazenavette2022dataset}. Lastly, the distilled dataset often performs poorly on architectures like ResNet \cite{he2016deep}, DenseNet \cite{huang2017densely} and ViT \cite{dosovitskiy2020image}. These issues partially stem from parameterizing the synthetic dataset in pixel space. Optimizing raw pixels can capture high-frequency, detailed information that is not essential for downstream tasks and is susceptible to over-fitting to the training architecture \cite{cazenavette2023generalizing}. Moreover, employing pixel-level optimization on high-resolution, large-scale data incurs significant computational and memory costs, rendering it unscalable for such datasets.

In this paper, we propose a novel framework, Data-to-Model distillation (\texttt{\texttt{D2M}}), that transfers information from the original dataset into a generative model rather than relying on raw pixel data (\cref{fig:distill}\textcolor{black}{(c)}). \texttt{D2M} effectively addresses the aforementioned issues by parameterizing the synthetic dataset within the parameter space of a generative model, such as Generative Adversarial Networks (GANs) \cite{goodfellow2014generative, brock2018large}. Following the approach of \cite{zhoudataset, wang2023dim, sajedi2023datadam, zhao2023improved}, our method employs a model pool to extract rich representations that are essential for learning. We design embedding matching and prediction matching modules to minimize differences in channel attention maps (at various layers) and output predictions between real and generated images, respectively. As depicted in \cref{fig:D2M}, these modules facilitate the distillation of different types of knowledge into the generative model using suitable matching losses, all geared towards improving classification performance. Following the distillation stage, our generative model can create training samples for classification tasks with \textit{any} number of images per class from random noises. Unlike other dataset distillation methods, \texttt{D2M} circumvents the need for retraining the distillation when changing the distillation ratio, which offers clear advantages in re-distillation efficiency. As \texttt{D2M} stores information in models rather than pixels, it stands as one of the pioneer distillation methods scalable to high-resolution (256$\times$256) and large-scale datasets. This makes it a promising tool for efficient synthetic data generation, requiring significantly less memory storage. The contributions of our study are:

\textbf{[C1]}: We propose a novel framework to distill the knowledge of large-scale datasets into a \textit{parameter space of a generative model} that can produce informative images for classification tasks. The method distills various representations from the real data to provide diverse supervision.

\textbf{[C2]}: We conduct extensive experiments on 15 datasets, differing in resolution, label complexity, and application domains. We achieve state-of-the-art results across all settings and enable the distillation of the ImageNet-1K at a resolution of 128$\times$128 with a fixed storage complexity across different settings.

\textbf{[C3]}: We show that data-to-model distillation outperforms prior dataset distillation algorithms in re-distillation efficiency and cross-architecture generality. The high-quality images generated by \texttt{D2M} also notably improve downstream applications in neural architecture search.

\section{Related Work} \label{sec:related}
\textbf{Dataset Distillation.} Wang \etal \cite{wang2018dataset} introduced dataset distillation to synthesize small-scale data from a large real dataset using meta-learning, minimizing training loss differences between the synthetic and original data. Meta-learning involves bi-level optimization and incurs significant computational costs. To mitigate this, researchers employed kernel methods to facilitate the inner optimization loop, as seen in KIP \cite{nguyen2021dataset, nguyen2021dataset2}, RFAD \cite{loo2022efficient}, and FRePo \cite{zhoudataset}. Other studies adopted surrogate objectives to tackle unrolled optimization problems in meta-learning. DC \cite{zhao2021datasetDC}, DSA \cite{zhao2021datasetDSA}, and IDC \cite{kim2022dataset} align gradients between synthetic and real datasets for distillation. Meanwhile, DM \cite{zhao2023dataset}, CAFE \cite{wang2022cafe}, DAM \cite{sajedi2023datadam}, and ATOM \cite{khaki2024atom} use distribution-matching to alleviate bias caused by original samples with large gradients. Further, MTT \cite{cazenavette2022dataset}, FTD \cite{du2023minimizing}, and TESLA \cite{cui2023scaling} match model parameter trajectories to improve performance. However, these methods distill information into pixels, leading to a linear growth in computational costs with class numbers and resolutions. This limits scalability with larger datasets and poses challenges for cross-architecture generalization and re-distillation efficiency \cite{zhao2021datasetDC, sajedi2023datadam, cazenavette2022dataset, wang2022cafe, liu2022dataset}. In contrast, \texttt{D2M} distills the knowledge into the generative model to address the limitations posed by pixel-wise dataset distillation.

\textbf{Generative Models.} 
This work is closely related to generative models, specifically generative adversarial networks (GANs) \cite{goodfellow2014generative, zhu2017unpaired, ledig2017photo, wang2018high}. Goodfellow et al. \cite{goodfellow2014generative} introduced vanilla GANs to generate realistic images that deceive human observers. A variety of GAN models, such as CycleGAN \cite{zhu2017unpaired}, InfoGAN \cite{chen2016infogan}, and BigGAN \cite{brock2018large}, have since emerged for tasks like image manipulation \cite{wang2018high, zhu2017unpaired, lee2020maskgan}, super-resolution \cite{ledig2017photo, he2022gcfsr}, and object detection \cite{li2017perceptual, liu2019generative}. However, these GAN-generated images often prioritize visual realism over informativeness, which may not be ideal for data-efficient classification tasks \cite{zhao2022synthesizing, cazenavette2023generalizing, wang2023dim, wang2022cafe}. Some methods have been proposed to address this by generating samples that can be used to train deep neural networks more efficiently. GLaD \cite{cazenavette2023generalizing} distills numerous images into a few intermediate feature vectors in the GAN’s latent space to help cross-architecture generalization. Wang et al. \cite{wang2023dim} introduced a two-stage algorithm for optimizing generative models to create training samples for 10-class classification tasks. Nevertheless, these methods face challenges such as high training costs \cite{wang2023dim}, the necessity of retraining for different distillation ratios \cite{cazenavette2023generalizing}, and scalability constraints for complex, high-resolution datasets \cite{wang2023dim}. In contrast, our approach distills discriminative knowledge from the dataset into the \textit{parameter space} of generative models, reducing re-distillation costs and achieving superior performance on standard benchmarks for both low- and high-resolution datasets.

\section{Methodology} \label{sec:method}
\begin{figure} [t]
    \centering
    \includegraphics[width=1\textwidth]{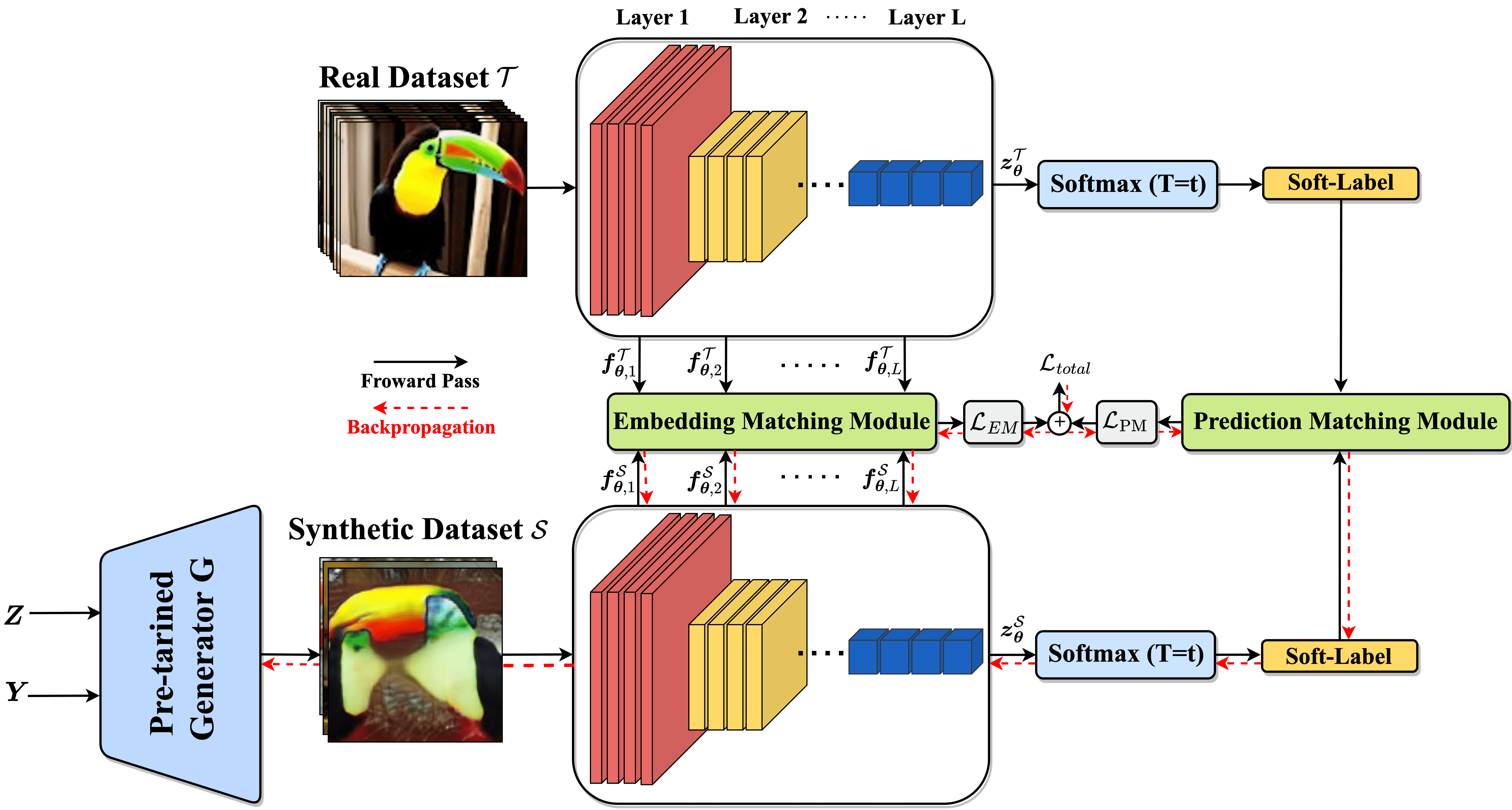}
    \caption{An overview of the proposed \texttt{D2M} framework. \texttt{D2M} distills the knowledge of large-scale datasets into the parameter space of a pre-trained generator through embedding matching and prediction matching modules. The learned generator can then produce small yet informative training images for the downstream classification tasks. $Z$ and $Y$ represent the random noises and labels, respectively.}
    \label{fig:D2M}
\end{figure}

In this section, we introduce a novel framework called \texttt{D2M}, which distills knowledge from a large-scale training dataset $\mathcal{T} = \{(\bm{x}_{i}, y_{i})\}_{i=1}^{|\mathcal{T}|}$, with $|\mathcal{T}|$ image-label pairs, into a generative model $G$. In the distillation phase, we extract comprehensive knowledge from the real dataset $\mathcal{T}$, including embedded representations and logit prediction, and then distill them into the parameters of a pre-trained generative model. This procedure refines the generative model to synthesize a small yet informative dataset $\mathcal{S} = \{(\bm{s}_{j}, y_{j})\}_{j=1}^{|\mathcal{S}|}$ that has comparable training power to the real dataset. Following distillation, the learned generative model produces training images from random noises, enabling flexibility in the distillation ratios. These samples are subsequently deployed in classification tasks to evaluate \texttt{D2M}'s performance. The overall \texttt{D2M} pipeline is depicted in \cref{fig:D2M}.

\subsection{Data-to-Model Distillation}
The proposed \texttt{D2M} differs from data distillation works that directly distill knowledge into raw pixels. In \texttt{D2M}, we distill the information into the learnable parameters of a generative model, offering a new path for efficient learning.

A randomly initialized generative model yields noisy outputs, making it challenging to establish meaningful correspondences with real images during the distillation process. Therefore, we employ a pre-trained Generative Adversarial Network (GAN) with a generator $G$ and a discriminator $D$, trained using the following standard loss function:
\begin{flalign} 
  {\min_{G} \max_{D}  \mathop{\mathbb{E}}_{\bm{x}\sim P_{{\mathcal{T}}}}\big[\log D(\bm{x}|y)\big] +  \mathop{\mathbb{E}}_{\bm{z}\sim P_{{Z}}}\big[\log (1-D(G(\bm{z}|y)))\big],} 
    \end{flalign}
where $P_{\mathcal{T}}$ and $P_{Z}$ denote the distributions of real training images and latent vectors, respectively. As highlighted by \cite{zhao2022synthesizing}, conventional GANs typically generate images that deviate from the original data distribution and are less informative than real samples for training deep networks. To mitigate this limitation, we introduce a distillation stage to refine the generative model. This refinement enables the model to synthesize images that are more discriminative and better suited to downstream classification tasks.

\textbf{Distillation Stage.} 
We randomly select $B$ pairs consisting of random noises $Z$ and labels $Y$ in each batch. These pairs are then fed into the pre-trained generative model to produce synthetic images $\mathcal{S}$. Meanwhile, we also randomly pick $B$ real images from the original dataset $\mathcal{T}$ with corresponding labels $Y$. During each iteration, both real and synthetic batches undergo a forward pass using a randomly chosen neural network $\phi_{\boldsymbol{\theta}}(\cdot)$ with different initializations from a model pool. This network is responsible for extracting features and predicting classification logits for both real and synthetic images. The network generates feature maps and predicted logits for each dataset, denoted as $\phi_{\boldsymbol{\theta}}(\mathcal{T}) = [\bm{f}^{\mathcal{T}}_{\boldsymbol{\theta},1}, \cdots, \bm{f}^{\mathcal{T}}_{\boldsymbol{\theta}, L}, \bm{z}_{\boldsymbol{\theta}}^{\mathcal{T}}]$ for the real dataset and $\phi_{\boldsymbol{\theta}}(\mathcal{S}) = [\bm{f}^{\mathcal{S}}_{\boldsymbol{\theta},1}, \cdots, \bm{f}^{\mathcal{S}}_{\boldsymbol{\theta}, L}, \bm{z}_{\boldsymbol{\theta}}^{\mathcal{S}}]$ for the synthetic one. The feature map $\bm{f}^{\mathcal{T}}_{\boldsymbol{\theta},l}$ and logit $\bm{z}_{\boldsymbol{\theta}}^{\mathcal{T}}$ are multi-dimensional arrays obtained from the real dataset in the $l^\text{th}$ and pre-softmax layers of the network, respectively. Similarly, feature $\bm{f}^{\mathcal{S}}_{\boldsymbol{\theta},l}$ and logits $\bm{z}_{\boldsymbol{\theta}}^{\mathcal{S}}$ are derived from the synthetic set. We then adopt embedding matching and prediction matching modules to capture the representations of the real training set across various layers.

The embedding matching module aligns the channel-wise attention maps between the real and synthetic sets across different feature extraction layers using the loss function $\mathcal{L}_\text{EM}$, formulated as:
\begin{flalign} \label{eq:fm}
    \displaystyle \mathop{\mathbb{E}}_{\boldsymbol{\theta}\sim P_{\boldsymbol{\theta}}}\Bigg[\sum_{l=1}^{L}\bigg\lVert \displaystyle {\mathbb{E}}_{\mathcal{T}}\Big[\tilde{\bm{f}}^{\mathcal{T}}_{\boldsymbol{\theta},l}\Big] - \displaystyle \mathbb{E}_{{\mathcal{S}}}\Big[\tilde{\bm{f}}^{\mathcal{S}}_{\boldsymbol{\theta},l}\Big]\bigg\rVert^{2}\Bigg],
\end{flalign}
where $P_{\boldsymbol{\theta}}$ is the distribution of neural network parameters and $\tilde{\bm{f}}^{\mathcal{T}}_{\boldsymbol{\theta},l}$ is defined as:
\begin{equation}
     \tilde{\bm{f}}^{\mathcal{T}}_{\boldsymbol{\theta},l} := 
\begin{cases}
  \bm{a}^{\mathcal{T}}_{\boldsymbol{\theta},l} & \text{if $l = 1, \cdots, L-1$} \\
 \bm{f}^{\mathcal{T}}_{\boldsymbol{\theta},L} & \text{if $l=L$},
\end{cases}
\end{equation}
in which $\bm{a}^{\mathcal{T}}_{\boldsymbol{\theta},l}$ represents the vectorized channel attention maps in the $l^{\text{th}}$ layer for the real dataset. In the same way, the $\tilde{\bm{f}}^{\mathcal{S}}_{\boldsymbol{\theta},l}$ can be defined for the synthetic set. In particular, we characterize each training image through the use of channel-wise attention maps created by different layers. As noted in \cite{chen2017sca, khaki2024atom}, these attention maps highlight the most discriminative regions of the input image, revealing the network's focus across various layers (early and intermediate) for obtaining information at low- and mid-level representations. The last layer of feature extraction in the neural network contains the highest-level abstract information \cite{ma2015hierarchical}. This embedded has been shown to effectively capture semantic information from the input data \cite{zhao2023dataset, saito2018maximum}. Thus, we leverage the mean square error loss to match the vectorized versions of the final feature maps with real and synthetic data. Our embedding matching uses the most \textit{discriminative} regions of feature maps using the concept of attention, differing significantly from pure feature matching approaches like CAFE \cite{wang2022cafe}. Notably, in cases where ground-truth data distributions are unavailable, we empirically estimate the expectation term in \cref{eq:fm}.

Although $\mathcal{L}_{\text{EM}}$ effectively approximates large-scale real data distribution, its matching loss primarily minimizes the mean feature distance within each batch without explicitly constraining the diversity of synthetic images. Therefore, we use a complementary loss as a regularizer to provide more specific supervision. This regularization promotes similarity in the output probability predictions between the two datasets, directly influencing the results of the downstream classification task. Drawing inspiration from knowledge distillation works \cite{yuan2020revisiting, hinton2015distilling, jin2023multi}, we employ logit-based matching to minimize the differences in the \textit{softened} output predictions between real and generated synthetic images. The soft-label predictions introduce additional information to the generative model that can be helpful for classification tasks \cite{yin2023sre2l}. The prediction matching loss $\mathcal{L}_{\text{PM}}$ can then be written as follows:
\begin{flalign} \label{eq:pm}
    \displaystyle \mathop{\mathbb{E}}_{\boldsymbol{\theta}\sim P_{\boldsymbol{\theta}}}\Bigg[\sum_{i=1}^{B} KL\bigg(\sigma\Big(\frac{\bm{z}_{\boldsymbol{\theta}}^{\bm{x}_{i}}}{T}
\Big), \sigma\Big(\frac{\bm{z}_{\boldsymbol{\theta}}^{\bm{s}_{i}}}{T}
\Big)\bigg)\Bigg],
\end{flalign}
where KL stands for Kullback-Leibler divergence, $\sigma(\cdot)$ denotes the softmax function, and $\bm{z}_{\boldsymbol{\theta}}^{\bm{x}_{i}}$ and $\bm{z}_{\boldsymbol{\theta}}^{\bm{s}_{i}}$ are the prediction logits for real and synthetic images sharing the same label $y_{i}$, respectively. The temperature hyperparameter $T$ is used to generate soft-label predictions while regulating the entropy of the output distribution \cite{hinton2015distilling, yuan2020revisiting}. Additional details on the impact of $T$ are discussed in \cref{subsec:ablation}. Finally, the training loss for the \texttt{D2M} framework is formulated as the augmented Lagrangian of the two mentioned losses. We learn the \textit{parameters of a generative model} by solving the following optimization problem using SGD:
\begin{flalign} \label{eq:loss}
{G}^{*} = \argmin_{G}\:\big(\mathcal{L}_{\text{EM}} + \lambda \mathcal{L}_{\text{PM}}\big),
\end{flalign}
where $\lambda$ serves as a Lagrangian multiplier to balance the gradients of $\mathcal{L}_{\text{EM}}$ and $\mathcal{L}_{\text{PM}}$. The impact of $\lambda$ is examined in \cref{subsec:ablation} and a learning algorithm is summarized in Alg. \ref{alg:d2m}.

\begin{algorithm}[t] 
\textbf{Input:} \text{Real training dataset $\mathcal{T}=\{(\bm{x}_{i}, y_{i})\}_{i=1}^{|\mathcal{T}|}$}\\
\textbf{Required:} Pre-trained generative model $G$, Latent vector set $Z$, Deep neural network $\phi_{\bm{\theta}}$ from a model pool with parameter distribution $P_{\boldsymbol{\theta}}$, Learning rate $\eta_{G}$, Task balance parameter $\lambda$, Number of training iterations $I$.
\begin{algorithmic}[1]
\For{$i = 1, 2, \cdots, I$}  \label{line:1}
	\State Sample $\bm{\theta}$ from $P_{\bm{\theta}}$ 
        \State Sample $B$ pairs from the real dataset and the generated images of $G(Z)$
        \State Compute $\mathcal{L}_{\text{EM}}$ and $\mathcal{L}_{\text{PM}}$ using Equations \ref{eq:fm} and \ref{eq:pm}
        \State Calculate \colorbox{bluenice}{$\mathcal{L} = \mathcal{L}_{\text{EM}} + \lambda \mathcal{L}_{\text{PM}}$}
        \State Update the generator using {$G \leftarrow G - \eta_{G}\nabla_{G}\mathcal{L}$}
\EndFor
\end{algorithmic}
\textbf{Output:} \text{Synthetic dataset $\mathcal{S}= G(Z) = \{(\bm{s}_{i}, y_{i})\}_{i=1}^{|\mathcal{S}|}$}
\caption{Data-to-Model Distillation (\texttt{D2M})}
\label{alg:d2m}
\end{algorithm}

\textbf{Comparing with Other Generative Prior Methods.} Recent generative-based dataset distillation methods typically distill large-scale datasets into the latent space of a specific GAN model \cite{cazenavette2023generalizing, zhao2022synthesizing}. For example, GLaD \cite{cazenavette2023generalizing} utilizes various intermediate feature spaces of StyleGAN-XL \cite{sauer2022stylegan} to characterize synthetic datasets. IT-GAN \cite{cazenavette2023generalizing} creates synthetic datasets in a BigGAN's latent space \cite{brock2018large} by initially inverting the full training set and then fine-tuning the latent representations based on the distillation objective. Both GLaD and IT-GAN necessitate complete retraining when distillation ratios change due to their information containers, which require updates when changing the IPC. In contrast, \texttt{D2M} proposes a novel approach by parameterizing the synthetic dataset within the learnable \textit{parameter space} of {any} generative model using our carefully designed framework. Unlike GLaD and IT-GAN, whose storage complexity grows linearly with increasing IPCs, our framework maintains a constant storage complexity across various distillation ratios. To demonstrate the efficacy of our framework, we empirically compare \texttt{D2M} with GLaD and IT-GAN, consistently outperforming them in all cases, as detailed in \cref{subsec:comp_sota} and supplementary materials.

\section{Experiments}
\label{sec:experiments}

\subsection{Experimental Setup} \label{subsec:expset}

\textbf{Datasets.}
In line with prior studies, we evaluate the proposed framework across diverse image classification datasets with varying resolutions and label complexity. Our method is applied to CIFAR10/100 \cite{krizhevsky2009learning} for low-resolution data. The TinyImageNet \cite{le2015tiny} and ImageNet-1K \cite{deng2009imagenet} datasets are resized to 64$\times$64 for medium-resolution ones. For high-resolution data (\ie 128$\times$128), we employ ImageNet-1K and its subsets—ImageNette \cite{howard2019imagenette}, ImageWoof \cite{howard2019imagenette}, ImageSquawk \cite{cazenavette2022dataset}, ImageFruit \cite{cazenavette2022dataset}, and ImageMeow \cite{cazenavette2022dataset}—each emphasizing specific categories such as assorted objects, dog breeds, birds, fruits, and cat breeds. We also use the recently introduced 10-class ImageNet-1K subsets (ImageNet-[A, B, C, D, E]) \cite{cazenavette2023generalizing}, which were determined through the evaluation performance of a pre-trained ResNet-50 on ImageNet-1K. Finally, we extend \texttt{D2M} to the medical imaging domain, focusing on high-resolution 256$\times$256 images in DermaMNIST \cite{yang2023medmnist}. This dataset features dermatoscopic images of typical pigmented skin lesions. More details on the datasets can be found in the supplementary materials.

\textbf{Network Architectures.}
Following recent dataset distillation works \cite{zhoudataset, sajedi2023datadam, wang2023dim}, we leverage a model pool to mitigate overfitting and offer diverse views for matching tasks. The model pool includes Depth-$n$ ConvNets \cite{gidaris2018dynamic} and ResNet-18/32 \cite{he2016deep} with different initialization random seeds \cite{zhoudataset, sajedi2023datadam}. To ensure diversity, we randomly select one model from the pool at each iteration. In a Depth-$n$ ConvNet, $n$ blocks precede a fully connected layer, with each block containing a 3$\times$3 convolutional layer (128 filters), instance normalization \cite{ulyanov2016instance}, ReLU activation, and 2$\times$2 average pooling (stride of 2). The number of blocks depends on dataset resolution: $n=$ 3, 4, 5, and 6 for 32$\times$32, 64$\times$64, 128$\times$128, and 256$\times$256, respectively. We initialize networks using normal initialization \cite{he2015delving} and use BigGAN \cite{brock2018large} as a default generative model.

\textbf{Evaluation Protocol.} Our method uniquely allows the dynamic generation of images for training. To ensure fairness with conventional studies like \cite{zhao2023dataset, zhao2021datasetDC, zhao2021datasetDSA, sajedi2023datadam}, we first measure the end-to-end latency for training each model on the given dataset at a particular IPC. When evaluating our method, we generate images on-the-fly during downstream model training, ensuring that the total time for both generating and training matches the latency constraints set by previous works. This approach explores a tradeoff between model training and image generation on the fly during evaluation. Details on the generation and training time costs are available with our source code \footnote{\url{https://github.com/DataDistillation/D2M}}. Apart from dynamic generation, we use the same ConvNet networks and optimizers as previous works for valid comparison. We also quantify the computational costs during distillation in GPU hours and the number of learnable parameters. For fairness and reproducibility, we employ publicly available datasets of baselines and train models using our experimental settings. For datasets without specific experiments, we implemented them using the released author codes. Additional implementation details are provided in the supplementary materials.

\begin{table*}[t]
\centering
\caption{The performance (\%) comparison to priors on CIFAR and TinyImageNet. ConvNet is used for evaluation. For more details, refer to the supplementary materials.}

\scriptsize
\renewcommand{\arraystretch}{1.1}
\resizebox{1\linewidth}{!}{
\begin{tabular}{c|ccc|ccc|ccc}
\toprule
Dataset   & \multicolumn{3}{c|}{CIFAR-10} & \multicolumn{3}{c|}{CIFAR-100} & \multicolumn{3}{c}{Tiny ImageNet}       \\
IPC       & 1     & 10  & 50 

& 1      & 10    & 50    

& 1        & 10      & 50      \\

Ratio \%    & 0.02  & 0.2 & 1  
& 0.2    & 2     & 10    
& 0.2      & 2       & 10      \\
\midrule
Random       
& $14.4_{\pm2.0}$ & $26.0_{\pm1.2}$ & $43.4_{\pm1.0}$ 
& $4.2_{\pm0.3}$ & $14.6_{\pm0.5}$ & $30.0_{\pm0.4}$ 
& $1.4_{\pm0.1}$ & $5.0_{\pm0.2}$ & $15.0_{\pm0.4}$ \\

Herding       
& $21.5_{\pm1.2}$ & $31.6_{\pm0.7}$ & $40.4_{\pm0.6}$ 
& $8.4_{\pm0.3}$ & $17.3_{\pm0.3}$ & $33.7_{\pm0.5}$ 
& $2.8_{\pm0.2}$ & $6.3_{\pm0.2}$ & $16.7_{\pm0.3}$ \\

Forgetting      
& $13.5_{\pm1.2}$ & $23.3_{\pm1.0}$ & $23.3_{\pm1.1}$ 
& $4.5_{\pm0.2}$ & $15.1_{\pm0.3}$ & $30.5_{\pm0.3}$ 
& $1.6_{\pm0.1}$ & $5.1_{\pm0.2}$ & $15.0_{\pm0.3}$ \\
\midrule

DC \cite{zhao2021datasetDC}         
& $28.3_{\pm0.5}$ & $44.9_{\pm0.5}$ & $53.9_{\pm0.5}$ 
& $12.8_{\pm0.3}$ & $25.2_{\pm0.3}$ & $30.6_{\pm0.6}$ 
& $5.3_{\pm0.1}$ & $12.9_{\pm0.1}$ & $12.7_{\pm0.4}$ \\

DSA \cite{zhao2021datasetDSA}          
& $28.8_{\pm0.7}$ & $52.1_{\pm0.5}$ & $60.6_{\pm0.5}$ 
& $13.9_{\pm0.3}$ & $32.3_{\pm0.3}$ & $42.8_{\pm0.4}$ 
& $6.6_{\pm0.2}$ & $16.3_{\pm0.2}$ & $25.3_{\pm0.2}$ \\

DM \cite{zhao2023dataset}          
& $26.0_{\pm0.8}$ & $48.9_{\pm0.6}$ & $63.0_{\pm0.4}$ 
& $11.4_{\pm0.3}$ & $29.7_{\pm0.3}$ & $43.6_{\pm0.4}$ 
& $3.9_{\pm0.2}$ & $12.9_{\pm0.4}$ & $24.1_{\pm0.3}$ \\

CAFE \cite{wang2022cafe}       
& $30.3_{\pm1.1}$ & $46.3_{\pm0.6}$ & $55.5_{\pm0.6}$ 
& $12.9_{\pm0.3}$ & $27.8_{\pm0.3}$ & $37.9_{\pm0.3}$ 
& - & - & - \\

CAFE+DSA \cite{wang2022cafe}        
& $31.6_{\pm0.8}$ & $50.9_{\pm0.5}$ & $62.3_{\pm0.4}$ 
& $14.0_{\pm0.3}$ & $31.5_{\pm0.2}$ & $42.9_{\pm0.2}$  
& - & - & - \\

DAM \cite{sajedi2023datadam}     
& $32.0_{\pm1.2}$ & $54.2_{\pm0.8}$ & $67.0_{\pm0.4}$ 
& $14.5_{\pm0.5}$ & $34.8_{\pm0.5}$ & $49.4_{\pm0.3}$ 
& $8.3_{\pm0.4}$ & $18.7_{\pm0.3}$ & $28.7_{\pm0.3}$ \\

ATOM \cite{khaki2024atom}        
& $34.8_{\pm1.0}$ & $57.9_{\pm0.7}$ & $68.8_{\pm0.5}$ 
& $18.1_{\pm0.4}$ & $35.7_{\pm0.4}$ & $50.2_{\pm0.3}$ 
& $9.1_{\pm0.2}$ & $19.5_{\pm0.4}$ & $29.1_{\pm0.3}$ \\

IDM \cite{zhao2023improved}      
& $45.6_{\pm0.7}$ & $58.6_{\pm0.1}$ & $67.5_{\pm0.1}$ 
& $20.1_{\pm0.3}$ & $45.1_{\pm0.1}$ & $50.0_{\pm0.2}$ 
& $10.1_{\pm0.2}$ & $21.9_{\pm0.2}$ & $27.7_{\pm0.3}$ \\

KIP \cite{nguyen2021dataset}         
& $49.9_{\pm0.2}$ & $62.7_{\pm0.3}$ & $68.6_{\pm0.2}$ 
& $15.7_{\pm0.2}$ & $28.3_{\pm0.1}$ & - 
& - & - & - \\

FRePo \cite{zhoudataset}          
& $46.8_{\pm0.7}$ & $65.5_{\pm0.4}$ & $71.7_{\pm0.2}$ 
& $28.7_{\pm0.1}$ & $42.5_{\pm0.2}$ & $44.3_{\pm0.2}$ 
& $15.4_{\pm0.3}$ & $25.4_{\pm0.2}$ & - \\

MTT \cite{cazenavette2022dataset}           
& $46.3_{\pm0.8}$ & $65.3_{\pm0.7}$ & $71.6_{\pm0.2}$ 
& $24.3_{\pm0.3}$ & $40.1_{\pm0.4}$ & $47.7_{\pm0.2}$ 
& $8.8_{\pm0.3}$ & $23.2_{\pm0.2}$ & $28.0_{\pm0.3}$ \\

TESLA \cite{cui2023scaling}         
& $48.5_{\pm0.8}$ & $66.4_{\pm0.8}$ & $72.6_{\pm0.7}$ 
& $24.8_{\pm0.4}$ & $41.7_{\pm0.3}$ & $47.9_{\pm0.3}$ 
& - & - & - \\

\textbf{D2M} (Ours)        
& $\mathbf{50.2_{\pm0.3}}$ & $\mathbf{67.8_{\pm0.5}}$ & $\mathbf{74.4_{\pm0.3}}$
& $\mathbf{29.8_{\pm0.4}}$ & $\mathbf{46.6_{\pm0.3}}$ & $\mathbf{51.2_{\pm0.2}}$
& $\mathbf{16.7_{\pm0.2}}$ & $\mathbf{26.1_{\pm0.4}}$ & $\mathbf{30.1_{\pm0.2}}$\\ \midrule

Full Dataset & \multicolumn{3}{c|}{$84.8_{\pm0.1}$} & \multicolumn{3}{c|}{$56.2_{\pm0.3}$}  & \multicolumn{3}{c}{$37.6_{\pm0.4}$} \\ \bottomrule
\end{tabular}
}
\label{tab:low-res}
\end{table*}

\textbf{Implementation Details.} We deploy the pre-trained BigGAN with the default hyperparameters and learning strategy outlined in \cite{brock2018large}. In the distillation, we refine BigGAN using the SGD optimizer with a batch size of $B =$ 128 over $K =$ 60 epochs while setting the task balance $\lambda$ and the temperature $T$ to 100 and 4, respectively. We also apply the same set of differentiable augmentations \cite{zhao2021datasetDSA} in all experiments, during distillation and evaluation. Experiments are conducted on two RTX A-6000 GPUs with 50 GB of memory. Additional hyperparameter details can be found in the supplementary materials.

\subsection{Comparison to State-of-the-art Methods} \label{subsec:comp_sota}

\textbf{CIFAR and Tiny ImageNet.} We compare the proposed \texttt{D2M} with 3 coreset selection and 12 dataset distillation approaches on the CIFAR-10/100 \cite{krizhevsky2009learning} and Tiny ImageNet \cite{le2015tiny} datasets, as detailed in \cref{tab:low-res}. \texttt{D2M} consistently outperforms all baselines across different settings for these low- and medium-resolution datasets. Specifically, it achieves up to 88\% of CIFAR-10's upper bound performance with only 1\% of its training data and up to 70\% of the upper bound with 2\% of Tiny ImageNet's training data. These results highlight the \texttt{D2M} framework's robustness for data-efficient learning and its ability to distill informative knowledge for improved classification task performance.
\begin{table}[t]
\renewcommand{\arraystretch}{1.0}
\footnotesize
\centering
\caption{The performance comparison (\%) to state-of-the-art methods on ImageNet-1K \cite{deng2009imagenet} and its high-resolution subsets \cite{cazenavette2022dataset}.}
 \setlength{\tabcolsep}{1pt}
 \setlength{\abovecaptionskip}{0.8cm}
\resizebox{1\linewidth}{!}{
\begin{tabular}{cc|cccccc|c}
\toprule
         Dataset & {IPC} & Random & DAM \cite{sajedi2023datadam} & GLaD \cite{cazenavette2023generalizing}  & TESLA \cite{cui2023scaling}  & FRePo \cite{zhoudataset} & \textbf{D2M} (Ours) & {Full Dataset} \\ 
 \midrule
 
\multirow{4}{*}{} 
& 1  &  $0.5_{\pm0.1}$ & $2.0_{\pm0.1}$  & - & $7.7_{\pm0.2}$ &  $7.5_{\pm0.3}$ & $\mathbf{8.3_{\pm0.2}}$ & \multirow{4}{*}{$33.8_{\pm0.3}$}\\ 
ImageNet-1K & 2  &  $0.9_{\pm0.1}$  & $2.2_{\pm0.1}$ & - & $10.5_{\pm0.3}$ & $9.7_{\pm0.2}$ & $\mathbf{11.1_{\pm0.2}}$\\
 (64$\times$64)& 10  & $3.1_{\pm0.2}$  & \multicolumn{1}{c} {$6.3_{\pm0.0}$} & - & $17.8_{\pm1.3}$ & - & $\mathbf{18.2_{\pm0.9}}$\\
& 50  & $7.6_{\pm1.2}$ &\multicolumn{1}{c}{$15.5_{\pm0.2}$}& - & $27.9_{\pm1.2}$ & - & $\mathbf{28.3_{\pm1.1}}$\\

\midrule

\multirow{2}{*}{ImageNette} 
& 1 &  $23.5_{\pm4.8}$  & $34.7_{\pm0.9}$ & $38.7_{\pm1.6}$ & - & $48.1_{\pm0.7}$ & $\mathbf{49.9_{\pm0.2}}$ &\multirow{2}{*}{$87.4_{\pm1.0}$}\\
& 10 &   $47.7_{\pm2.4}$  & ${59.4_{\pm0.4}}$& - & - & $66.5_{\pm0.8}$ & $\mathbf{67.7_{\pm0.3}}$\\
\midrule
    
\multirow{2}{*}{ImageWoof}  
& 1 &  $14.2_{\pm0.9}$  & ${24.2_{\pm0.5}}$ & $23.4_{\pm1.1}$ & -  & $29.7_{\pm0.6}$ & $\mathbf{32.3_{\pm0.4}}$ & \multirow{2}{*}{$67.0_{\pm1.3}$}\\
& 10 &  $27.0_{\pm1.9}$ & ${34.4_{\pm0.4}}$ & - & - & $42.2_{\pm0.9}$ & ${\mathbf{44.9_{\pm0.3}}}$\\
\midrule

\multirow{2}{*}{ImageSquawk}  
& 1 &  $21.8_{\pm0.5}$  & ${36.4_{\pm0.8}}$ & $35.8_{\pm1.4}$ & - & - & $\mathbf{40.3_{\pm0.5}}$ &\multirow{2}{*}{$87.5_{\pm0.3}$}\\
& 10 & $40.2_{\pm0.4}$  & $55.4_{\pm0.9}$ & - & - & - & $\mathbf{59.8_{\pm0.2}}$\\     
\bottomrule
\end{tabular}}
\label{tab:imagenet}
\end{table}

\begin{wrapfigure}[10]{R}{0.5\linewidth}
   \vspace{-2em}
    \centering
    \includegraphics[width=0.48\textwidth]{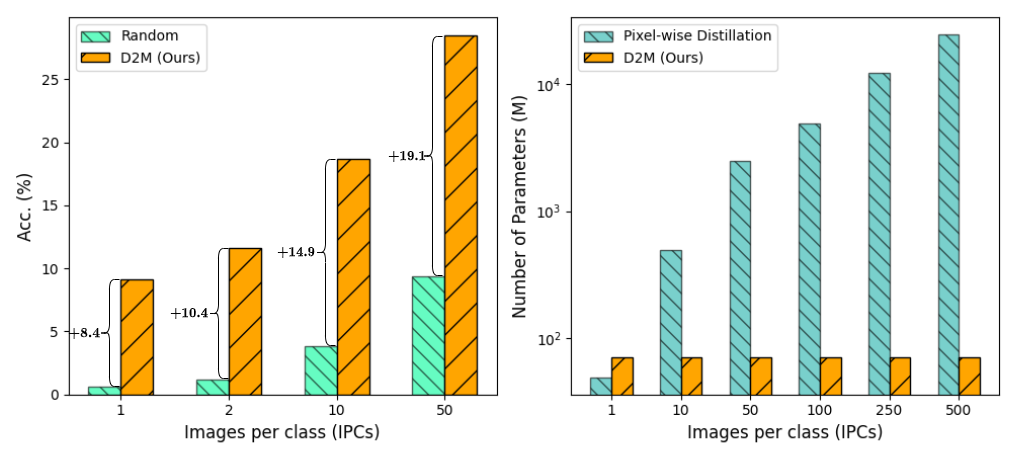}
        \caption{Performance comparison and count of parameters on 128$\times$128 ImageNet-1K.}
    \label{fig:costs}
\end{wrapfigure}

\textbf{ImageNet-1K and Its Subsets.} ImageNet-1K \cite{deng2009imagenet} and its subsets are more challenging than the CIFAR-10/100 and Tiny ImageNet datasets due to their complex label spaces and higher resolutions. Consequently, memory and time constraints prevented previous data distillation works from scaling up to ImageNet-1K at high resolution \cite{cazenavette2022dataset, zhoudataset, du2023minimizing, zhao2021datasetDSA, sajedi2023datadam, zhao2022synthesizing}. However, \texttt{D2M} overcomes this by distilling information from the original dataset into the generative model's parameters rather than relying on complex raw pixels. This enables \texttt{D2M}, as one of the pioneering works to generate synthetic 128$\times$128 ImageNet-1K images, outperforming Random by a significant margin up to 19.1\% (\cref{fig:costs}). To understand how well our method performs on high-resolution images, we also extended the evaluation to three high-resolution ImageNet-1K subsets: ImageNette (assorted objects), ImageWoof (dog breeds), and ImageSquawk (birds) datasets, all with a resolution of 128$\times$128. As shown in \cref{tab:imagenet}, we outperform prior works with the same model architecture across all settings. Notably, \texttt{D2M} improves the performance of the top competitor, {FRePo}, on the ImageWoof by more than 2.7\%. This improvement stems from the fact that \texttt{D2M} effectively distills essential information from the original dataset into the generative model through channel-wise attention matching and softened output prediction alignment. Our ablation studies in \cref{subsec:ablation} further confirm that the performance gain is directly related to the method's ability to generate informative images. 

To further evaluate our method's effectiveness, we compare \texttt{D2M} to GLaD \cite{cazenavette2023generalizing} using ImageNet-A to ImageNet-E, as well as ImageFruit and ImageMeow (cat breeds). It should be noted that GLaD is a plug-and-play method and can be added to existing dataset distillation studies. \cref{tab:glad} presents an analysis of performance between \texttt{D2M} and GLaD when applied to previous techniques, namely DC \cite{zhao2021datasetDC}, DM \cite{zhao2023dataset}, and MTT \cite{cazenavette2022dataset}. In each case, we distilled a synthetic dataset with one image per class and then evaluated it on the Depth-5 ConvNet. The results show \texttt{D2M}'s superiority across all experimental settings. 

\begin{table}[t]
\centering
\caption{The performance (\%) comparison to prior works on 128$\times$128 ImageNet-Subset (ImageFruit, ImageMeow, and ImageNet-[A, B, C, D, E]) with IPC1.}
\footnotesize
\resizebox{1\linewidth}{!}{
\begin{tabular}{c|ccccccc}
\toprule
Methods & ImageFruit   & ImageMeow   &     ImageNet-A     & ImageNet-B        & ImageNet-C   & ImageNet-D   & ImageNet-E \\ \midrule
DC \cite{zhao2021datasetDC}   & $21.0_{\pm0.9}$ & $22.0_{\pm0.6}$   & $43.2_{\pm0.6}$ & $47.2_{\pm0.7}$  & $41.3_{\pm0.7}$ & $34.3_{\pm1.5}$  & $34.9_{\pm1.5}$     \\
DM  \cite{zhao2023dataset}  & $20.4_{\pm1.9}$  & $20.1_{\pm1.2}$   & $39.4_{\pm1.8}$ & $40.9_{\pm1.7}$  & $39.0_{\pm1.3}$  & $30.8_{\pm0.9}$  & $27.0_{\pm0.8}$     \\
MTT \cite{cazenavette2022dataset}  & $22.4_{\pm1.1}$  & $26.6_{\pm0.4}$  & $51.7_{\pm0.2}$ & $53.3_{\pm1.0}$  &  $48.0_{\pm0.7}$ & $43.0_{\pm0.6}$  & $39.5_{\pm0.9}$     \\
DC+GLaD \cite{cazenavette2023generalizing}  & $20.7_{\pm1.1}$  & $22.6_{\pm0.8}$  & $44.1_{\pm2.4}$ & $49.2_{\pm1.1}$  &  $42.0_{\pm0.6}$ & $35.6_{\pm0.9}$  & $35.8_{\pm0.9}$     \\
DM+GLaD \cite{cazenavette2023generalizing}   & $21.8_{\pm1.8}$  & $22.3_{\pm1.6}$  & $41.0_{\pm1.5}$ & $42.9_{\pm1.9}$  &  $39.4_{\pm0.7}$ & $33.2_{\pm1.4}$  & $30.3_{\pm1.3}$     \\
MTT+GLaD \cite{cazenavette2023generalizing}   & $23.1_{\pm0.4}$  & $26.0_{\pm1.1}$  & $50.7_{\pm0.4}$ & $51.9_{\pm1.3}$  &  $44.9_{\pm0.4}$ & $39.9_{\pm1.7}$  & $37.6_{\pm0.7}$     \\ 
\textbf{D2M} (Ours)   & $\mathbf{24.4_{\pm0.5}}$  & $\mathbf{28.4_{\pm0.3}}$  &  $\mathbf{55.2_{\pm0.2}}$ & $\mathbf{56.8_{\pm0.6}}$  &  $\mathbf{50.4_{\pm0.4}}$ & $\mathbf{45.2_{\pm0.6}}$  & $\mathbf{41.6_{\pm0.8}}$  \\\bottomrule
\end{tabular}}
\label{tab:glad}
\end{table}

\textbf{Higher Resolution Datasets (256$\times$256).}
We extend the scaling of \texttt{D2M} to higher resolutions, specifically datasets with a resolution of 256$\times$256. To that end, we conduct experiments on two datasets: ImageSquawk and DermaMNIST. The results, shown in \cref{tab:imagesquack256}, demonstrate superior performance across both datasets, surpassing Random by a significant margin.

\begin{table} [t]
\centering
 \caption{GPU hours for re-distillation process from IPC1 to IPC10 and IPC50 on CIFAR-10. Total denotes the GPU hours spent on the distillation for all three IPCs.}
\scriptsize
 \setlength{\tabcolsep}{4pt}
  \begin{tabular}{c|cccccc}
    \toprule
    \multirow{2}{*}{Method}&\multicolumn{6}{c}{Distillation Time (GPU hours)} \\
    & DC\cite{zhao2021datasetDC} & DSA\cite{zhao2021datasetDSA} & DM\cite{zhao2023dataset} & MTT\cite{cazenavette2022dataset} & DAM\cite{sajedi2023datadam} & \textbf{D2M}
\\ 
    \midrule

    IPC1 & 0.2 & 0.2 & 1.2 & 1.4 & 1.2 & 4.9 \\

    IPC1$\rightarrow{}$IPC10 & 1.8 & 2.0 & 1.5 & 7.8 & 1.5 & \textbf{0}\\

    IPC1$\rightarrow{}$IPC50 & 8.3 & 8.9 & 4.7 & 38.2 & 4.8 & \textbf{0}\\
    \midrule
    Total & 10.3 & 11.1 & 7.4 & {47.4} & 7.5 & {\textbf{4.9}} \\
    \bottomrule
  \end{tabular}
    \label{tab:cross-ipc}
\end{table}

\textbf{Distillation Cost Analysis.}
In efficient learning, it is critical to consider computational costs during the distillation. We compare the re-distillation cost and the number of learnable parameters for \texttt{D2M} against priors in \cref{tab:cross-ipc} and \cref{fig:costs}, respectively. When IPCs change, previous methods require dataset re-distillation, incurring substantial computational expenses and increasing GPU hours (\cref{tab:cross-ipc}). In contrast, \texttt{D2M} conducts optimization on generative models' {parameters} rather than raw pixels, requiring only a \textit{single} distillation for varying IPCs. This precludes the need for repeated distillation, saving up to 42.5 GPU hours. Additionally, while the number of learnable parameters in pixel-wise data distillation increases with dataset size and resolution, \texttt{D2M} keeps the cost unchanged across all IPCs, saving up to 35x for IPC50, as shown in \cref{fig:costs}.

\subsection{Ablation Studies and Analysis} \label{subsec:ablation}

\textbf{Cross-Architecture Generalization.}
Building on prior studies \cite{zhao2021datasetDC, wang2022cafe, sajedi2023datadam, cazenavette2022dataset, zhao2023dataset}, we evaluate the ability of \texttt{D2M}-generated images for understanding classification tasks without overfitting to a specific architecture. We use distillation methods to create synthetic data with a default network and evaluate their performance on unseen architectures. Following the settings of \cite{sajedi2023datadam, du2023minimizing}, we experiment on the CIFAR-10 dataset with IPC 50 and test generalization with architectures like AlexNet \cite{krizhevsky2017imagenet}, VGG-11 \cite{simonyan2014very}, ResNet-50 \cite{he2016deep}, DenseNet-121 \cite{huang2017densely}, and ViT \cite{dosovitskiy2020image}. The findings in \cref{tab:cross} show that synthetic images from \texttt{D2M} not only generalize well across various models but also outperform the best competitor on average up to a margin of 3.9$\%$. \texttt{D2M} illustrates the robustness of vision transformers and complex architectures like ResNet-50 and DenseNet-121 due to its powerful generative capability and distilled knowledge. This suggests \texttt{D2M} can identify crucial learning features beyond mere knowledge matching.

\begin{table} [t]
\centering
\caption{Cross-architecture performance on CIFAR-10 with 50 images per class.}
 \scriptsize
\resizebox{1\linewidth}{!}{
  \begin{tabular}{c|ccccc|c}
   \toprule
 \multirow{2}{*}{Method}&\multicolumn{5}{c|}{\textbf{Unseen} Evaluation Architecture} & \multirow{2}{*}{Average (\%)}\\
   & AlexNet \cite{krizhevsky2017imagenet} & VGG-11 \cite{simonyan2014very} & ResNet-50 \cite{he2016deep} & DenseNet-121 \cite{huang2017densely} & ViT \cite{dosovitskiy2020image} &\\
   \midrule
DSA \cite{zhao2021datasetDSA} & $53.7_{\pm0.6}$ & $51.4_{\pm1.0}$ & $24.4_{\pm0.5}$ & $24.6_{\pm1.0}$ & $43.3_{\pm0.4}$ & $39.5_{\pm0.7}$\\
CAFE \cite{wang2022cafe} & $34.0_{\pm0.6}$ & $40.5_{\pm0.8}$ & $46.7_{\pm1.2}$ & $49.7_{\pm1.3}$ & $22.7_{\pm0.7}$ & $38.7_{\pm0.9}$ \\
MTT \cite{cazenavette2022dataset} & $48.2_{\pm1.0}$ & $55.4_{\pm0.8}$ & $61.3_{\pm0.4}$  & $62.3_{\pm1.1}$ & $47.7_{\pm0.6}$ & $55.0_{\pm0.8}$ \\
DM \cite{zhao2023dataset} & $60.1_{\pm0.5}$ & $57.4_{\pm0.8}$ & $51.5_{\pm0.4}$ & $58.9_{\pm0.7}$ & $45.2_{\pm0.4}$ & $54.6_{\pm0.6}$ \\
FTD \cite{du2023minimizing} & $53.8_{\pm0.9}$ & $58.4_{\pm1.6}$ & $59.9_{\pm1.3}$ & $58.3_{\pm1.5}$ & $49.3_{\pm0.7}$ & $55.9_{\pm1.2}$ \\
DAM \cite{sajedi2023datadam} & $63.9_{\pm0.9}$ & $64.8_{\pm0.5}$ & $59.6_{\pm0.8}$ & $62.2_{\pm0.4}$ & $48.2_{\pm0.8}$ & $59.7_{\pm0.7}$ \\
\textbf{D2M} (Ours) & {$\mathbf{65.4_{\pm0.7}}$} & {$\mathbf{67.3_{\pm0.5}}$} & {$\mathbf{64.8_{\pm0.6}}$} & {$\mathbf{66.1_{\pm0.5}}$} & $\mathbf{54.7_{\pm0.6}}$ & $\mathbf{63.6_{\pm0.6}}$\\
\bottomrule
     \end{tabular}}
 \label{tab:cross}
\end{table}

\textbf{Exploring the effect of temperature $\boldsymbol{T}$ in D2M.} In our framework, soft label predictions enrich the information in the generated synthetic dataset. Inspired by \cite{hinton2015distilling, romero2014fitnets}, we use a temperature parameter $T$ to control the entropy of the output distribution while generating soft label predictions. This ablation explores the impact of temperature on overall performance. As shown in \cref{fig:ablation}\textcolor{black}{(a)}, increasing the $T$ value improves \texttt{D2M}'s performance up to a certain point ($T=$ 4). A lower temperature emphasizes only maximal logits, while a higher value flattens the distribution, focusing on different logits uniformly. Tuning the softmax temperature allows soft labels to reveal extra knowledge like class relationships for training, enabling information flow across classes (\textit{dark knowledge} \cite{hinton2015distilling}).

\textbf{Evaluation of task balance $\boldsymbol{\lambda}$ in D2M.}
In the \texttt{D2M} framework, we train the generative model by minimizing a linear combination of embedded-based and prediction-based matching losses, as expressed in \cref{eq:loss}. The parameter $\lambda$ serves as the regularizing coefficient, determining the balance between the losses. In \cref{fig:ablation}\textcolor{black}{(c)}, we study different values of $\lambda$ (ranging from 0.01 to 1000) on the CIFAR10 dataset with IPC10 to assess its sensitivity. Our default value of $\lambda = 10$ yields the best results. Values that are too high, such as 1000, reduce the effectiveness of embedding matching, resulting in a loss of informative knowledge. Conversely, too small values compromise regularization and impede the distillation of discriminative label knowledge, which is beneficial for classification tasks.

\begin{figure*} [t]
    \centering
    \includegraphics[width=\textwidth]{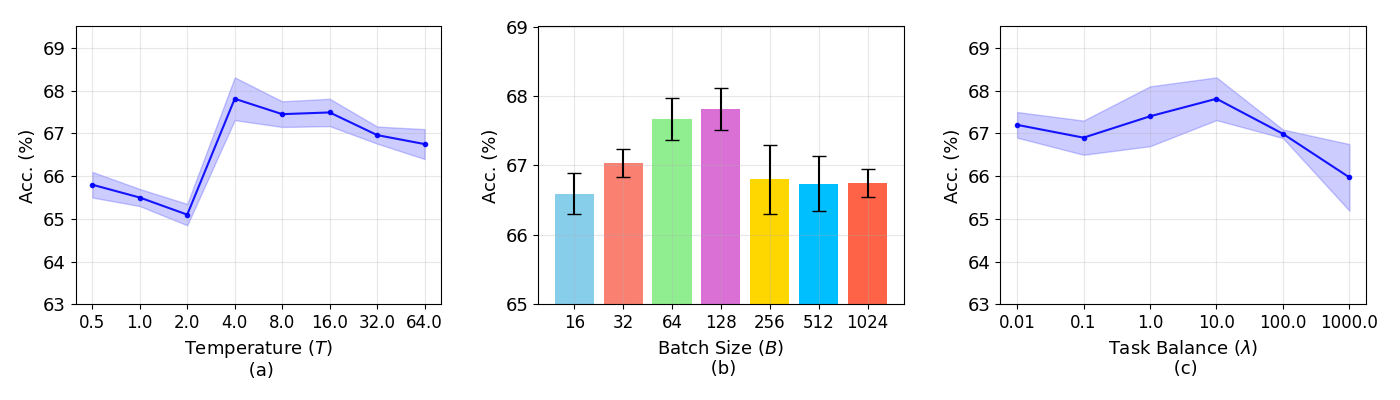}
    \caption{The effect of (a) temperature, (b)  batch size, and (c) task balance on the D2M's performance for CIFAR10 with IPC10.}
    \label{fig:ablation}
\end{figure*}

\textbf{Exploring the effect of loss components in D2M.}
We conduct an experiment to evaluate the effect of loss components, $\mathcal{L}_{\text{EM}}$ and $\mathcal{L}_{\text{PM}}$, on \texttt{D2M}'s performance. Conventional generative models strive to produce real-looking images but may lack the informative value of real images for training due to inherent information loss \cite{sajedi2023datadam, zhao2023dataset, wang2022cafe}. We refine these generators during distillation with different loss designs and observe \textcolor{black}{improvements of 10.3$\%$ and 7.9$\%$ for $\mathcal{L}_{\text{EM}}$ and $\mathcal{L}_{\text{PM}}$, respectively, on CIFAR10 with IPC10. This enhancement is attributed to the transfer of informative and discriminative knowledge for classification tasks. Moreover, combining $\mathcal{L}_{\text{EM}}$ and $\mathcal{L}_{\text{PM}}$ led to a remarkable 12.7\% improvement}. 

\vspace{+5pt}
\begin{wraptable}[]{r}{0.45\linewidth}
\centering
\vspace{-1.9em}
\caption{The performance comparison for high-resolution (256$\times$256) datasets.}
 \resizebox{1\linewidth}{!}{
\begin{tabular}{cc|cc}
\toprule
     Dataset   & IPC & Random &  \textbf{D2M}\\ 
 \midrule
\multirow{2}{*}{ImageSquawk}   
& 1 &  $22.3_{\pm0.5}$  &  $\mathbf{48.2_{\pm0.4}}$\\
& 10 & $41.3_{\pm0.6}$  &  $\mathbf{63.4_{\pm0.3}}$\\ 
\hline
\multirow{2}{*}{DermaMNIST} 
& 1 &  $21.3_{\pm1.3}$  &  $\mathbf{46.4_{\pm0.4}}$\\
& 10 & $38.4_{\pm1.1}$  &  $\mathbf{67.2_{\pm0.3}}$\\ 
\bottomrule
\end{tabular}}
\label{tab:imagesquack256}
\end{wraptable}

\begin{wraptable}[0]{r}{0.52\linewidth}
  \centering
  \vspace{-13.5em}
    \caption{The effect of different generators on D2M's performance for the CIFAR10 dataset.}
  \resizebox{1\linewidth}{!}{
  \begin{tabular}{c|ccc}
    \toprule
   IPC & 1   &  10  & 50  \\
    \midrule
    CGAN & $49.6_{\pm0.7}$ & $67.4_{\pm0.3}$  & $72.2_{\pm0.8}$ \\
   BigGAN & $50.2_{\pm0.2}$ & $67.8_{\pm0.5}$ &  $74.4_{\pm0.3}$ \\
    StyleGAN-XL & $\mathbf{50.7_{\pm0.5}}$ & $\mathbf{68.2_{\pm0.4}}$ &  $\mathbf{75.2_{\pm0.4}}$ \\   
   CVAE & $49.8_{\pm0.3}$ & $67.6_{\pm0.2}$  & $73.1_{\pm0.4}$ \\
    \bottomrule
  \end{tabular}
   }
    \label{tab:gan}
\end{wraptable} 
\vspace{+10pt}
\textbf{Evaluation of different generative models in D2M.}
We investigate the effect of different pre-trained generators on \texttt{D2M} by deploying BigGAN \cite{brock2018large}, CGAN \cite{mirza2014conditional}, StyleGAN-XL \cite{sauer2022stylegan}, and a CVAE model \cite{child2020very, sohn2015learning} on CIFAR-10 with IPC1, 10, and 50. \cref{tab:gan} demonstrates that all models effectively synthesize images for classification, with StyleGAN-XL outperforming the rest. However, BigGAN was chosen as the default due to its balance of performance and computational efficiency. Details on the computational costs are given in the supplementary.

\subsection{Visulization and Application} \label{subsec:vis}
\begin{figure}[t]
  \centering
  \begin{subfigure}[b]{0.235\textwidth}
    \includegraphics[width=\linewidth]{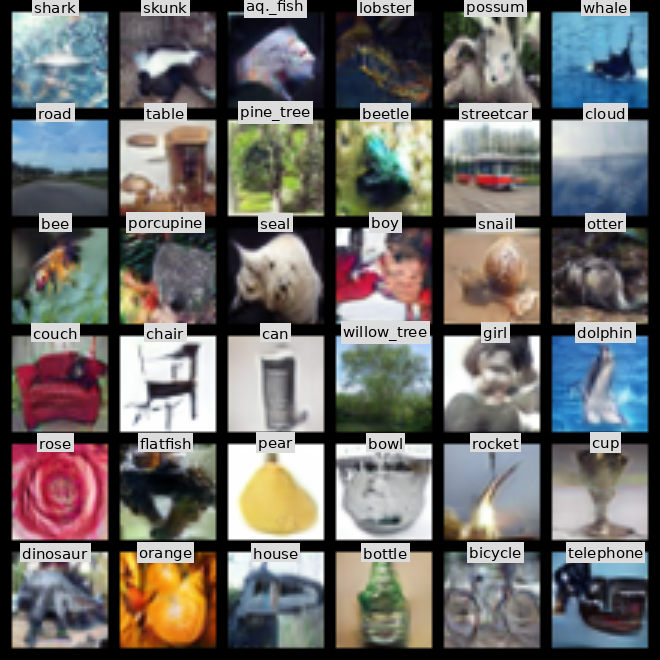}
    \caption{CIFAR-100}
    \label{fig:sub1}
  \end{subfigure}\hfill
  \begin{subfigure}[b]{0.235\textwidth}
    \includegraphics[width=\linewidth]{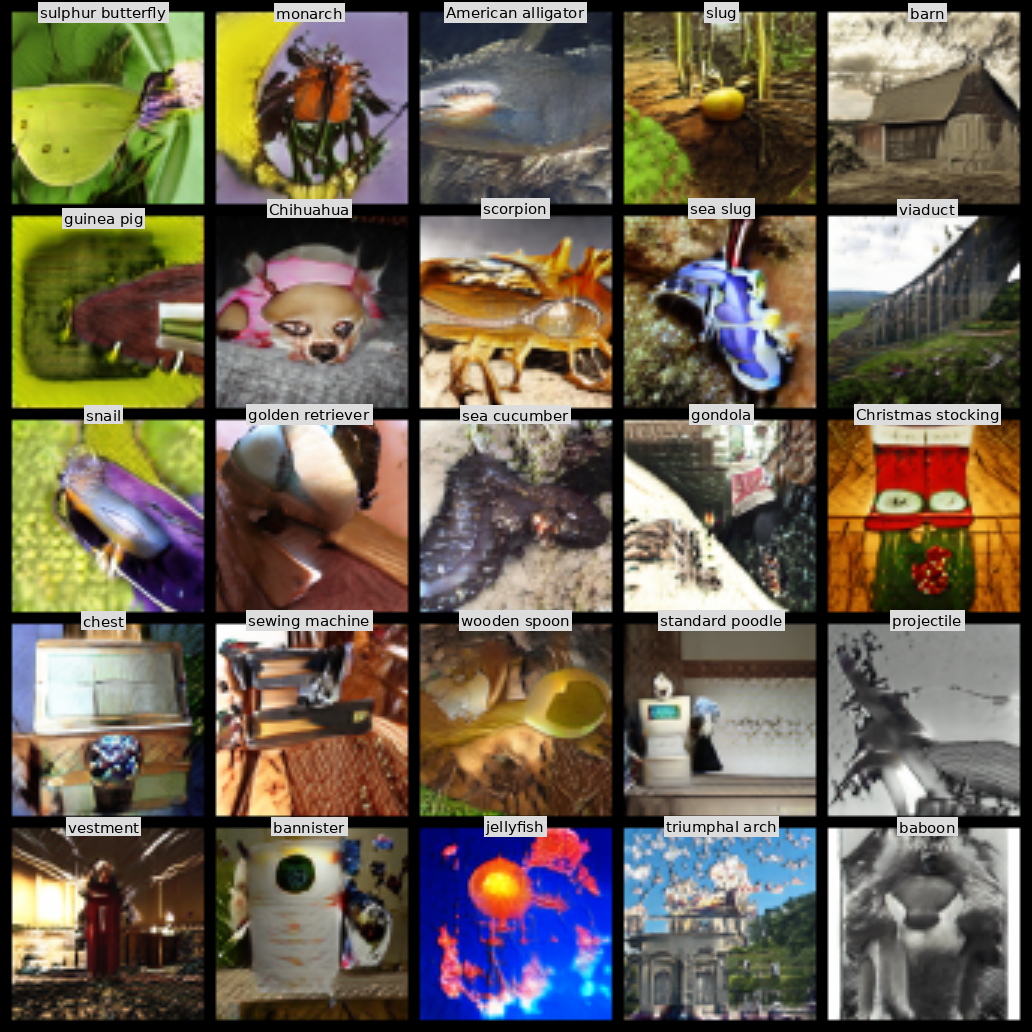}
    \caption{Tiny ImageNet}
    \label{fig:sub2}
  \end{subfigure}\hfill
  \begin{subfigure}[b]{0.235\textwidth}
    \includegraphics[width=\linewidth]{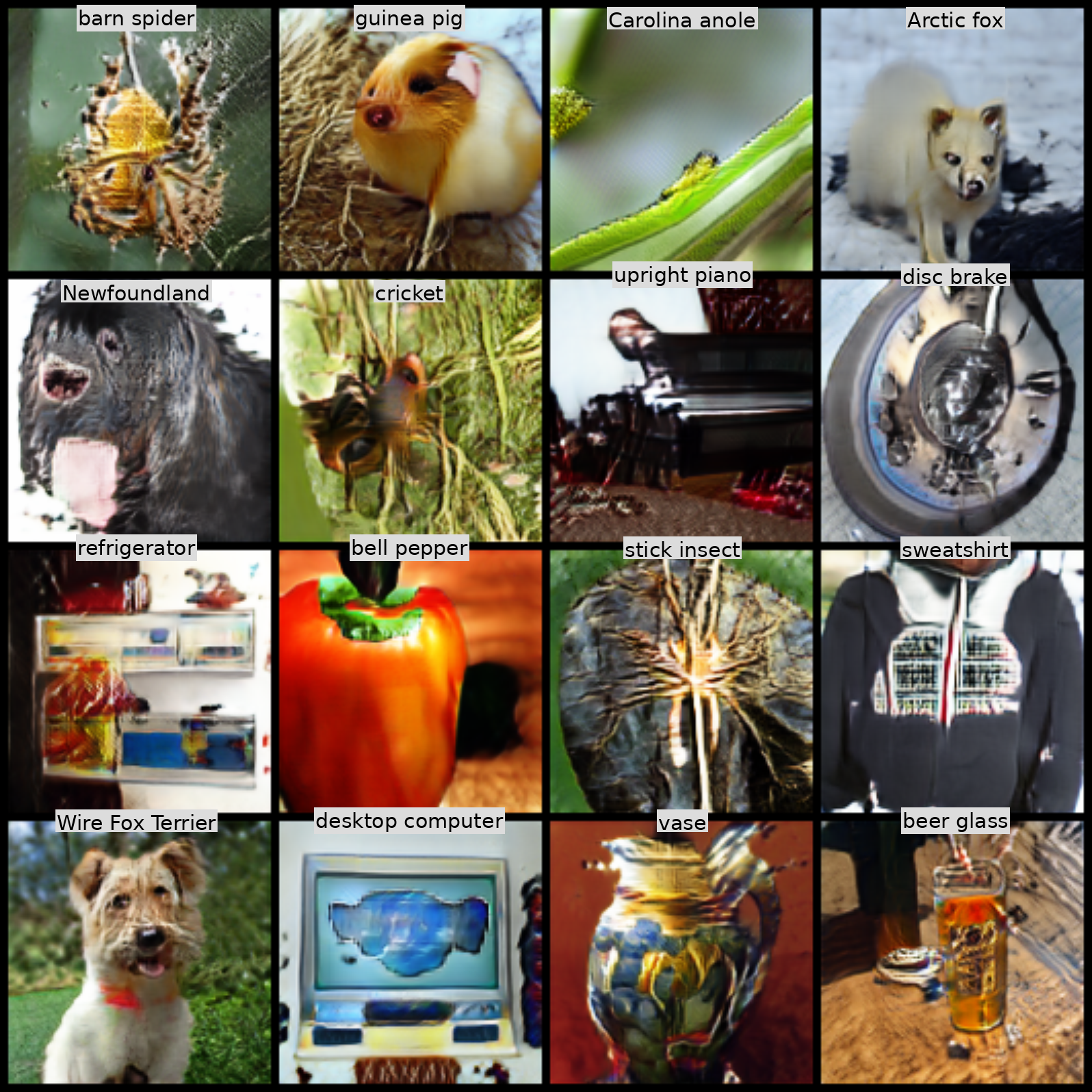}
    \caption{ImageNet-1K}
    \label{fig:sub3}
  \end{subfigure}\hfill
  \begin{subfigure}[b]{0.235\textwidth}
    \includegraphics[width=\linewidth]{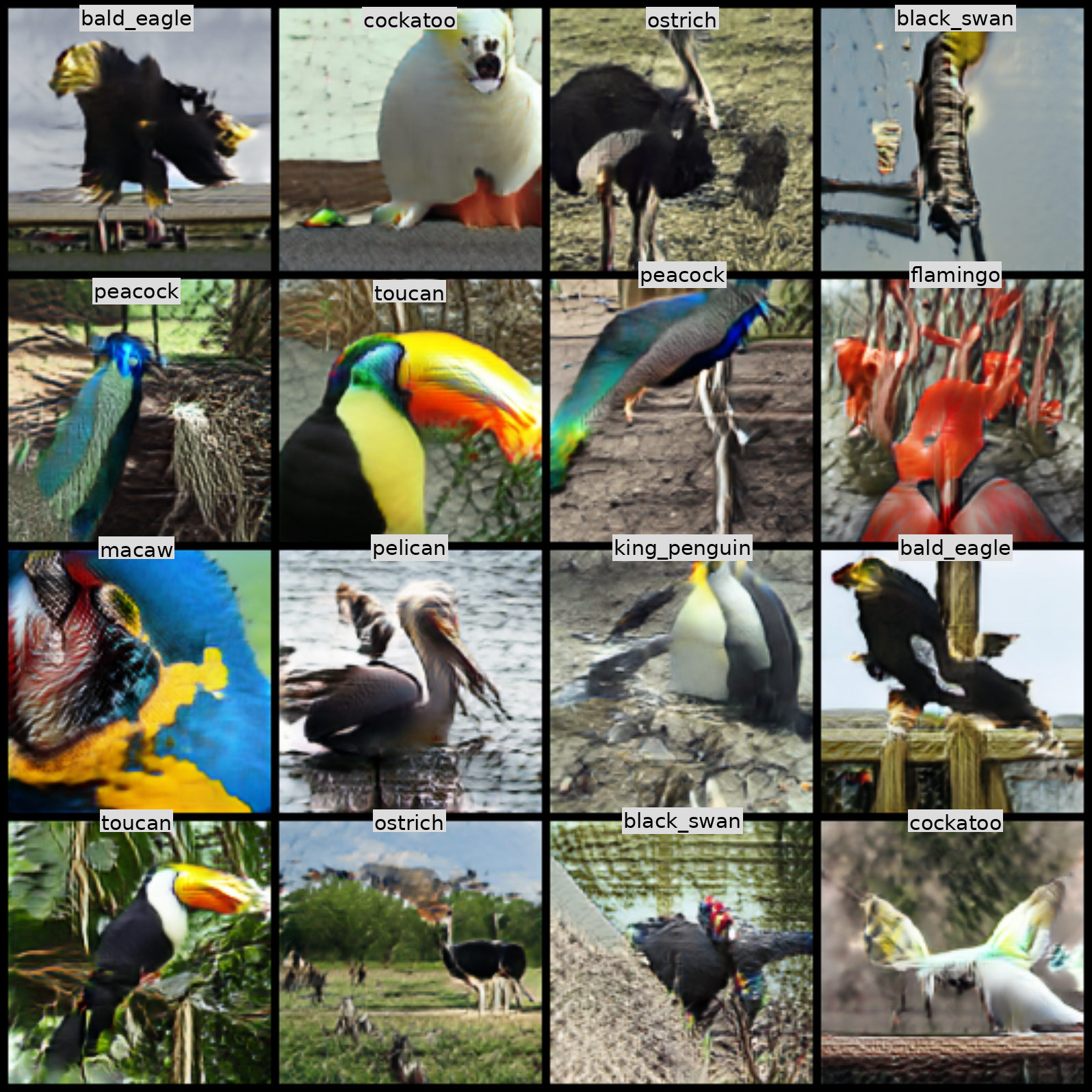}
    \caption{ImageSquawk}
    \label{fig:sub4}
  \end{subfigure} 
  \caption{Example generated images from 32$\times$32 CIFAR-100, 64$\times$64 Tiny ImageNet, 128$\times$128 ImageNet-1K, and 128$\times$128 ImageSquawk.}
\label{fig:vis}
\end{figure}

\textbf{Synthetic Images.} 
We visualize synthetic images for various datasets and resolutions in \cref{fig:vis}. These images are identifiable, although they may contain artificial patterns that improve their informativeness, like discriminative features of the animals. Note that our generative model prioritizes informativeness over realism. More visualizations are available in the supplementary materials.

\textbf{Neural Architecture Search.}
The synthetic images can act as a proxy set to accelerate model evaluation in the Neural Architecture Search (NAS). Following \cite{sajedi2023datadam, zhao2021datasetDC}, we conduct NAS on CIFAR-10 with a search space of 720 ConvNets, varying in network depth, width, activation, normalization, and pooling layers. More implementation details can be found in the supplementary materials. We train all architectures on the Random, {DSA}, {CAFE}, {DM}, DAM, and \texttt{D2M} synthetic sets with IPC50 for 200 epochs and on the full dataset for 100 epochs to establish a baseline. We rank architectures based on validation performance and report the testing accuracy of the best-selected model when trained on the test set. Spearman’s rank correlation also evaluates the reliability of synthetic images for architecture search, comparing testing performances between models trained on proxy and real training sets. As illustrated in \cref{tab:nas}, \texttt{D2M}-generated images achieve the highest performance (88.8\%) and a strong rank correlation (0.80) compared to priors. 
\begin{table}[t]
\renewcommand\arraystretch{0.7}
\centering
\caption{{Neural architecture search on the CIFAR-10 dataset.}}
\scriptsize
\resizebox{1\linewidth}{!}{
 \setlength{\tabcolsep}{2.2pt}
\begin{tabular}{ccccccc|c}
\toprule
                    & Random            & DSA \cite{zhao2021datasetDSA}               & DM \cite{zhao2023dataset}   & DAM \cite{sajedi2023datadam} & CAFE \cite{wang2022cafe} & \textbf{D2M} (Ours) & Whole Dataset \\ \midrule
Performance (\%)   &        88.9         &             87.2   &         87.2  &  88.8 &  83.6  & {\bf{88.9}}  & 89.2  \\
Correlation    &        0.70         &        0.66        &       0.71     &  0.72  & 0.59 &  {\bf{0.80}}   &  1.00\\
Time cost (min)     & 206.4  &   206.4    &  206.6 &   206.4 &  206.4  &   206.5     & 5168.9 \\ 
 Number of Images      & \bf{500}          & \bf{500}  &   \bf{500}     & \bf{500}        &   \bf{500}  &   \bf{500}  & $5\times 10^4$   \\
\bottomrule
\end{tabular}}
\label{tab:nas}
\end{table}

\section{Conclusion and Limitations} \label{sec:conclusion}
In this work, we introduce an efficient framework (\texttt{D2M}), which distills information from large-scale training datasets into the parameters of a pre-trained generative model instead of synthetic images. Our generative model employs carefully designed matching modules for distillation to enhance re-distillation efficiency, scalability to high-resolution datasets, and cross-architecture generality. We conduct extensive experiments across various datasets to demonstrate the efficiency and effectiveness of \texttt{D2M}. However, \texttt{D2M} needs more GPU memory in IPC1 compared to data distillation methods (\cref{fig:costs}). In the future, we plan to apply a more efficient generative model to small IPCs and explore \texttt{D2M}'s potential for other downstream tasks like object detection and segmentation.


%
%
\bibliographystyle{splncs04}
\bibliography{main}

\clearpage
\appendix
\setcounter{page}{1}

\section{Supplementary Materials}
Here we include additional details of our experiments, evaluation framework, and further results.


\section{Clarifying the Evaluation Framework} \label{sec:eval}
Early on in this project, we struggled to find a fair way to evaluate our distillation method against other image distillation methods. In previous iterations of this work, we proposed our setting as a fixed IPC comparison. Under this setting, we generate a fixed number of images to compare with other IPC-based works such as CAFE\cite{wang2022cafe}, DataDAM\cite{sajedi2023datadam}, MTT\cite{cazenavette2022dataset}, etc. Ultimately we evolved this setting into our current version at ECCV. In our current setting, and previous iterations, we generate a fixed number of images per class, however there are multiple ways to \textit{fix} the number of images per class. One could match the storage costs, or in our case, the latency. We measure the time it takes to train a downstream network on IPC images, this creates a fixed time cost per network, and per configuration. We can use this time cost to distribute the number of images we produce against the time taken to train our downstream network. We note that this \textit{still} retains a fixed number of images and that the raw amount of images is determined by the fixed time ratio per configuration. In previous iterations of our work, we simply refer to this setting as \textit{fixed} IPC, which in turn is correct, but we feel adding this message is valuable for the community to discuss and understand how our definition of \textit{fixed IPC} may differ from previous works, while still retaining a fair comparison. In conclusion our evaluation protocols match the evaluation time constraints of training a target downstream model on IPC images, however we divide this time between generation of a fixed number of images (determined by the time ratio) and the actual training. This protocol still differs from previous works such as DiM\cite{wang2023dim}, as we are not generating on the fly -- rather allocated a fixed time block to generate a fixed number of image such that the total time remains equivalent. 

\section{Implementation Details} \label{impl}
\subsection{Datasets} \label{datasets}
We conduct a comprehensive series of experiments involving a variety of datasets, including CIFAR10/100 \cite{krizhevsky2009learning}, Tiny ImageNet \cite{le2015tiny}, ImageNet-1K \cite{deng2009imagenet}, and specific subsets of ImageNet-1K, namely ImageNette \cite{howard2019imagenette}, ImageWoof \cite{howard2019imagenette}, and ImageSquawk \cite{cazenavette2022dataset}.
The CIFAR dataset, widely recognized in the field of low-resolution computer vision, comprises images of common objects rendered in 32$\times$32 pixels. This dataset is divided into two parts: CIFAR10, containing 10 broad categories, and CIFAR100, featuring 100 more detailed categories. Each part consists of 50,000 training images and 10,000 for testing. The CIFAR10 dataset classifies images into distinct categories such as "airplane", "car", "bird", "cat", "deer", "dog", "frog", "horse", "ship", and "truck".
The Tiny ImageNet dataset is derived from the larger ImageNet-1K and includes 200 classes, with 100,000 training images and 10,000 testing images, all downsized to 64$\times$64 pixels. In contrast, the more extensive ImageNet-1K dataset encompasses 1,000 categories, boasting over 1.2 million training images and 50,000 testing images. To align with the specifications of Tiny ImageNet, we resized the ImageNet-1K images to 64$\times$64 pixels, following established procedures outlined in \cite{zhoudataset, zhao2023dataset, sajedi2023datadam}.
The Tiny ImageNet and ImageNet-1K datasets present a higher level of complexity compared to CIFAR-10/100 due to their broader class range and increased image resolution. Our study goes further by incorporating high-resolution images, specifically the 128$\times$128 pixel versions of ImageNet-1K and its subsets. While previous dataset distillation studies \cite{cazenavette2022dataset, cazenavette2023generalizing} focused on subsets categorized by themes such as birds, fruits, and cats, our research takes a more comprehensive approach by including several subsets like ImageNette (assorted items), ImageWoof (dog breeds), ImageSquawk (bird species), ImageFruit, and ImageMeow (cat species). This demonstrates the robustness of our method, \texttt{D2M} across a broad range of datasets. 

It is worth noting that we have finally extended the application of \texttt{D2M} to higher resolutions, specifically 256$\times$256 for the Image-Squawk dataset and medical imaging DermaMNIST \cite{yang2023medmnist}. The DermaMNIST makes use of the HAM10000 dataset \cite{tschandl2018ham10000}, which is a comprehensive collection of dermatoscopic images depicting various common pigmented skin lesions. This dataset includes 10,015 dermatoscopic images classified into seven different diseases, posing a multi-class classification challenge. To facilitate model training and evaluation, we divided the images into training, validation, and test sets with a 7:1:2 ratio. An overview of the statistical characteristics of the datasets used in our study is presented in \Cref{tab:data}.
\begin{table}[t]
\renewcommand\arraystretch{0.7}
\centering
\caption{Statistical properties of the datasets employed in the study.}
\footnotesize
 \setlength{\tabcolsep}{2.2pt}
\begin{tabular}{c|ccc}
\toprule
                   Training Dataset & $\#$ of Classes  & $\#$ of Images per class & Resolution \\ \midrule
CIFAR-10 \cite{krizhevsky2009learning}   &    10  &   5000  &   32$\times$32   \\
CIFAR-100 \cite{krizhevsky2009learning}     &        100        &       500      &       32$\times$32   \\
Tiny ImageNet  \cite{le2015tiny}   & 200  &   500    &  64$\times$64 \\ 
ImageNet-1K ($128 \times 128$)  \cite{deng2009imagenet}   & 1000         &   $\approx$ 1200         &  64$\times$64 \\
ImageNet-1K Subsets (256$\times$256) \cite{cazenavette2023generalizing, cazenavette2022dataset}   & 10        &   $\approx$ 1000-1200         &  128$\times$128 \\
DermaMNIST  \cite{yang2023medmnist}   & 7  &   $\approx$ 1400    &  256$\times$256 \\ 
\bottomrule
\end{tabular}
\label{tab:data}
\end{table}

Additionally, in the supplementary materials, we conducted a comparative analysis between our methodology and GLaD \cite{cazenavette2023generalizing}, applying it to newly created high-resolution (128$\times$128 pixel) 10-class subsets from ImageNet-1K. These high-resolution subsets were created based on the performance evaluation of a pre-trained ResNet-50 \cite{he2016deep} model on ImageNet-1K, as detailed in \cite{cazenavette2023generalizing}. Our experiments contained "ImageNet-A", which included the top 10 classes, followed by "ImageNet-B" with the next 10, and so forth, up to "ImageNet-E". For a comprehensive list of the ImageNet classes within each subset, please refer to \Cref{tab:classes}.

\begin{table*}[t]
\centering
\caption{Class listings for the ImageNet-1K \cite{deng2009imagenet} subsets.}
\setlength{\tabcolsep}{3pt}
\resizebox{1\linewidth}{!}{
\begin{tabular}{c|cccccccccc}
\toprule
Dataset    & 0                & 1                & 2               & 3                  & 4              & 5           & 6               & 7           & 8                  & 9                \\ \hline
ImageNette \cite{howard2019imagenette}   & Tench            & English Springer & Cassette Player & Chainsaw           & Church         & French Horn & Garbage Truck   & Gas Pump    & Golf Ball          & Parachute          \\\hline
ImageWoof  \cite{howard2019imagenette}     & Australian Terrier        & Border Terrier       & Samoyed     & Beagle        & Shih-Tzu   & English Foxhound        & Rhodesian Ridgeback           & Dingo     & Golden Retriever       & English Sheepdog             \\\hline
ImageSqauwk \cite{cazenavette2022dataset}     & Peacock          & Flamingo              & Macaw        & Pelican             & King Penguin     & Bald Eagle     & Toucan      & Ostrich    & Black Swan              & Cockatoo           \\\hline
ImageFruit \cite{cazenavette2022dataset}     & Pineapple    & Banana   & Strawberry  & Orange   &  Lemon  & Pomegranate  & Fig      & Bell Pepper  & Cucumber  & Granny Smith Apple   \\\hline
ImageMeow \cite{cazenavette2022dataset}  & Tabby Cat  & Bengal Cat    & Persian Cat    & Siamese Cat   & Egyptian Cat   & Lion    & Tiger  & Jaguar &  Snow Leopard  & Lynx           \\\hline
ImageNet-A \cite{cazenavette2023generalizing}      & Leonberg          & Probiscis Monkey         & Rapeseed           & Three-Toed Sloth            & Cliff Dwelling   & Yellow Lady's Slipper  & Hamster          & Gondola     & Orca         & Limpkin         \\ \hline
ImageNet-B  \cite{cazenavette2023generalizing}    & Spoonbill        & Website           & Lorikeet      & Hyena             & Earthstar          & Trollybus & Echidna             & Pomeranian & Odometer           & Ruddy Turnstone     \\ \hline
ImageNet-C  \cite{cazenavette2023generalizing}   & Freight Car      & Hummingbird        & Fireboat      & Disk Brake             & Bee Eater       & Rock Beauty   & Lion        & European Gallinule        & Cabbage Butterfly            & Goldfinch \\\hline
ImageNet-D   \cite{cazenavette2023generalizing}    & Ostrich     & Samoyed           & Snowbird         & Brabancon Griffon             & Chickadee    & Sorrel   & Admiral       & Great Gray Owl   & Hornbill              & Ringlet           \\\hline
ImageNet-E \cite{cazenavette2023generalizing} & Spindle          & Toucan        & Black Swan      &  King Penguin & Potter's Wheel         & Photocopier         & Screw & Tarantula  & Sscilloscope & Lycaenid            \\\hline
\end{tabular}}
\label{tab:classes}
\end{table*}

\subsection{Implementations of Prior Works} \label{prior}
To ensure a fair comparison with prior studies, including \cite{sajedi2023datadam, zhao2021datasetDC, zhao2021datasetDSA, zhao2022synthesizing, zhao2023dataset, zhao2023improved, wang2022cafe, cazenavette2022dataset, cazenavette2023generalizing, du2023minimizing, zhou2023dataset, nguyen2021dataset, nguyen2021dataset2, cui2023scaling}, we obtained publicly available distilled data corresponding to each standard method and maintained consistent experimental settings during the evaluation phase. Our experimental setup featured a ConvNet architecture, which consisted of three, four, or five layers depending on the image resolutions. We also applied the same preprocessing techniques across all methodologies. In cases where our results matched or fell below the findings reported in their original papers, we directly presented their default values.

Kernel-based approaches, such as Kernel Inducing Points (KIP) \cite{nguyen2021dataset2, nguyen2021dataset} and FRePo \cite{zhoudataset}, employ significantly larger neural networks compared to other baseline methods. Specifically, the original KIP uses a larger model with a width of 1024 for evaluation, as opposed to the 128 used by other approaches. KIP also incorporates an additional convolutional layer compared to other methods. In the case of FRePo, it utilizes a distinct model that doubles the number of filters when the feature map size is halved. Furthermore, FRePo employs batch normalization instead of instance normalization. We made diligent efforts to replicate previous methods that lacked experiments on specific datasets by following the provided author codes. However, for techniques that encountered scalability challenges with high-resolution datasets, we were unable to obtain relevant performance metrics.

\begin{table*} [t]
    \centering
        \caption{Hyperparameters Details used in the D2M framework.}
    \resizebox{\textwidth}{!}{
    \normalsize{
    \begin{tabular}{c|c|c||c||c}
    \hline  
    \multicolumn{3}{c||}{\textbf{Hyperparameters}} &
    \multirow{2}{*}{} \textbf{Options/} &
    \multirow{2}{*}{\textbf{Value}} \\
    \cline{1-3}
    \textbf{Category} & \textbf{Parameter Name} & \textbf{Description} & \textbf{Range} & \\
    \hline 
    \hline
    
    \multirow{9}{*}{\textbf{Optimization}}
    & \textbf{Learning Rate $\boldsymbol{\eta_{\mathcal{S}}}$ (generator) } & Step size towards global/local minima & $(0, 10.0]$ & $ 1.0e-6$ \\ 
    \cline{2-5}

    & \textbf{Learning Rate $\boldsymbol{\eta_{\mathcal{\boldsymbol{\theta}}}}$ (network)} & Step size towards global/local minima & $(0, 1.0]$ & $0.01$ \\ 
    \cline{2-5}
    
    \multirow{4}{*}{} & 
    
    \multirow{1}{*}{\textbf{Optimizer (generator)}} & \multirow{1}{*}{Updates synthetic set to approach global/local minima} & Adam & betas: ($0.5$,$0.9$) \\
    \cline{2-5}
    & \multirow{2}{*}{\textbf{Optimizer (network)}} & \multirow{2}{*}{Updates model to approach global/local minima} & SGD with & Momentum: $0.9$ \\
    & & & Momentum & Weight Decay: $5e-4$\\
    
    \cline{2-5}
    
    \multirow{3}{*}{} & 

    \multirow{1}{*}{\textbf{Scheduler (generator)}} & - & - & - \\
    \cline{2-5}
    &\multirow{2}{*}{\textbf{Scheduler (network)}} & \multirow{2}{*}{Decays the learning rate over epochs} & \multirow{2}{*}{StepLR} & Decay rate: $0.5$ \\
    &&&& Step size: $15.0$\\
    \cline{2-5}
     & \textbf{Epoch Count} $K$& Number of epochs for generator matching & $[1, \infty)$ & 60\\
    \hline
    \multirow{2}{*}{\textbf{Loss Function}}
    & \multirow{1}{*}{\textbf{Task Balance $\lambda$}} & \multirow{1}{*}{Regularization Multiplier} & \multirow{1}{*}{$[0, \infty)$} & 100 \\
    \cline{2-5}
    &  \textbf{Temperature Value} \bf{$T$} & Temperature in KL Divergence & $[0, \infty)$ & 4 \\
    \hline
    \multirow{8}{*}{\textbf{DSA Augmentations}}
    & \multirow{3}{*}{\textbf{Color}} &  \multirow{3}{*}{Randomly adjust (jitter) the color components of an image} & brightness & 1.0\\
    & & & saturation & 2.0\\
    & & & contrast & 0.5\\
    \cline{2-5}  
     & \textbf{Crop} & Crops an image with padding & ratio crop pad & 0.125 \\
     \cline{2-5}
     & \textbf{Cutout} & Randomly covers input with a square & cutout ratio & 0.5 \\
    \cline{2-5}
    & \textbf{Flip} & Flips an image with probability p in range: & $(0, 1.0]$ & $0.5$ \\
    \cline{2-5}
    & \textbf{Scale} & Shifts pixels either column-wise or row-wise & scaling ratio & $1.2$ \\
    \cline{2-5}
    & \textbf{Rotate} & Rotates image by certain angle & $0^{\circ} - 360^{\circ}$ & $[-15^{\circ}, +15^{\circ}]$ \\
    \cline{2-5}    
    \hline    
    \multirow{4}{*}{\textbf{Encoder Parameters}} & \textbf{Conv Layer Weights} & The weights of convolutional layers & $\mathbb{R}$ bounded by kernel size & Uniform Distribution  \\
    \cline{2-5}
    
    & \textbf{Activation Function} & The non-linear function at the end of each layer & - & ReLU \\
    \cline{2-5}
    & \textbf{Normalization Layer} & Type of normalization layer used after convolutional blocks & - & InstanceNorm\\
    \cline{2-5}
    & \textbf{Default Generator} & Type of Generator used to produce synthetic distilled images & - & BigGan\\
    \hline
    \end{tabular}
    }       }
    \label{tab: hyperparameters}
\end{table*}

\subsection{Hyperparameters} \label{hyperparam}
To ensure the reproducibility of our methodology, we have provided \Cref{tab: hyperparameters}, which details all the hyperparameters employed in this study. For the baseline methods, we adhered to the default parameters specified by the authors in their respective papers. Uniform hyperparameter settings were maintained throughout all experiments unless otherwise noted.
Specifically, we utilized an SGD optimizer with a learning rate of 1e-6 during the distillation stage and a learning rate of 0.01 when training neural network models during the evaluation phase. For low-resolution datasets, we implemented a 3-layer ConvNet architecture, while for medium- and high-resolution datasets, we adopted the recommendations outlined in \cite{zhao2023dataset}, employing 4-layer and 5-layer ConvNet architectures, respectively. Across all experiments, a mini-batch size of 128 real images was used for training over 60 epochs. We consistently applied a set of differentiable augmentations \cite{zhao2021datasetDSA} during both the distillation and evaluation phases.

\section{Additional Results and Further Analysis} \label{add-results}

\subsection{Comparison to More Baselines} \label{baselines}

\textbf{Comparison to IT-GAN \cite{zhao2022synthesizing}.} Similar to GLaD \cite{cazenavette2023generalizing}, IT-GAN generates synthetic datasets in a generative model's latent space. This is done by initially inverting the entire training set into the latent space and then fine-tuning the latent representations based on the distillation objective. IT-GAN does not reduce the number of synthesized images; they remain equal to the original dataset count. This differs from the typical aim of dataset distillation. For a fair comparison with \texttt{D2M}, we take a small subset (IPC10) of the IT-GAN-generated data (created with BigGAN \cite{brock2018large}) as the "distilled dataset". We then evaluate this subset using ConvNet-3, comparing the results with \texttt{D2M}. IT-GAN achieved a 59.7\% accuracy performance on CIFAR-10, a decrease of 8.1\% compared to \texttt{D2M}. This finding suggests that our loss design effectively distills more informative knowledge into the generative model, aiding downstream classification tasks. It is also worth noting that, unlike \texttt{D2M}, both GLaD and IT-GAN require retraining of the distillation stage when there is a change in IPC or the distillation ratio.

\begin{table}[t]
\footnotesize
\centering
\caption{The performance comparison (\%) for DREAM as a plug-and-play method with D2M.}
 \setlength{\tabcolsep}{8pt}
 \setlength{\abovecaptionskip}{0.8cm}
\begin{tabular}{cc|ccc}
\toprule
     Dataset   & {IPC} & DREAM & \textbf{D2M}  & DREAM + \textbf{D2M}\\ 
 \midrule

\multirow{2}{*}{CIFAR-10}  
& 10 &  68.6±0.5 & 67.8±0.5 &  \textbf{69.5±0.3}\\
& 50 &  74.6±0.4 & 74.4±0.4 & \textbf{74.9±0.5}\\
\bottomrule
\end{tabular}
\label{tab:DreamTable}
\end{table}

\textbf{Comparison to DREAM \cite{liu2023dream}.}
In this section, we conduct a comparative analysis with a concurrent plug-and-play dataset distillation approach known as DREAM \cite{liu2023dream}. DREAM was introduced as a pre-processing technique for dataset distillation methods, serving as an informed selector when choosing images from the real training dataset for matching purposes. In the context of \texttt{D2M}, we have the opportunity to incorporate this pre-processing step to enhance the selection of images from the training data for our matching strategy. The authors of DREAM make a compelling argument that not all training data is equally beneficial for distilling knowledge from images. In our evaluation, we present the best performance achieved by DREAM when reproduced on our hardware, utilizing the optimal configuration as presented in their paper, as shown in Table \ref{tab:DreamTable}. It is important to note that comparing DREAM to our baseline \texttt{D2M} in Tab. \textcolor{black}{1} of the main paper may not be fair. DREAM serves as a plug-and-play pre-processing step that complements other distillation algorithms, including DC \cite{zhao2021datasetDC}, DM \cite{zhao2023dataset}, and MTT \cite{cazenavette2022dataset}. Therefore, to provide a fair comparison, we have integrated DREAM pre-processing into our \texttt{D2M} method and reported the results in \Cref{tab:DreamTable}. We explicitly demonstrate that this augmented approach delivers superior performance, achieving the best results on CIFAR-10, particularly for IPC10 and 50, surpassing both our default \texttt{D2M} and the best results obtained by DREAM.

\textbf{Comparison to DiM \cite{wang2023dim}.} DiM represents a concurrent approach in the realm of dataset distillation. In DiM, Wang \etal introduced a two-stage distillation process designed specifically for 10-class datasets. Initially, they train the generative model from scratch and subsequently incorporate a distillation loss to minimize differences in knowledge at the penultimate layer between real and synthetic images. In contrast, we propose a \textit{single-stage} distillation applicable to \textit{any} dataset, leveraging our novel matching modules to incorporate \textit{{attention maps}} from intermediate layers and \textit{{soft} output predictions} (i.e., dark knowledge).

\textbf{Evaluation of batch size $\boldsymbol{B}$ in D2M.}
We evaluate the sensitivity of batch size $B$ during the distillation stage by varying it within the range of 16 to 1024 on ConvNet-3. Notably, the default batch size $B=$ 128 yields the highest accuracy, as illustrated in \Cref{fig:ablation}\textcolor{black}{(b)}. Smaller batch sizes perform poorly due to their insufficient representativeness for feature matching and limited supervision from the original dataset in each iteration. In contrast, excessively large batch sizes, while providing more information, introduce optimization challenges and lead to a performance decrease. We notice that our generators remain completely stable during training. The only consideration here is the size of the optimization problem we are trying to solve. Specifically, increasing the batch size can complicate the tractability of the optimization problem and result in a more difficult optimization problem.
\begin{figure*} [t]
    \centering
    \includegraphics[width=\textwidth]{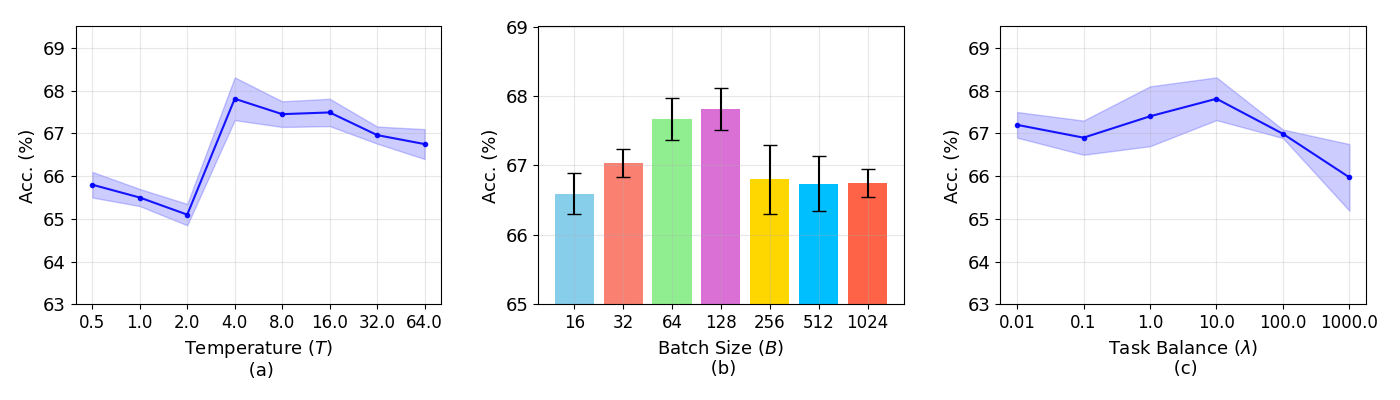}
    \caption{The effect of (a) temperature, (b)  batch size, and (c) task balance on the D2M's performance for CIFAR10 with IPC10.}
    \label{fig:ablation}
\end{figure*}

\textbf{Diffusion Model Analysis.} We extended our framework to the diffusion model (SD-XL) with LoRA fine-tuning on ImegeNette for IPC1. We achieved $52.1\%$ which indicates the generalizability of \texttt{D2M} to diffusion models.

\textbf{Discriminator in {D2M}.} Including a discriminator in the distillation process is an interesting approach, but it increases the computational graph size, making it difficult to use in all our experiments. However, we conducted experiments on CIFAR-10 IPC1, comparing three configurations of our distillation loss: matching loss only (50.2\%), matching loss with balanced discriminator loss (50.9\%), and matching loss with large discriminator loss (43.8\%). These results suggest that a strong discriminator can harm performance, while using discriminator loss as regularization improves results, likely due to reduced catastrophic forgetting. Thus, our matching loss is primarily responsible for making the images better suited for downstream training.
\subsection{Computational Cost Analysis} \label{section:costs}
\begin{figure}
    \centering
    \includegraphics[width=0.55\textwidth]{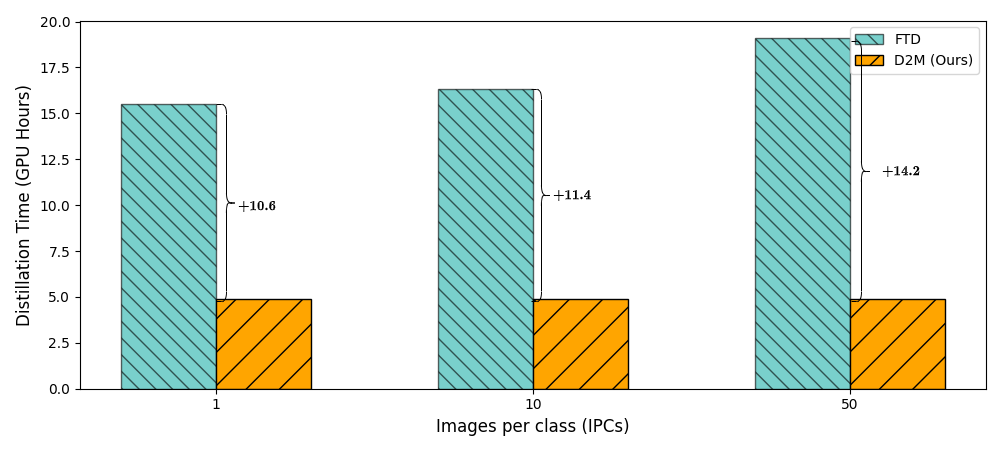}
    \caption{The GPU hours of the distillation time for FTD and D2M on CIFAR-10 with different IPCs.}
    \label{fig:ftd}
\end{figure}

In the main paper, we conducted a thorough comparison of the re-distillation costs associated with \texttt{D2M} in comparison to other leading methods on the CIFAR-10 dataset. However, it is equally crucial to assess the distillation costs for each individual IPC setting. As there is always a trade-off between performance and computational demands, we used FTD \cite{du2023minimizing} as a reference point. FTD is recognized as one of the top-tier methods in terms of performance for pixel-wise dataset distillation. We meticulously calculated the GPU hours required by both FTD and \texttt{D2M} during the distillation process on CIFAR-10 across IPC settings of 1, 10, and 50. As depicted in \Cref{fig:ftd}, \texttt{D2M}'s GPU hour requirements for distillation remain consistent across different IPCs. This is because our proposed method distills knowledge from a real dataset into the parameters of the generative model rather than the raw pixels, and this process is independent of the number of images per class used for distillation. Conversely, in the case of FTD, computational costs increase as we elevate the IPC. It is important to mention that our method consistently outperforms FTD in terms of performance and computational costs, regardless of the IPC settings, as shown in Tab. \textcolor{black}{1} of the main paper and \Cref{fig:ftd}, respectively. 

\begin{figure} [t]
    \centering
    \includegraphics[width=0.55\textwidth]{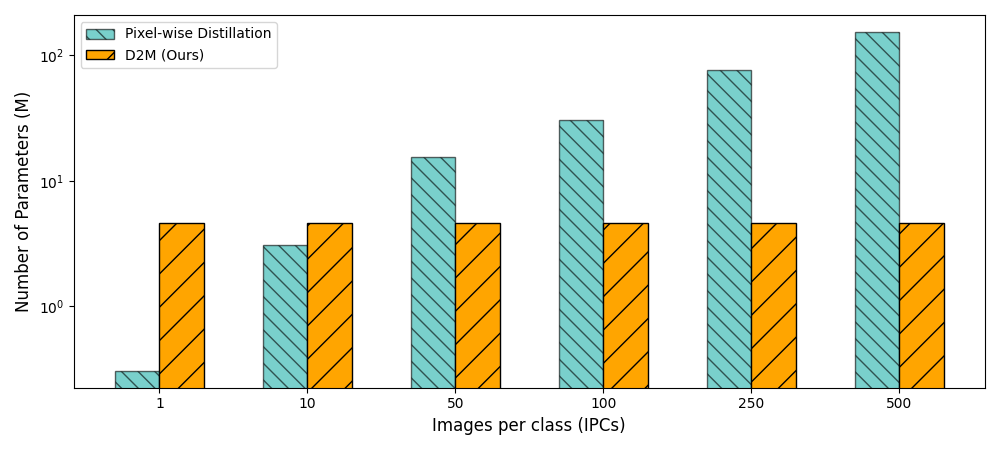}
    \caption{Comparative analysis of the number of learnable parameters on 32$\times$32 resolution CIFAR-100 with different IPCs.}
    \label{fig:costs-cifar100}
\end{figure}

The same trend holds for the number of learnable parameters. In the main paper (Fig. \textcolor{black}{3}), we observed that on ImageNet-1K, \texttt{D2M} outperforms all pixel-wise dataset distillation methods in terms of the count of learnable parameters across most IPCs. To further evaluate this, we compared the number of learnable parameters trained on less complex datasets like CIFAR-100. As illustrated in \Cref{fig:costs-cifar100}, for small IPCs, pixel-wise distillation does exhibit some advantages over \texttt{D2M}. However, as we increase the number of images per class, \texttt{D2M} significantly outperforms pixel-wise distillation by a considerable margin, especially with IPC settings higher than 20. Therefore, we can conclude that \texttt{D2M} excels in computational efficiency and performance when working with complex and high-resolution datasets such as ImageNet-1K or Tiny ImageNet. In contrast, for simpler datasets like CIFAR-10/100, \texttt{D2M} remains at the forefront in terms of performance, regardless of the IPCs; however, it proves to be more computationally efficient in larger IPC settings.

\begin{table}[t]
\centering
\caption{The GPU hours of the distillation time and the effort of image generation for different generative models on CIFAR-10.} 
\footnotesize
  \begin{tabular}{c|c|cccc}
   \toprule
Generative& Distillation Time& \multicolumn{3}{c}{Image Generation Time (ms)}\\
   Model&(GPU hours)&IPC1&IPC10&IPC50\\
   \midrule
CGAN & 2.1 & 1.46 & 5.08 & 20.58\\
BigGAN & 4.9 & 5.42 & 15.50 & 59.92\\
StyleGAN-XL & 8.9 & 26.84 & 125.33 & 518.53\\
CVAE & 7.1 & 83.06 & 481.90 & 2203.48\\
\bottomrule
     \end{tabular}
 \label{tab:compcost}
\end{table}

\begin{figure} [t]
    \centering
    \includegraphics[width=0.55\textwidth]{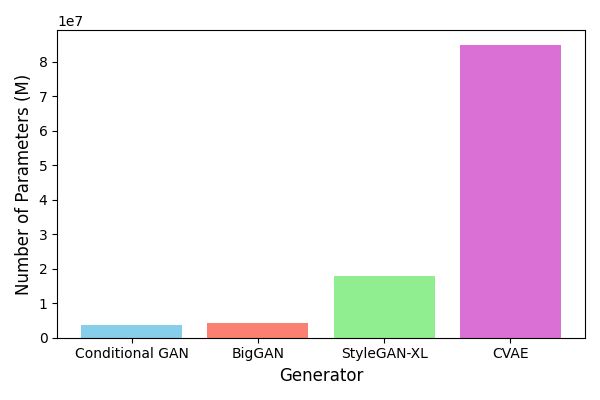}
    \caption{Number of learnable parameters for different generative models on 32$\times$32 resolution CIFAR-10.}
    \label{fig:costs-generators}
\end{figure}

\textbf{Computational costs of different generative models in D2M.}
To enhance the computational efficiency of \texttt{D2M} when applied to less complex datasets, we conducted experiments using CIFAR-10 with various generative models, including conditional GAN (CGAN) \cite{mirza2014conditional}, BigGAN \cite{brock2018large}, StyleGAN-XL \cite{sauer2022stylegan}, and CVAE \cite{child2020very, sohn2015learning}. As shown in \Cref{tab:compcost} and \Cref{fig:costs-generators}, conditional GAN stood out for its remarkably short distillation time and a minimal number of learnable parameters, respectively. However, it is worth noting that this efficiency comes at the cost of a slight reduction in performance, as evidenced in Tab. \textcolor{black}{7} of the main paper. Consequently, we can conclude that when working with simple low-resolution datasets, opting for conditional GAN is a viable choice, even if it means sacrificing a small degree of accuracy compared to other generative models. However, in this work, we have selected BigGAN as the default generative model across all experiments. This choice was made to strike a balance between performance and computational costs, particularly when dealing with extensive and high-resolution datasets. In such complex scenarios, \texttt{D2M} outperforms all other dataset distillation methods in terms of computational efficiency and performance, even in small IPCs.

Unlike the pixel-wise dataset distillation algorithms \cite{sajedi2023datadam, cazenavette2022dataset, wang2022cafe, zhao2021datasetDC, zhao2021datasetDSA, zhao2023improved, du2023minimizing, zhoudataset, liu2023dream}, \texttt{D2M} relies on generating training images in real-time during the evaluation stage. To understand the computational effort involved in this step, we conducted an experiment using the CIFAR-10 dataset to measure how long it takes to generate images with various generative models. The results, which you can find in \Cref{tab:compcost}, indicate that image generation typically takes just a few milliseconds when we use GAN-based generative models under different IPC settings. Importantly, the time spent on image generation is a small fraction of the total evaluation time, usually less than 0.6\% when using BigGAN.

\subsection{More Details on Neural Architecture Search} \label{nas}
Drawing on the studies in \cite{zhao2021datasetDC, zhao2021datasetDSA, zhao2023dataset}, we established a search domain containing 720 ConvNet configurations on the CIFAR-10 dataset. These models were evaluated using our distilled dataset, with an IPC of 50 serving as a proxy dataset within the neural architecture search (NAS) paradigm. We initiated this process with a foundational ConvNet, forming a systematic grid that varies by depth $D \in$ \{1, 2, 3, 4\}, width $W \in$ \{32, 64, 128, 256\}, activation function $A \in$ \{\text{Sigmoid, ReLu, LeakyReLu}\}, normalization technique $N \in$ \{None, BatchNorm, LayerNorm, InstanceNorm, GroupNorm\}, and pooling operation $P \in$ \{None, MaxPooling, AvgPooling\}. These variants were then evaluated and hierarchically ordered based on their validation outcomes. In line with \cite{zhao2023dataset, sajedi2023datadam}, we randomly designated 10\% of the CIFAR10 training samples as our validation subset, while the remainder formed the training set. The DSA augmentation \cite{zhao2021datasetDSA} was applied across all proxy-set methods.

\begin{figure*} [htp]
  \centering
  \begin{subfigure}[b]{0.32\textwidth}
    \includegraphics[width=\linewidth]{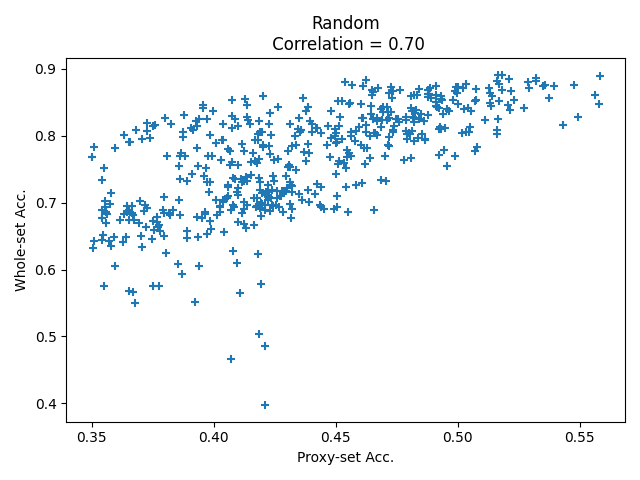}
    \caption{Random}
    \label{fig:sub2r}
  \end{subfigure}\hfill
  \begin{subfigure}[b]{0.32\textwidth}
    \includegraphics[width=\linewidth]{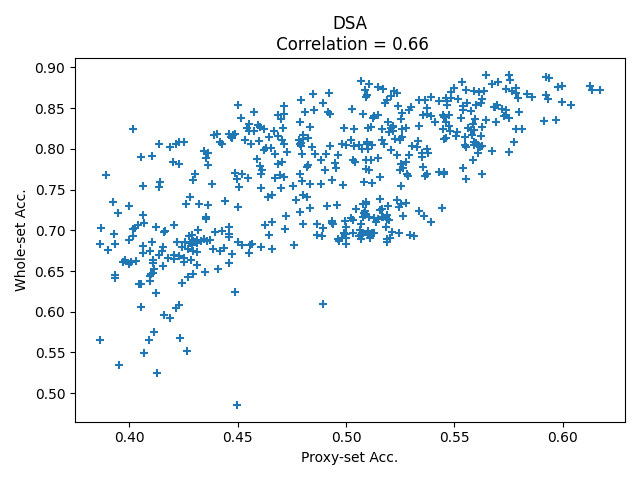}
    \caption{DSA}
    \label{fig:sub2dsa}
  \end{subfigure}\hfill
    \begin{subfigure}[b]{0.32\textwidth}
    \includegraphics[width=\linewidth]{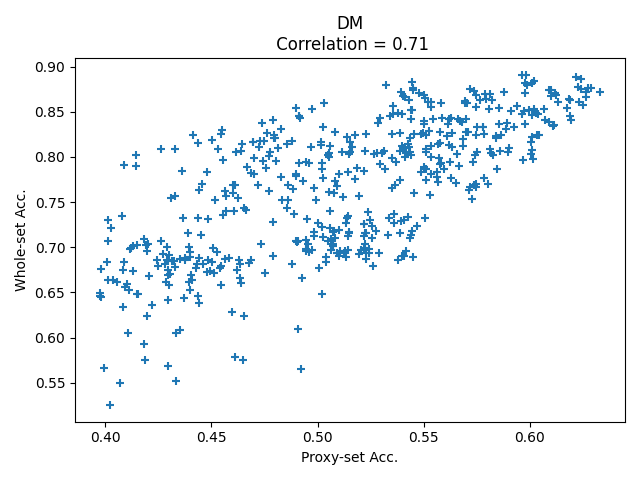}
    \caption{DM}
    \label{fig:sub2dm}
  \end{subfigure}\hfill
    \begin{subfigure}[b]{0.32\textwidth}
    \includegraphics[width=\linewidth]{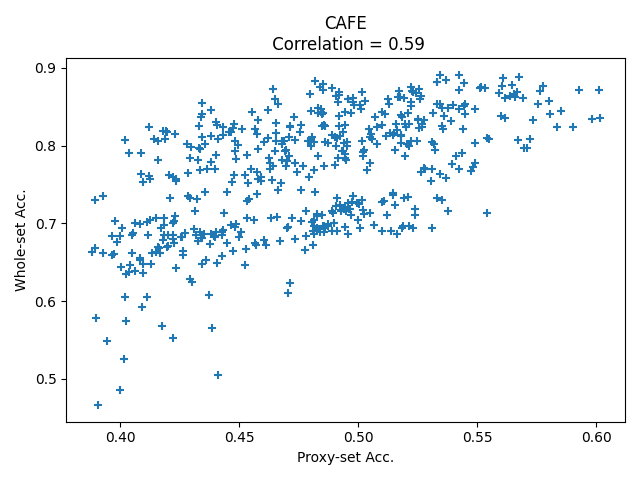}
    \caption{CAFE}
    \label{fig:sub2cafe}
  \end{subfigure}\hfill
    \begin{subfigure}[b]{0.32\textwidth}
    \includegraphics[width=\linewidth]{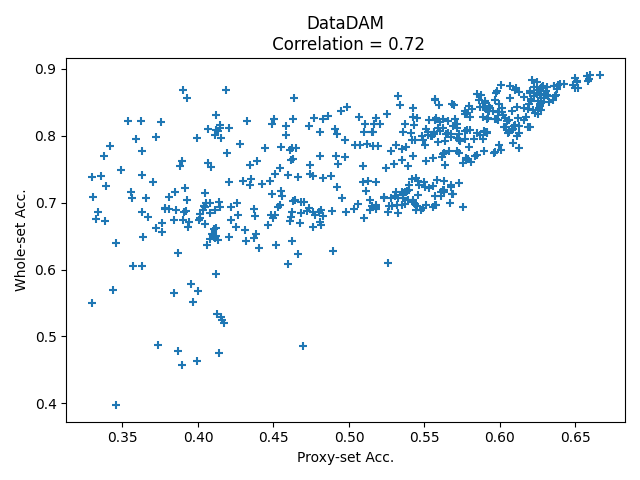}
    \caption{DataDAM}
    \label{fig:sub2dam}
  \end{subfigure}\hfill
\begin{subfigure}[b]{0.32\textwidth}
    \includegraphics[width=\linewidth]{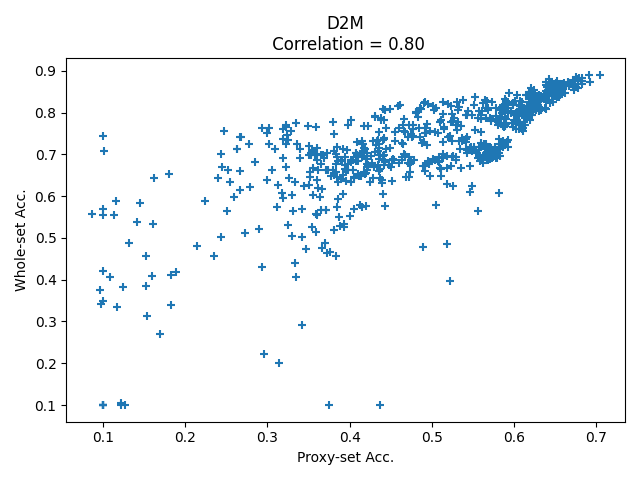}
    \caption{D2M (Ours)}
    \label{fig:sub1d2m}
  \end{subfigure}\hfill
  \caption{Performance rank correlation between proxy-set and whole dataset training across all 720 architectures.}
  \label{fig:nasfull}
\end{figure*}

In \Cref{fig:nasfull}, we illustrate the rank correlation of performance between the proxy dataset—derived from various methodologies, including Random \cite{rebuffi2017icarl}, DSA \cite{zhao2021datasetDSA}, DM \cite{zhao2023dataset}, CAFE \cite{wang2022cafe}, DataDAM \cite{sajedi2023datadam}, \texttt{D2M}—and the full training dataset. We used Spearman's rank correlation coefficient to analyze all 720 different architectural structures. Each point on the plot represents a unique architectural choice. The horizontal axis shows the test accuracy of models trained on the proxy dataset, while the vertical axis reflects the accuracy of models trained on the entire dataset. Our evaluation reveals that all methods are effective at generating reliable performance rankings for potential architectures. However, \texttt{D2M} stands out by having a larger cluster of points close to the straight line, indicating that it provides a more effective dataset for establishing dependable performance rankings of architectural choices. \texttt{D2M} achieves a remarkable correlation coefficient of 0.80, surpassing the benchmarks set by previous studies. The graph validates that our approach can yield a proxy dataset that obtains a more reliable performance ranking of candidate architectures.

\vspace{-18pt}

\section{Additional Visualizations} \label{vis}

\subsection{Data Distribution} \label{datadist}

In order to assess the ability of our method to faithfully represent the distribution inherent in the original dataset, we employ t-SNE \cite{van2008visualizing} for visualizing features extracted from both real and synthetically generated sets. These sets are produced by DC \cite{zhao2021datasetDC}, DSA \cite{zhao2021datasetDSA}, DM \cite{zhao2023dataset}, CAFE \cite{wang2022cafe}, DataDAM \cite{sajedi2023datadam}, and \texttt{D2M}, within the embedding space of the ResNet-18 architecture \cite{he2016deep}. The visualizations were conducted using the CIFAR-10 dataset with an IPC50, a consistent choice across all methodologies.

\begin{figure*} [t]
    \centering
    \includegraphics[width=\textwidth]{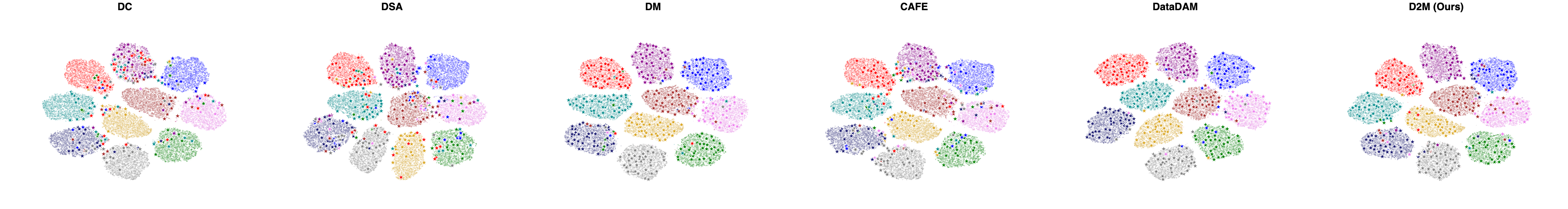}
    \caption{Distributions of the synthetic images learned by six methods on the CIFAR10 dataset with IPC 50. The stars represent the synthetic data dispersed amongst the original dataset. The classes are as follows: \color{green} plane, 
    \color{Goldenrod} car, 
    \color{blue} bird, 
    \color{violet} cat, 
    \color{MidnightBlue} deer, 
    \color{brown} dog, 
    \color{Mulberry} frog,
    \color{gray} horse,
    \color{Emerald} ship,
    \color{red} truck.}
    \label{fig:tsne}
\end{figure*}

As depicted in \Cref{fig:tsne}, our approach, akin to DM and DataDAM, successfully preserves the dataset's distribution, manifesting as a well-balanced dispersion across the entire dataset. Conversely, other methods such as DC, DSA, and CAFE exhibit noticeable biases towards certain cluster boundaries. In simpler terms, the t-SNE visualization validates that \texttt{D2M} maintains a significant degree of impartiality in accurately capturing the dataset's distribution consistently across all categories. Preservation of dataset distributions holds paramount importance, particularly in domains like ethical machine learning, as methodologies that fail to capture data distribution can inadvertently introduce bias and discrimination. Our method's capability to faithfully represent the data distribution renders it more suitable than alternative approaches, particularly in applications such as facial detection for privacy considerations \cite{ciftci2023my}.

\subsection{Visualization of the Generated Images} \label{sec:vis}

\begin{figure*} [htbp]
    \centering
    \begin{subfigure}{0.48\textwidth}
    \includegraphics[width=\linewidth]{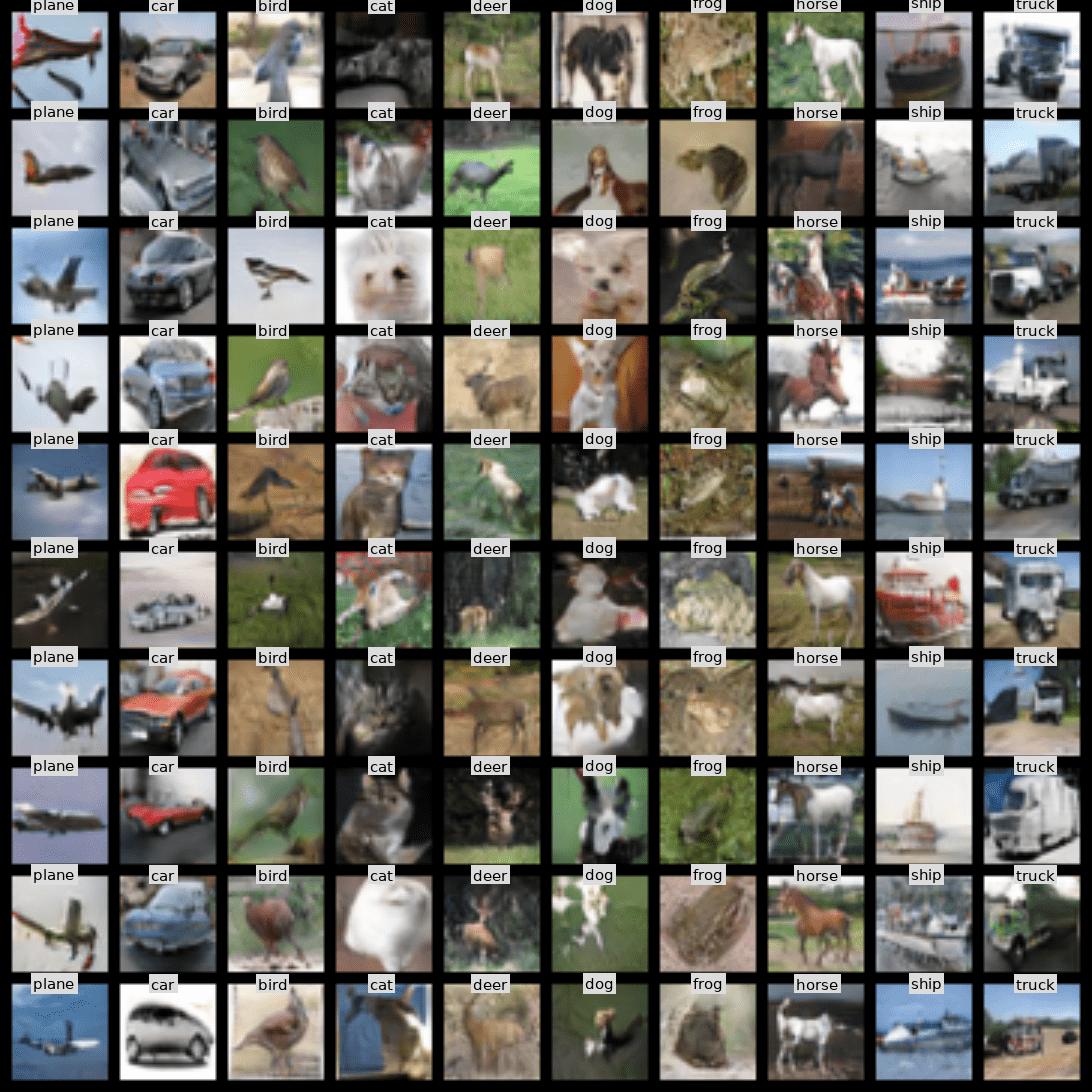}
    \caption{$T= 0.5$}
     \vspace{+5pt}
    \end{subfigure} \hfill
  \begin{subfigure}{0.48\textwidth}
    \includegraphics[width=\linewidth]{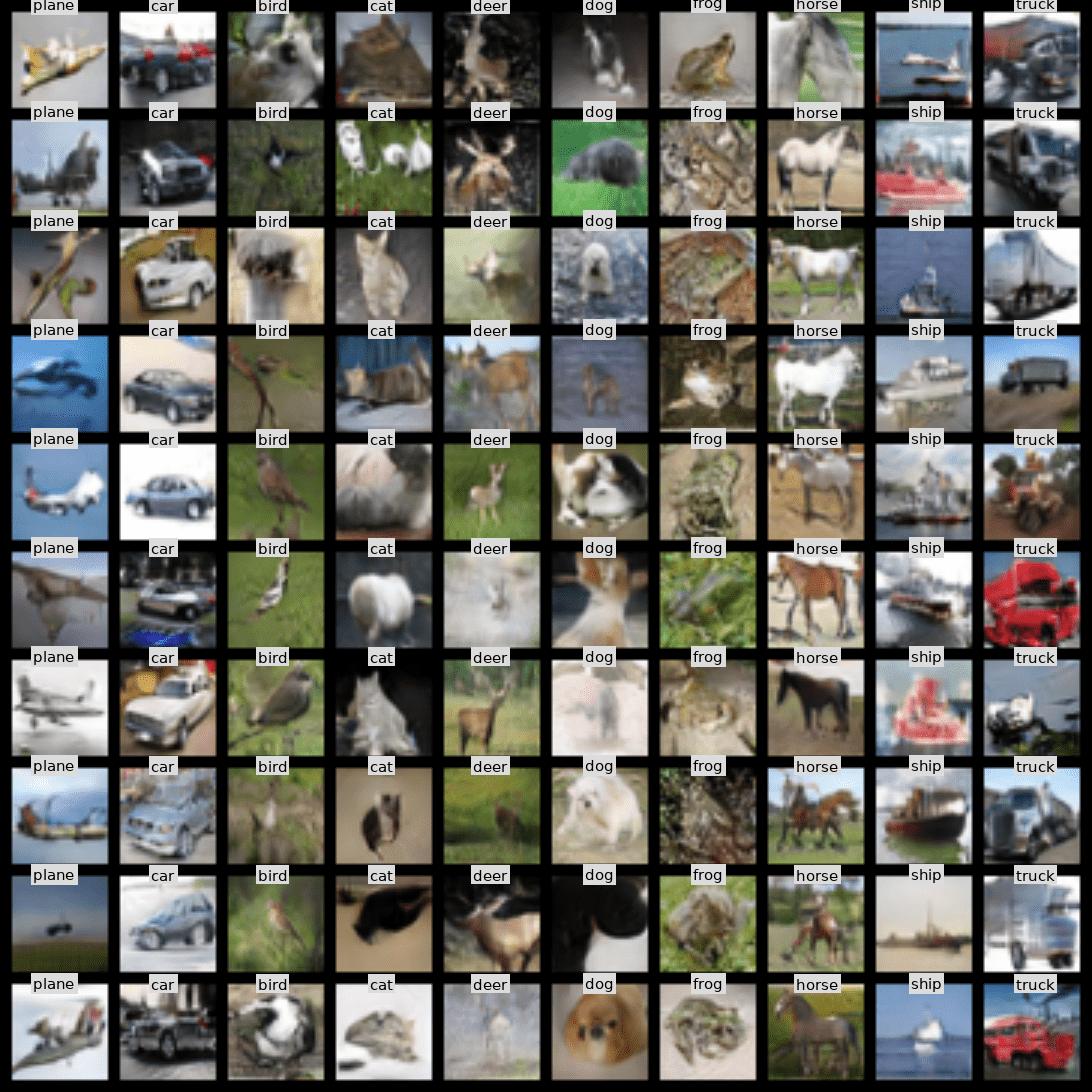}
    \caption{$T= 1.0$}
    \vspace{+5pt}
    \end{subfigure} \hfill
  \begin{subfigure}{0.48\textwidth}
    \includegraphics[width=\linewidth]{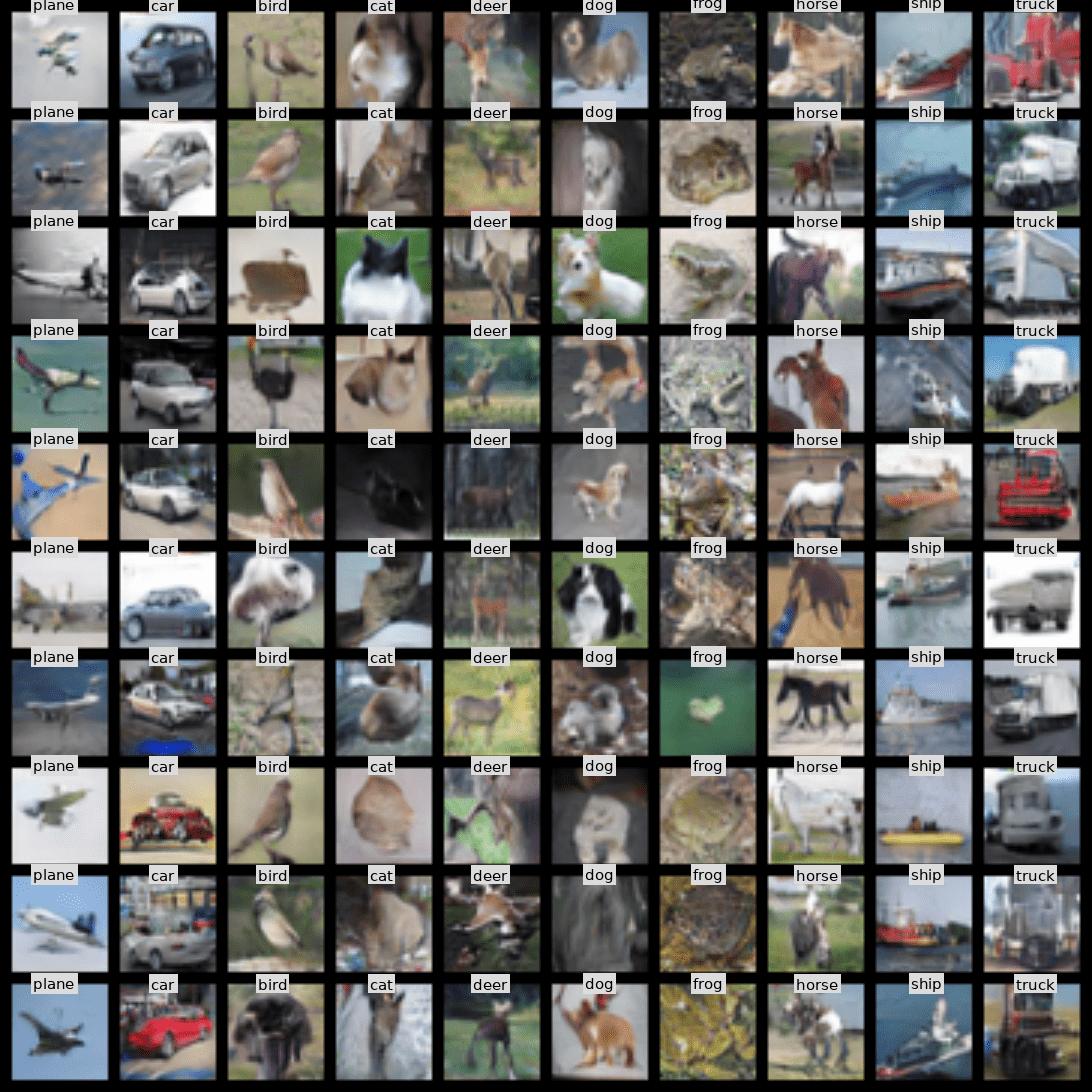}
    \caption{$T= 8.0$}
     \vspace{+5pt}
    \end{subfigure} \hfill
  \begin{subfigure}{0.48\textwidth}
    \includegraphics[width=\linewidth]{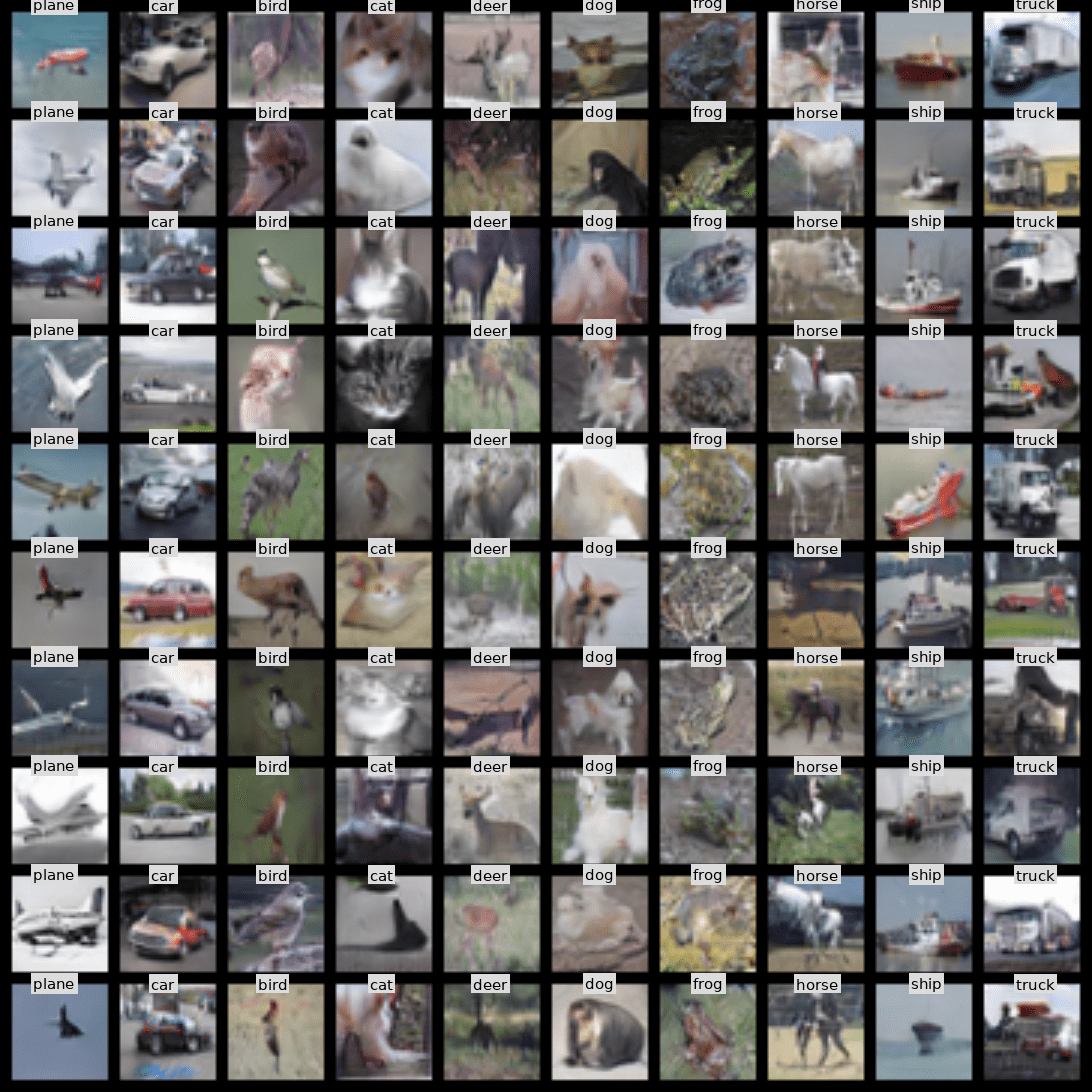}
    \caption{$T= 64.0$}
     \vspace{+5pt}
    \end{subfigure} \hfill
    \caption{D2M-generated image visualization (IPC10) for the CIFAR-10 dataset with different temperatures $(T)$ during distillation.}
    \label{fig:CIFAR10-Ablation-Temp}
\end{figure*}

\textbf{Visualization of the images generated with D2M trained on different temperatures $T$.}
This section examines the impact of temperature, denoted as $T$, on the generated synthetic images. As depicted in Fig. \textcolor{black}{4(a)} in the main paper, it becomes apparent that a moderate temperature value enhances the influence of logit matching in KL-Divergence, resulting in better-distilled images suitable for classification and downstream tasks. We illustrate this qualitative effect by comparing its impact on CIFAR10 in \Cref{fig:CIFAR10-Ablation-Temp}. We can clearly see that as we increase the temperature value from $0.5$ to $8$, there is a noticeable improvement in the localization and alignment of objects within the image, particularly within the 'car' and 'bus' classes. Meanwhile, when examining the image at $T=64$, it appears that the quality and robustness of the image features have slightly diminished in comparison to the previous images. Nevertheless, it still retains a richer set of feature information compared to $T=1$.

\begin{figure*}
    \centering
    \begin{subfigure}[b]{0.33\textwidth}
    \includegraphics[width=\linewidth]{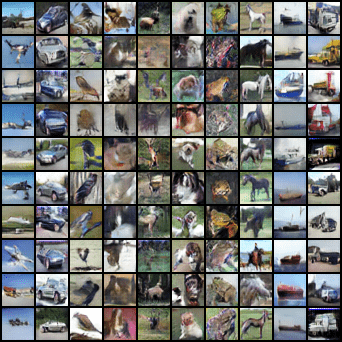}
            \caption{Conditioan GAN}
             \label{fig:sub1-cgan}
  \end{subfigure}\hfill
  \begin{subfigure}[b]{0.33\textwidth}
    \includegraphics[width=\linewidth]{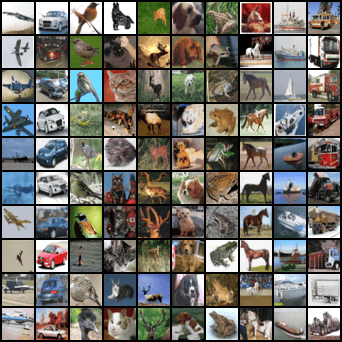}
        \caption{Style-GAN-XL}
    \label{fig:sub2-sgan}
  \end{subfigure}\hfill
    \begin{subfigure}[b]{0.33\textwidth}
    \includegraphics[width=\linewidth]{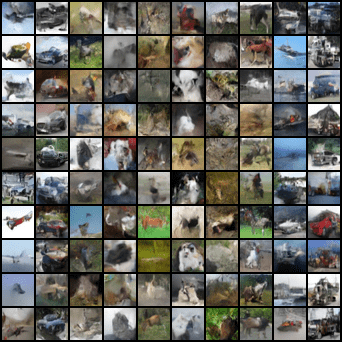}
    \caption{CVAE}
    \label{fig:sub3-vgan}
  \end{subfigure}\hfill
\caption{D2M-generated image visualization (IPC10) for the CIFAR-10 dataset with different generative models.}
    \label{fig:CIFAR10-Ablation-Generator}
\end{figure*}

\begin{figure*}
    \centering
    \begin{subfigure}[b]{0.49\textwidth}
    \includegraphics[width=\linewidth]{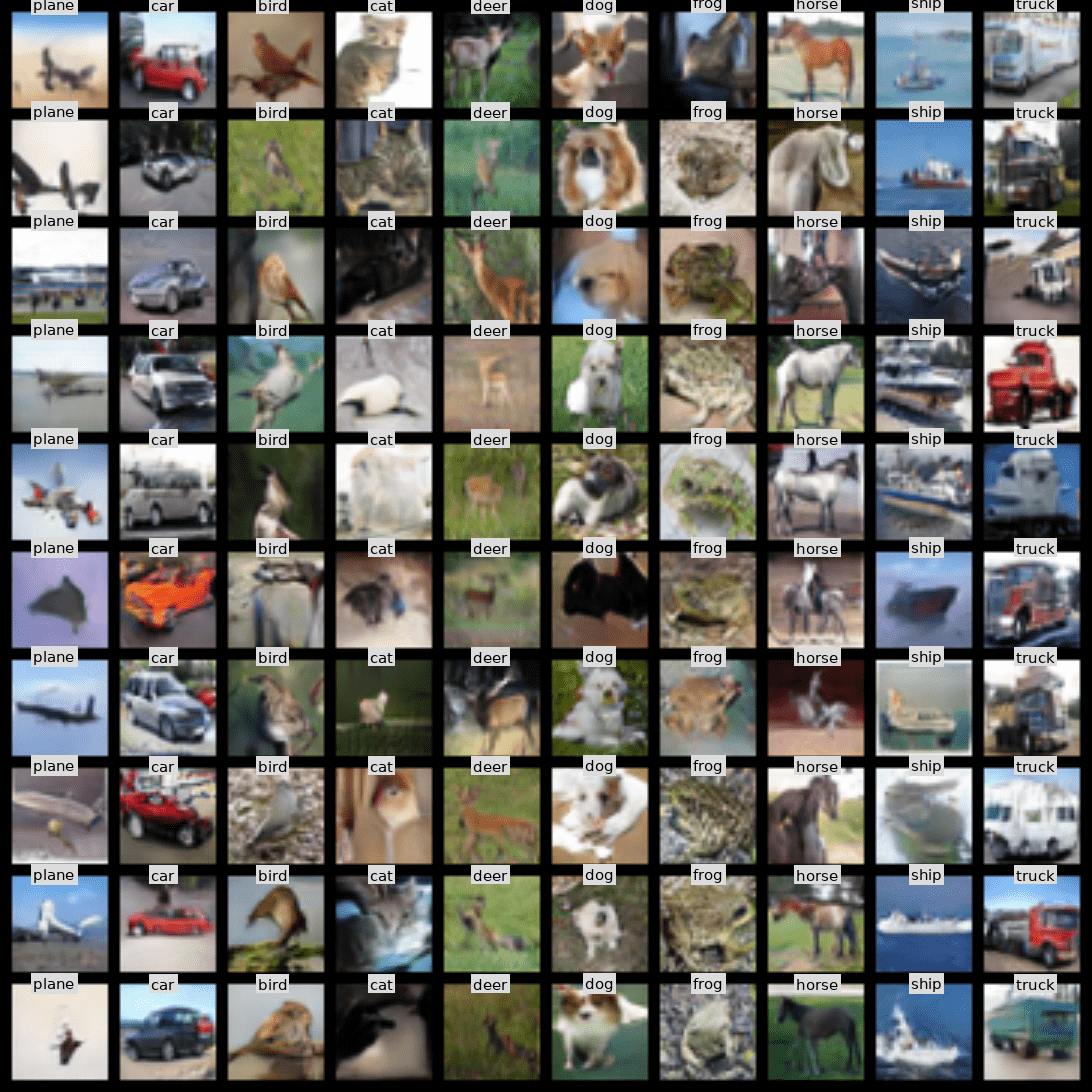}
    \caption{CIFAR-10 (IPC10)}
  \end{subfigure}\hfill
  \begin{subfigure}[b]{0.49\textwidth}
    \includegraphics[width=\linewidth]{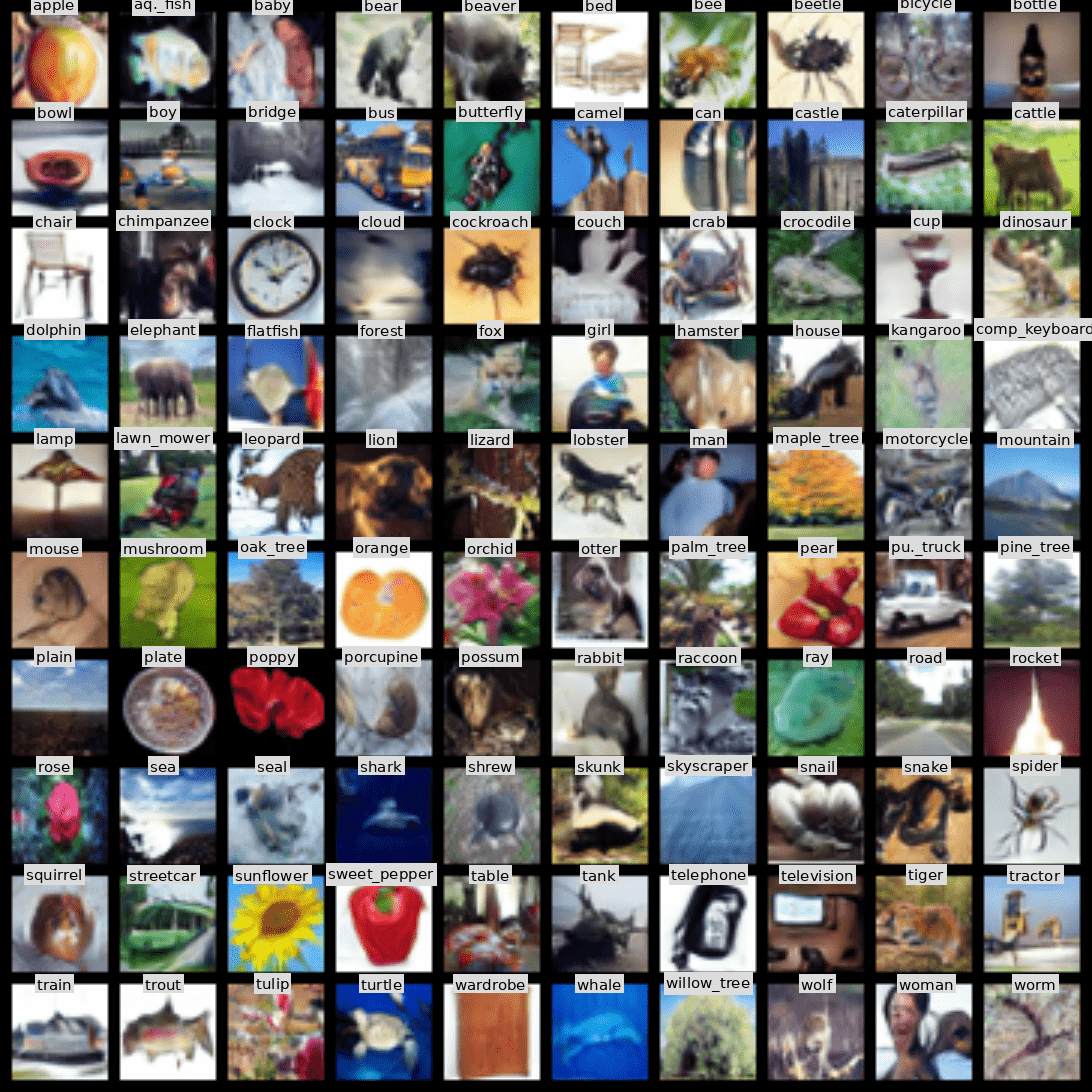}
    \caption{CIFAR-100 (IPC1)}
  \end{subfigure}\hfill
    \caption{D2M-generated image visualization: CIFAR datasets.}
    \label{fig:CIFAR10-100}
\end{figure*}

\textbf{Visualization of the images generated with D2M trained with different generative models.}
In this section, we present visual comparisons of different backbone generative models used to produce our synthetic dataset. Specifically, we evaluate conditional GAN (\Cref{fig:sub1-cgan}), StyleGAN-XL (\Cref{fig:sub2-sgan}), and CVAE (\Cref{fig:sub3-vgan}), in comparison to BigGAN (as depicted in \Cref{fig:CIFAR10-100} \textcolor{black}{(a)}). Upon qualitative analysis, we observe that the conditional GAN exhibits less distinct object localization compared to the other methods. This difference is particularly noticeable when examining the 'cat' and 'bird' classes. StyleGAN-XL stands out for producing high-quality images with rich colors and robust features across all categories. This is especially evident in the 'car,' 'airplane,' and 'ship' classes. Lastly, the CVAE also generates clear images; however, when compared to Style-GAN-XL, it tends to lose some relevant background information, notably in the 'airplane' class.

\begin{figure*}
    \centering
    \includegraphics[width=\textwidth]{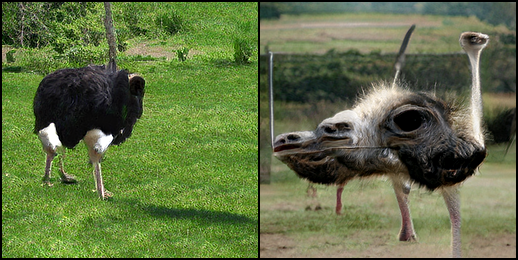}
    \includegraphics[width=\textwidth]{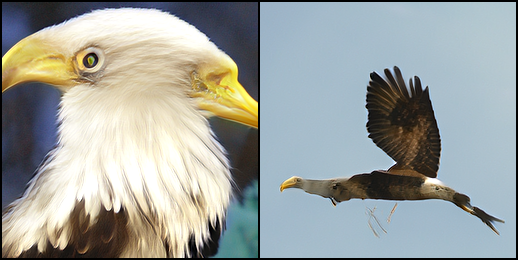}
    \caption{D2M-generated image visualization: ImageSquawk (256$\times$256): (Top) 'ostrich' class, (Bottom) 'eagle' class.}
    \label{fig:ImageSquackHR1}
\end{figure*}

\begin{figure*}
    \centering
    \includegraphics[width=\textwidth]{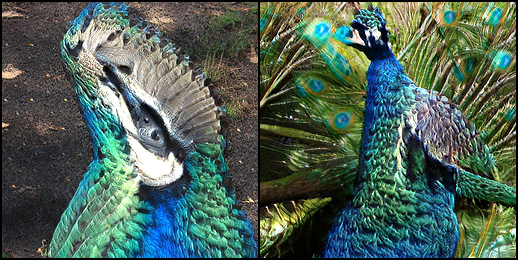}
    \includegraphics[width=\textwidth]{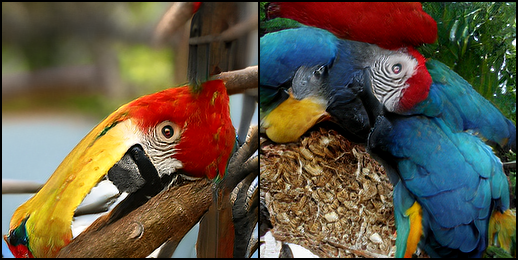}
    \caption{D2M-generated image visualization: ImageSquawk  (256$\times$256): (Top) 'peacock' class, (Bottom) 'macaw' class.}
    \label{fig:ImageSquackHR2}
\end{figure*}

\begin{figure*}
    \centering
    \includegraphics[width=\textwidth]{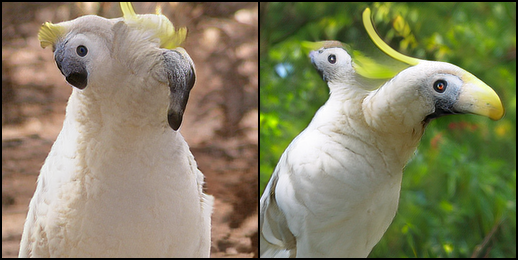}
    \includegraphics[width=\textwidth]{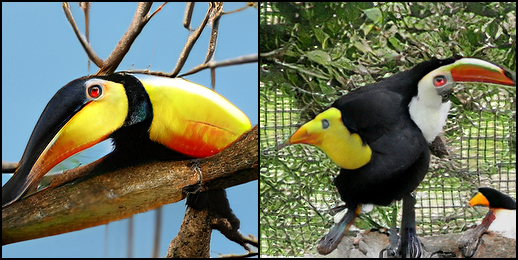}
    \caption{D2M-generated image visualization: ImageSquawk  (256$\times$256): (Top) 'cockatoo' class, (Bottom) 'toucan' class.}
    \label{fig:ImageSquackHR3}
\end{figure*}

\begin{figure*}
    \centering
    \includegraphics[width=\textwidth]{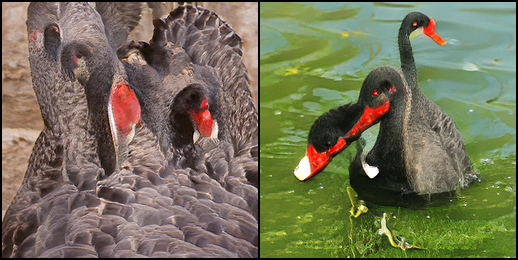}
    \includegraphics[width=\textwidth]{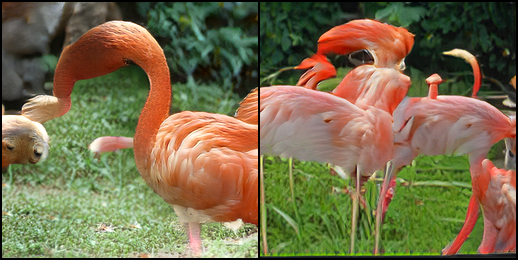}
    \caption{D2M-generated image visualization: ImageSquawk  (256$\times$256): (Top) 'black swan' class, (Bottom) 'flamingo' class.}
    \label{fig:ImageSquackHR4}
\end{figure*}

\begin{figure*}
    \centering
    \includegraphics[width=\textwidth]{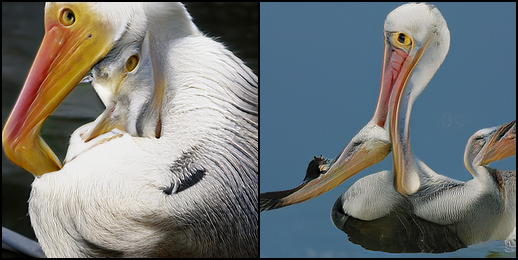}
    \includegraphics[width=\textwidth]{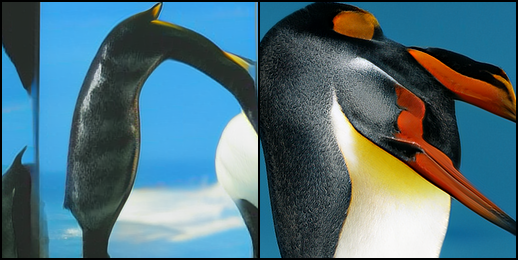}
    \caption{D2M-generated image visualization: ImageNet-1K (256$\times$256): (Top) 'pelican' class, (Bottom) 'penguin' class}
    \label{fig:ImageSquackHR5}
\end{figure*}

\begin{figure*}
    \centering
    \includegraphics[width=\textwidth]{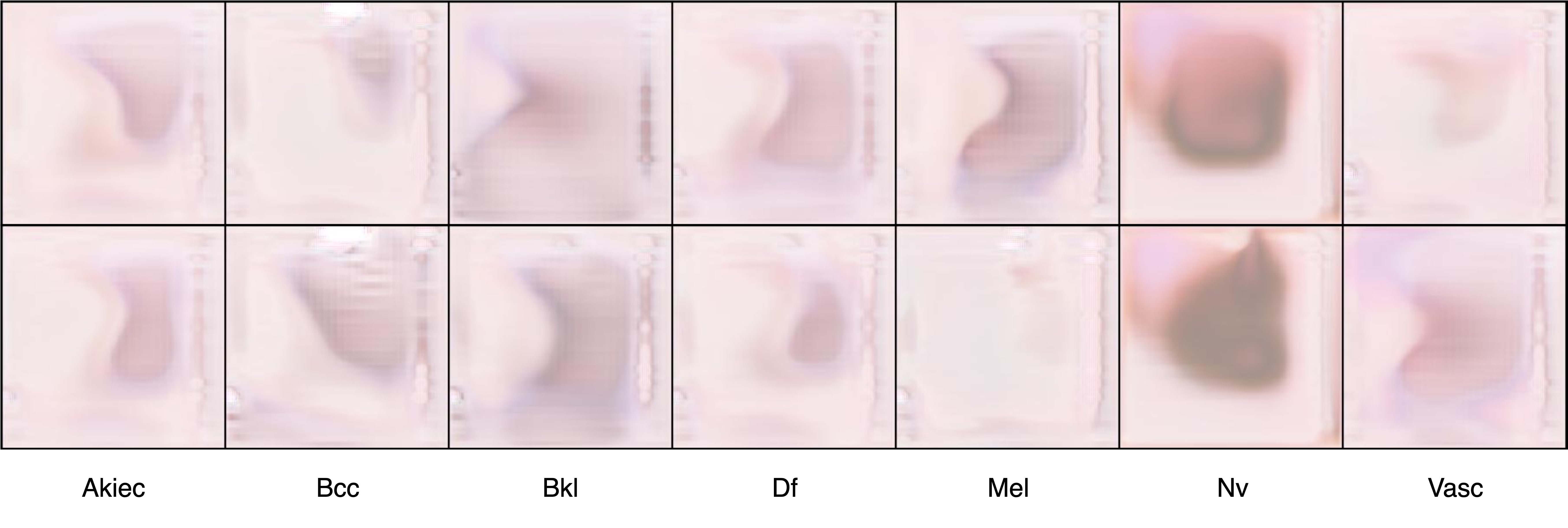}
    \caption{D2M-generated image visualization: DermaMNIST (256$\times$256) 7 classes. Class names under respective columns.}
    \label{fig:derma}
\end{figure*}

\begin{figure*}
    \centering
    \includegraphics[width=\textwidth]{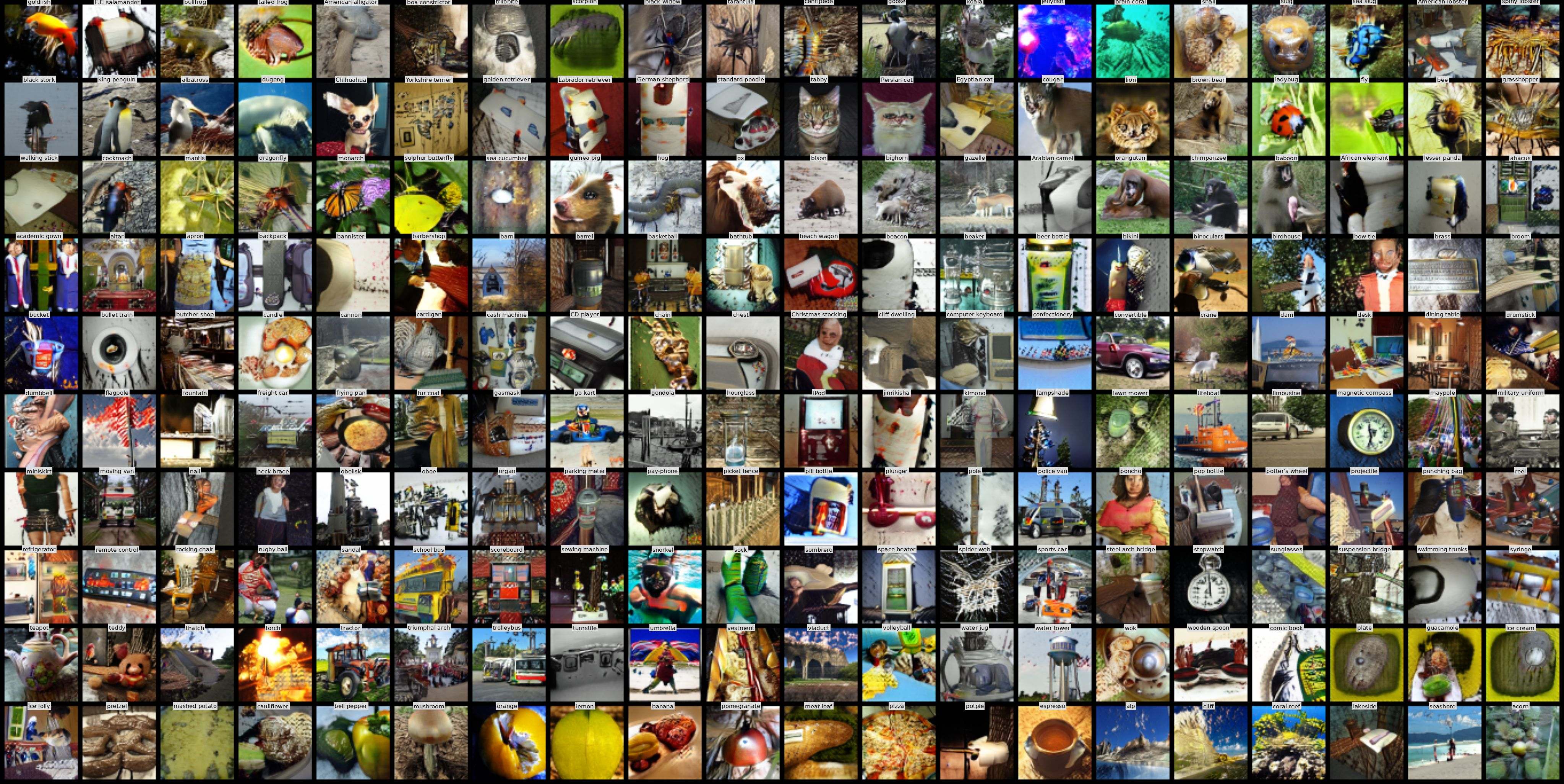}
    \caption{D2M-generated image visualization: Tiny ImageNet (64$\times$64) with IPC1.}
    \label{fig:TinyImagent}
\end{figure*}

\begin{figure*}
    \centering
    \includegraphics[width=\textwidth]{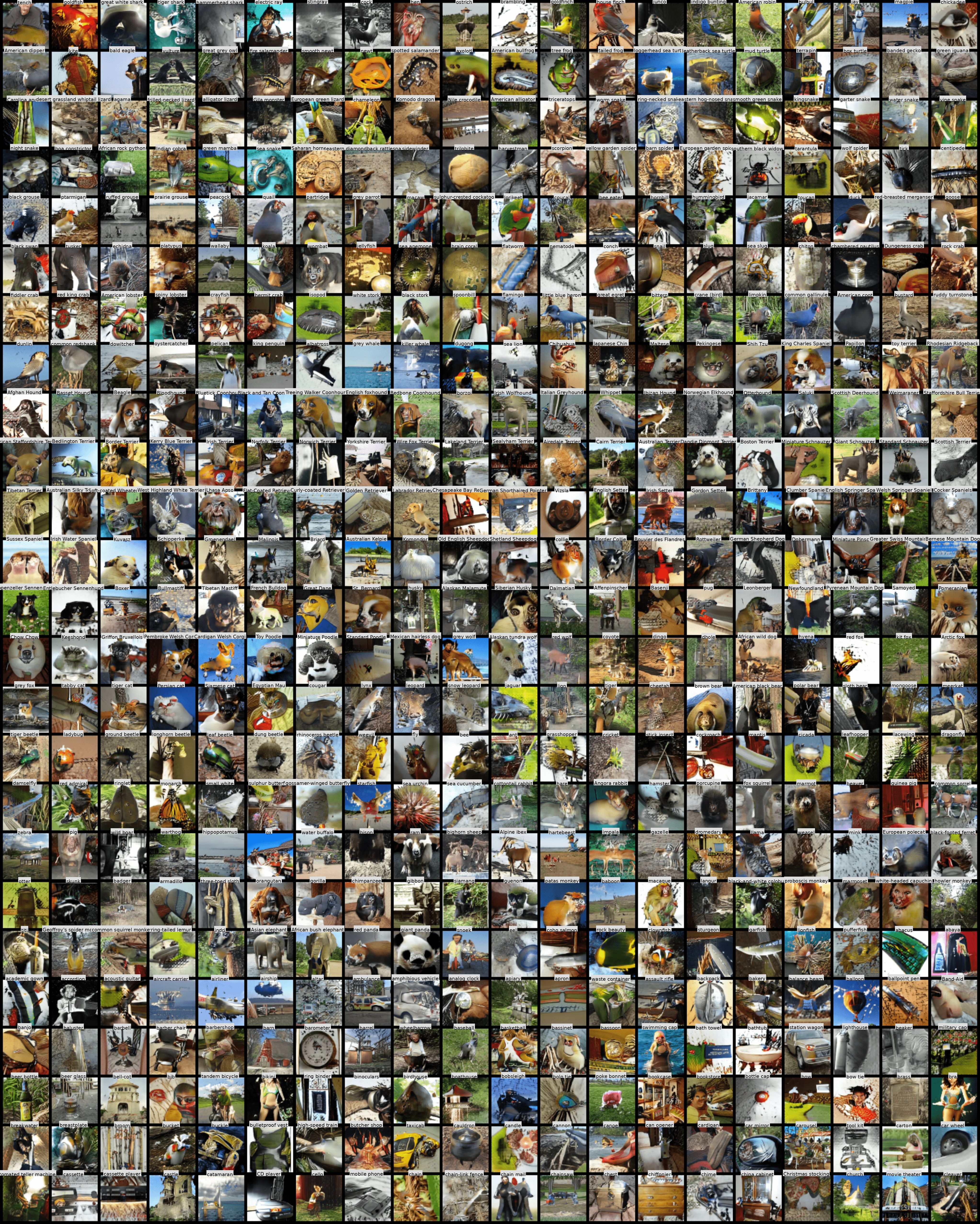}
    \caption{D2M-generated image visualization: ImageNet-1K (64$\times$64) Classes 0-500.}
    \label{fig:ImageNet64a}
\end{figure*}

\begin{figure*}
    \centering
    \includegraphics[width=\textwidth]{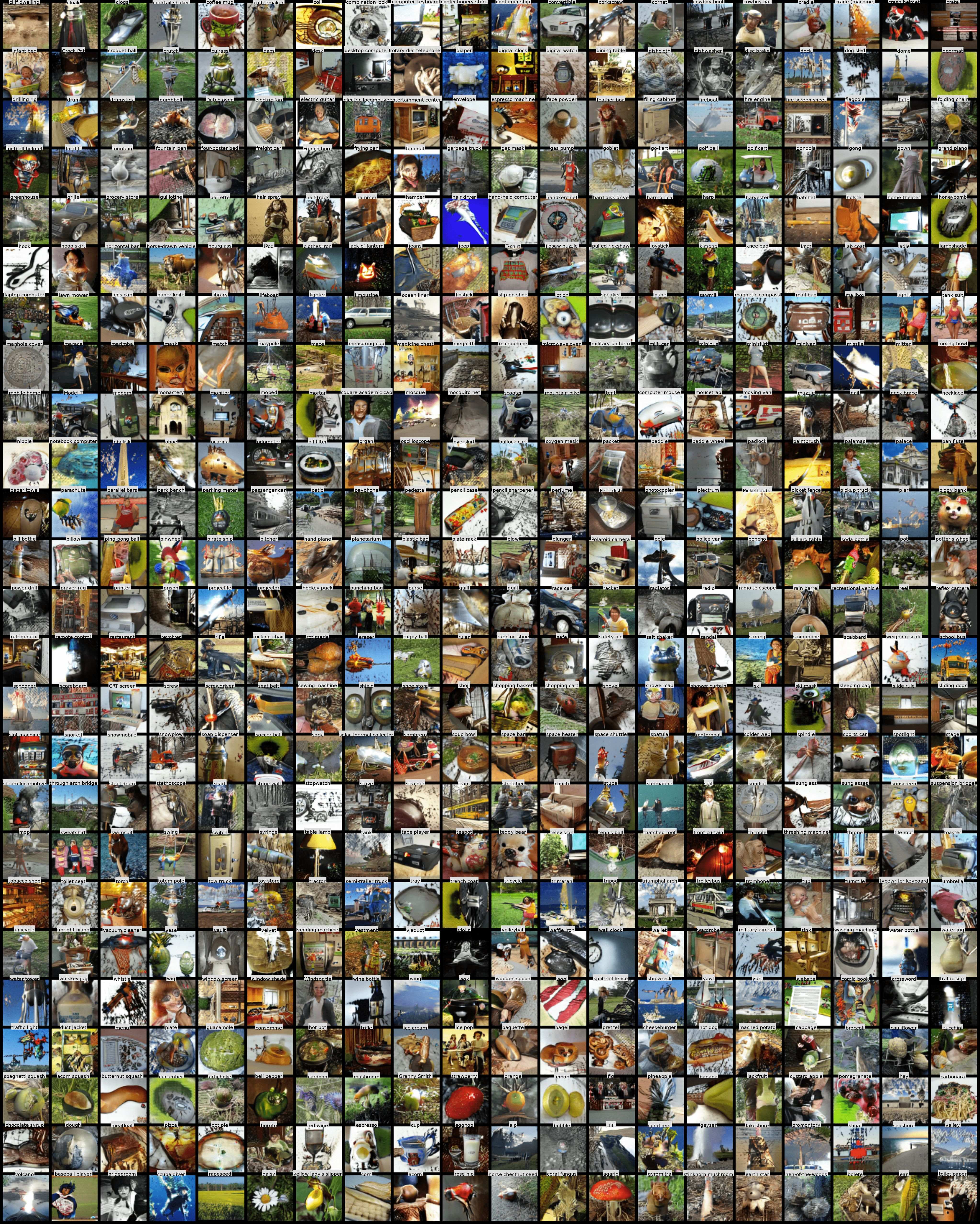}
    \caption{D2M-generated image visualization: ImageNet-1K (64$\times$64) Classes 500-1000.}
    \label{fig:ImageNet64b}
\end{figure*}

\begin{figure*}
    \centering
    \includegraphics[width=0.64\textwidth]{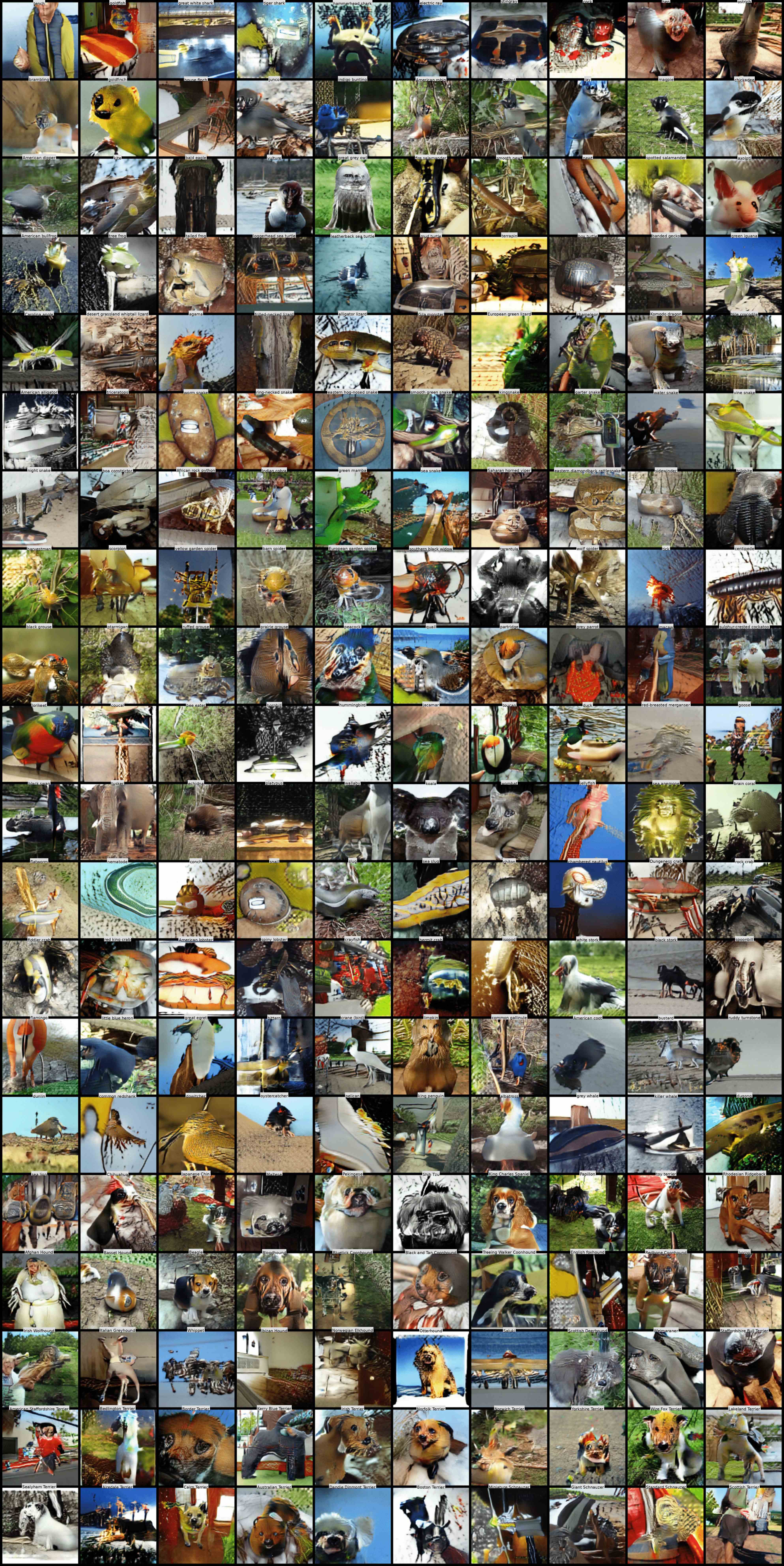}
    \caption{D2M-generated image visualization: ImageNet-1K (128$\times$128) Classes 0-200.}
    \label{fig:ImageNet128-PT1}
\end{figure*}

\begin{figure*}
    \centering
    \includegraphics[width=0.64\textwidth]{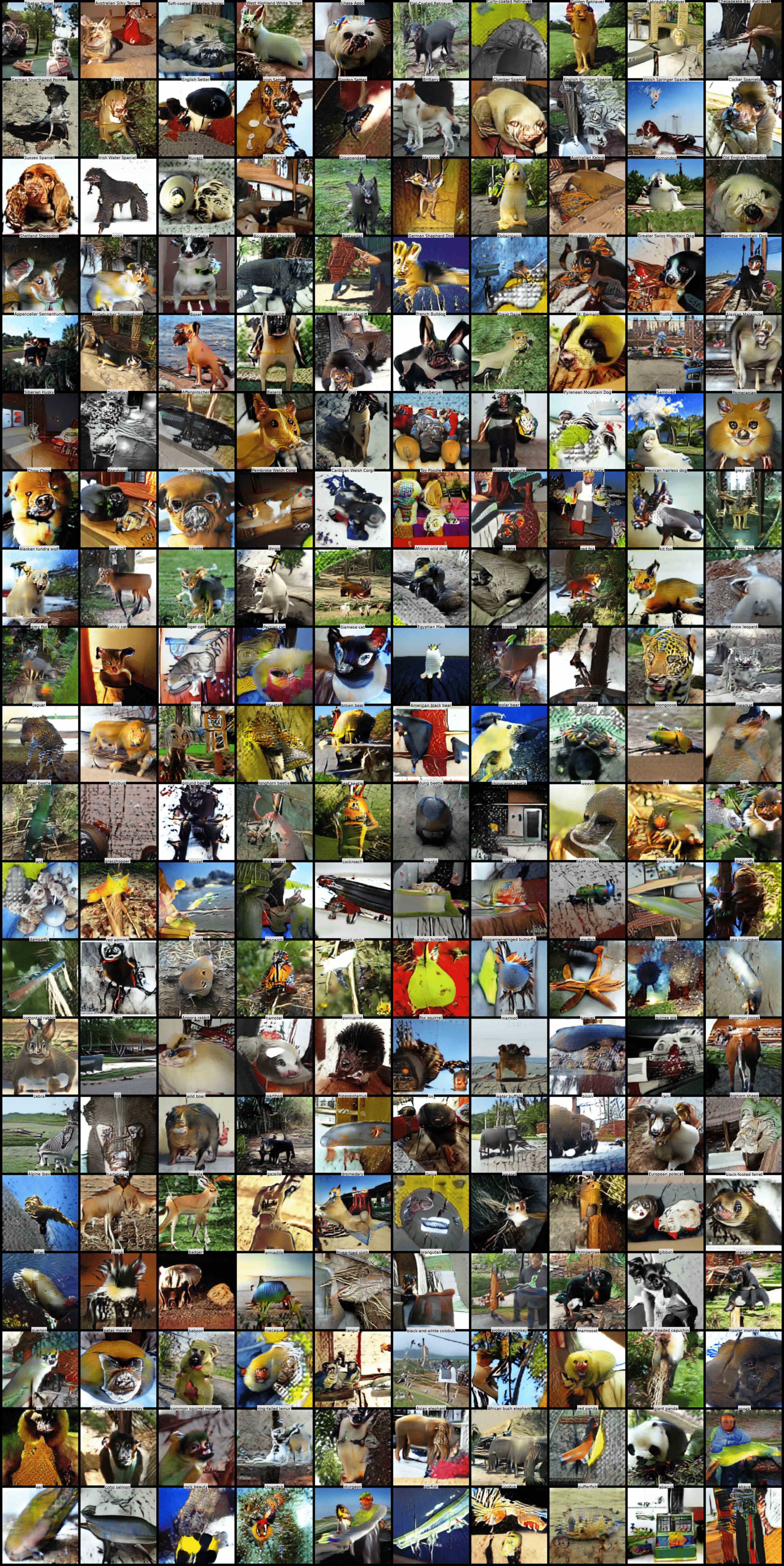}
    \caption{D2M-generated image visualization: ImageNet-1K (128$\times$128) Classes 200-400.}
    \label{fig:ImageNet128-PT2}
\end{figure*}

\begin{figure*}
    \centering
    \includegraphics[width=0.64\textwidth]{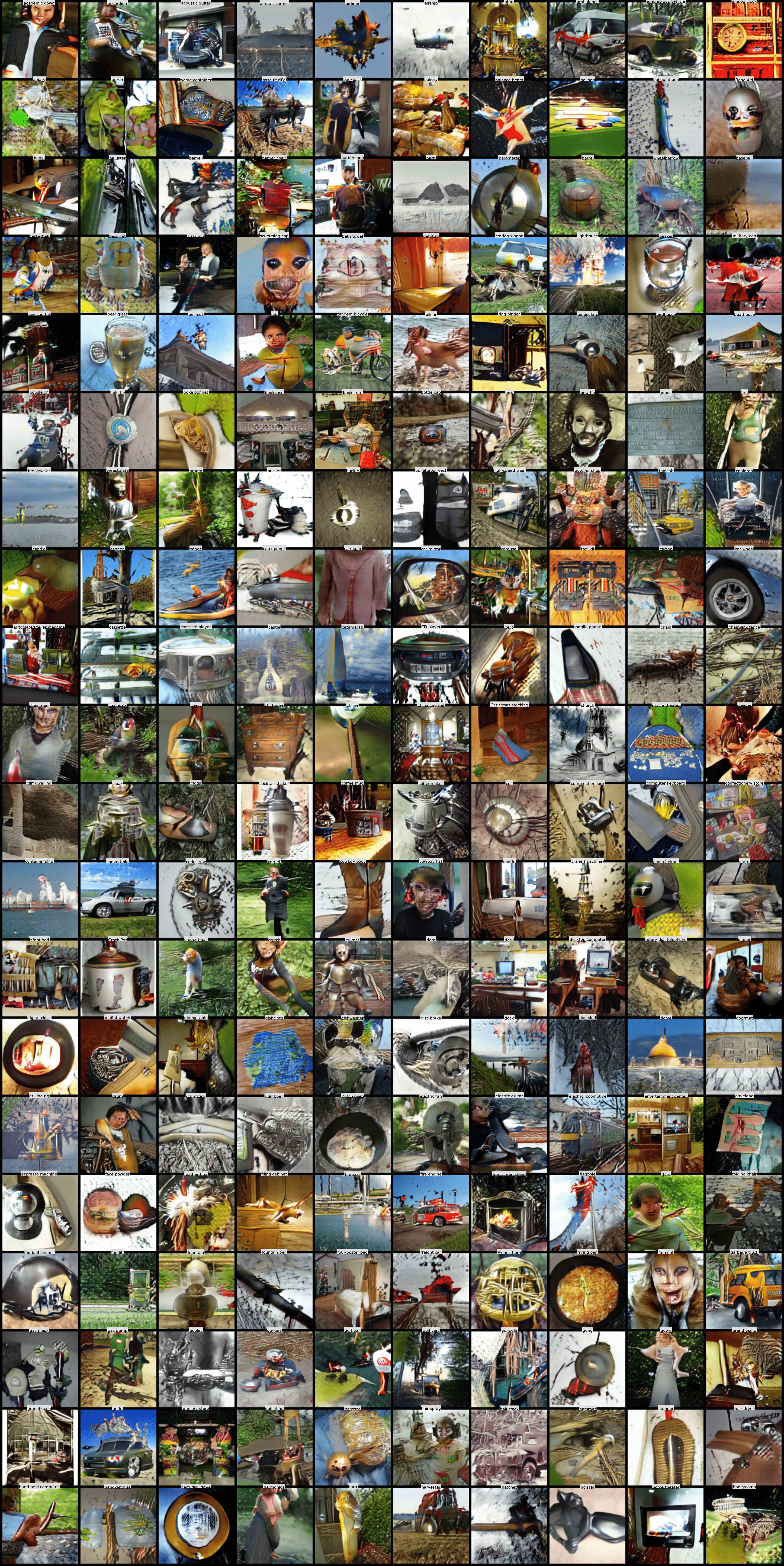}
    \caption{D2M-generated image visualization: ImageNet-1K (128$\times$128) Classes 400-600.}
    \label{fig:ImageNet128-PT3}
\end{figure*}

\begin{figure*}
    \centering
    \includegraphics[width=0.64\textwidth]{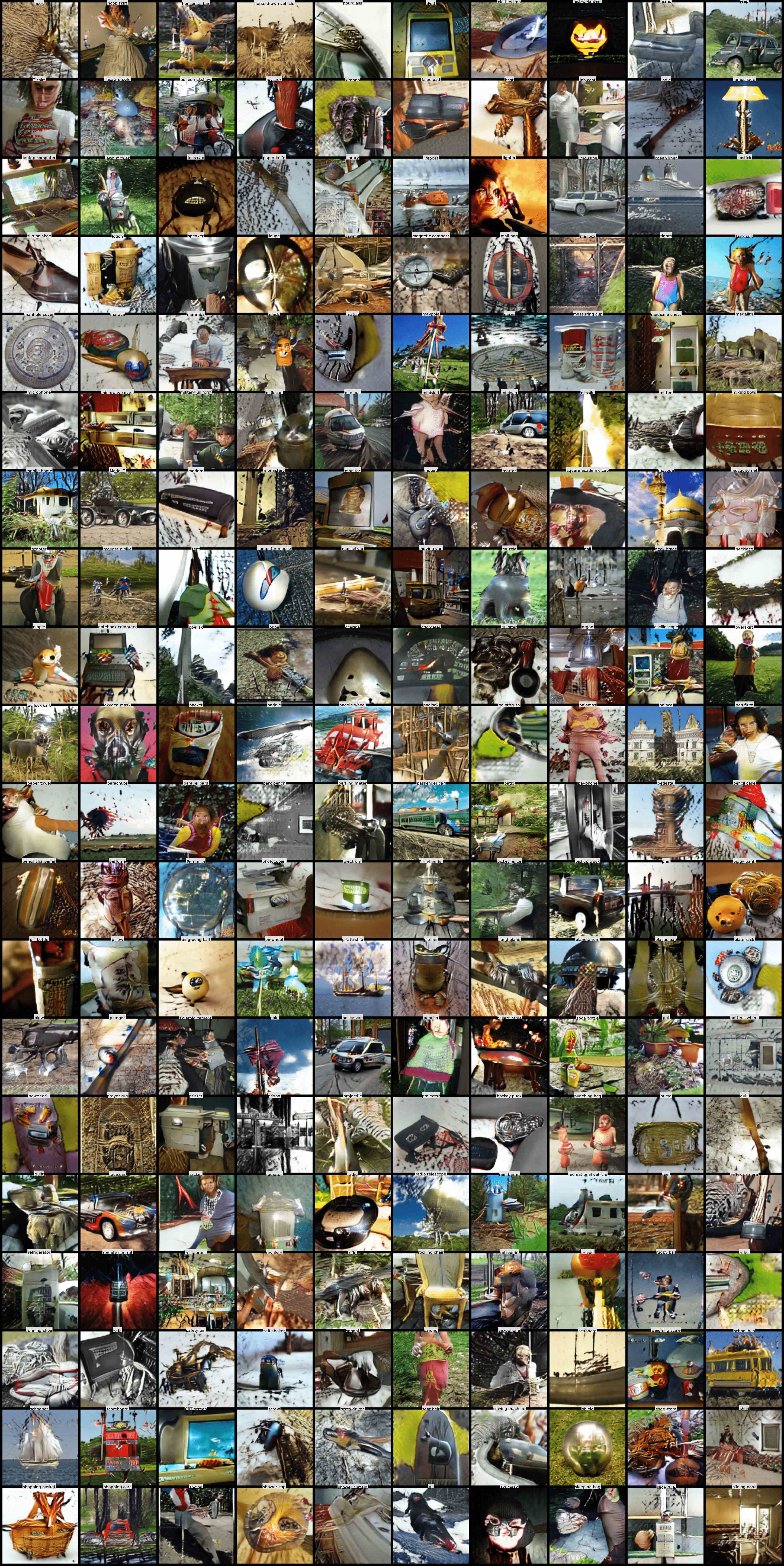}
    \caption{D2M-generated image visualization: ImageNet-1K (128$\times$128) Classes 600-800.}
    \label{fig:ImageNet128-PT4}
\end{figure*}

\begin{figure*}
    \centering
    \includegraphics[width=0.64\textwidth]{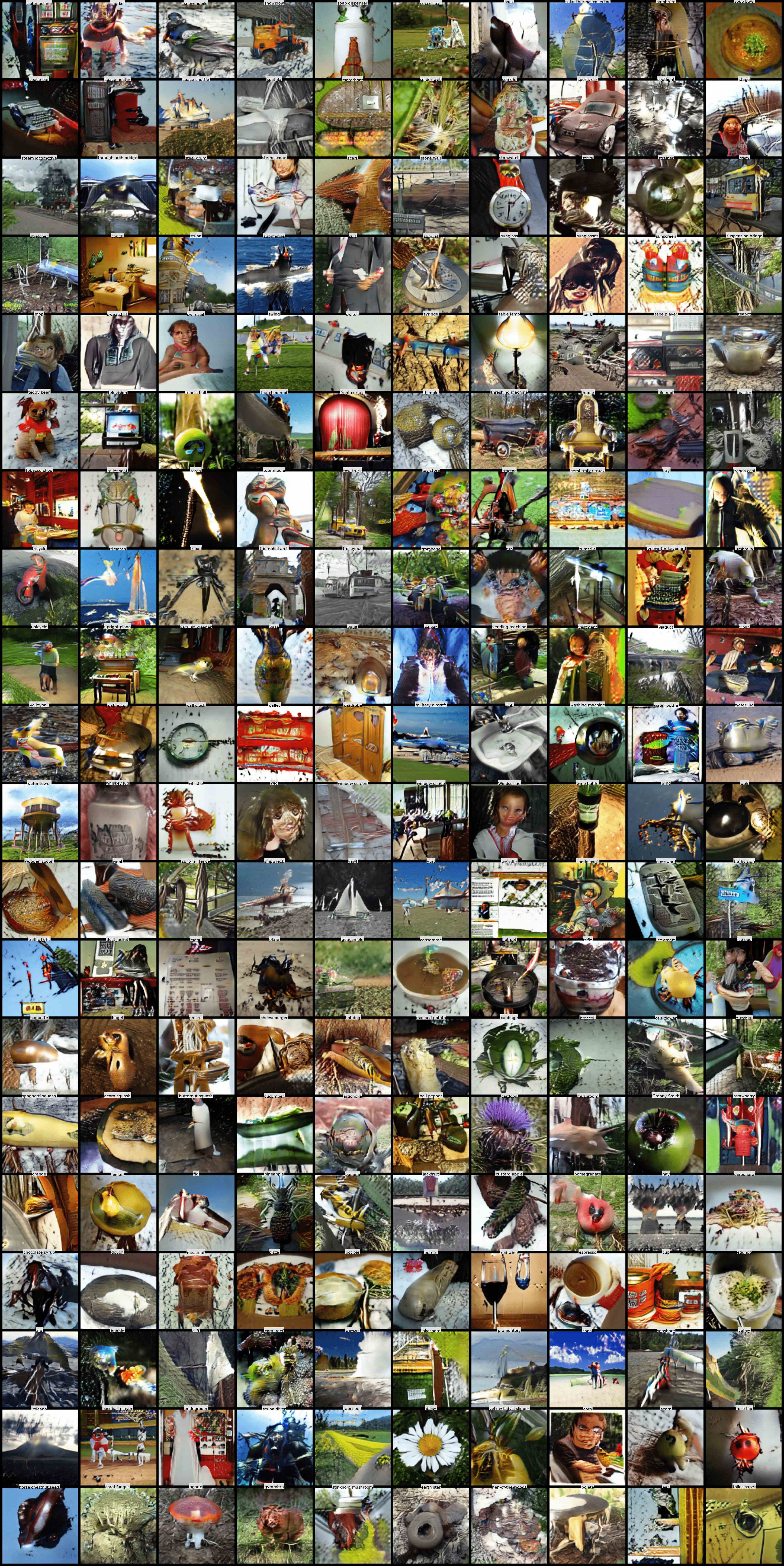}
    \caption{D2M-generated image visualization: ImageNet-1K (128$\times$128) Classes 800-1000.}
    \label{fig:ImageNet128-PT5}
\end{figure*}

\begin{figure*}
    \centering
    \includegraphics[width=\textwidth]{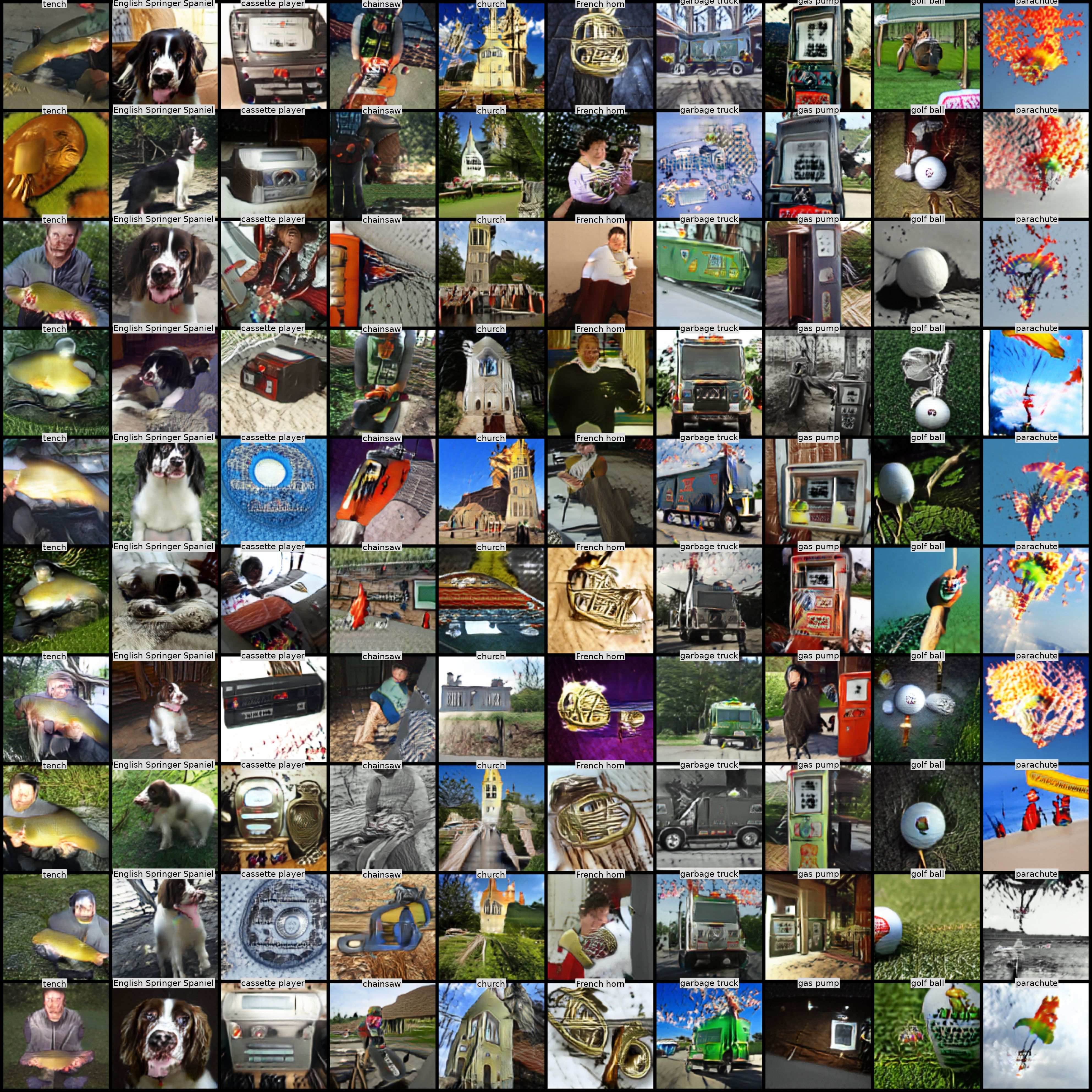}
    \caption{D2M-generated image visualization: ImageNette Subset (128$\times$128) with IPC1.}
    \label{fig:ImageNette}
\end{figure*}

\begin{figure*}
    \centering
    \includegraphics[width=\textwidth]{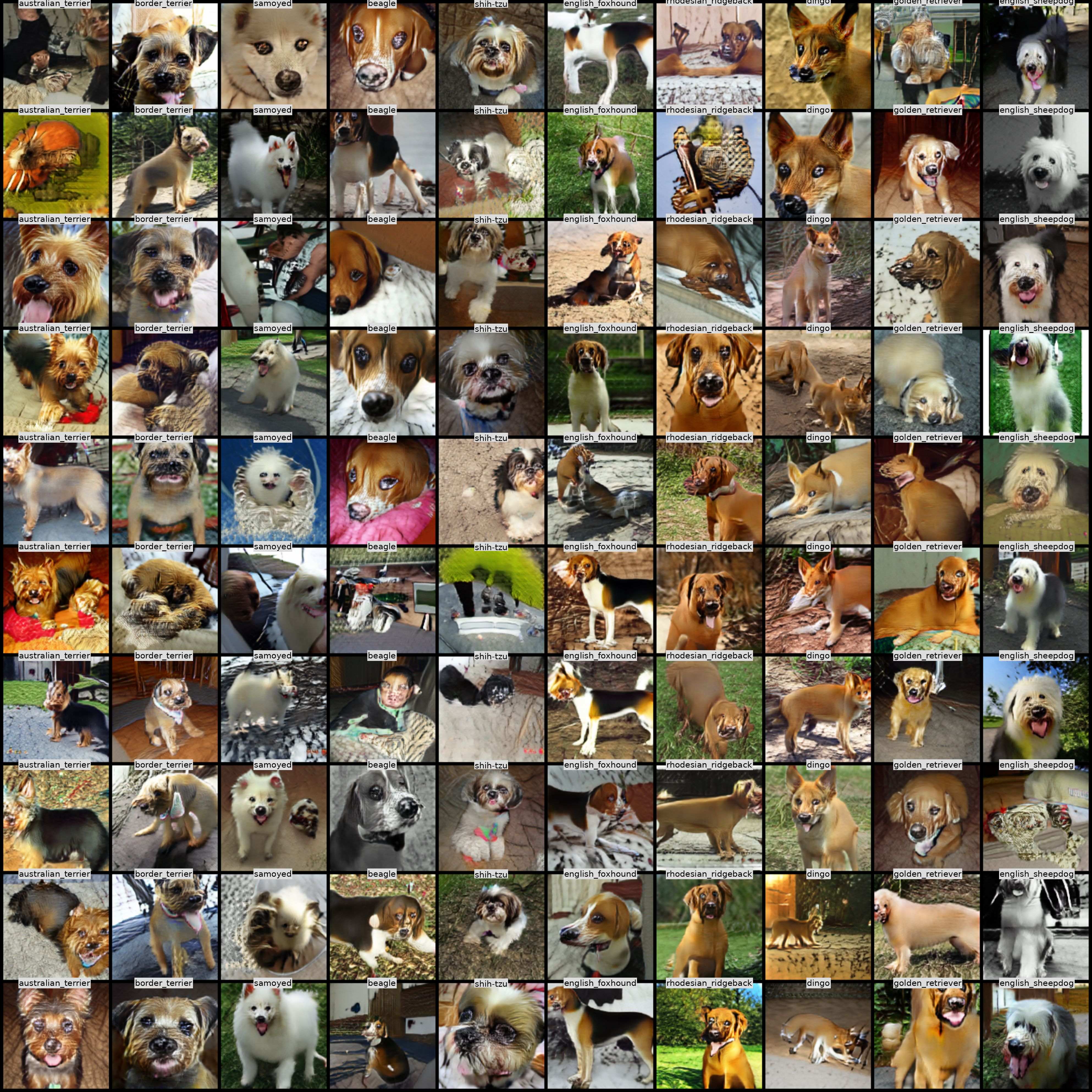}
    \caption{D2M-generated image visualization: ImageWoof Subset (128$\times$128) with IPC1.}
    \label{fig:ImageWoof}
\end{figure*}

\begin{figure*}
    \centering
    \includegraphics[width=\textwidth]{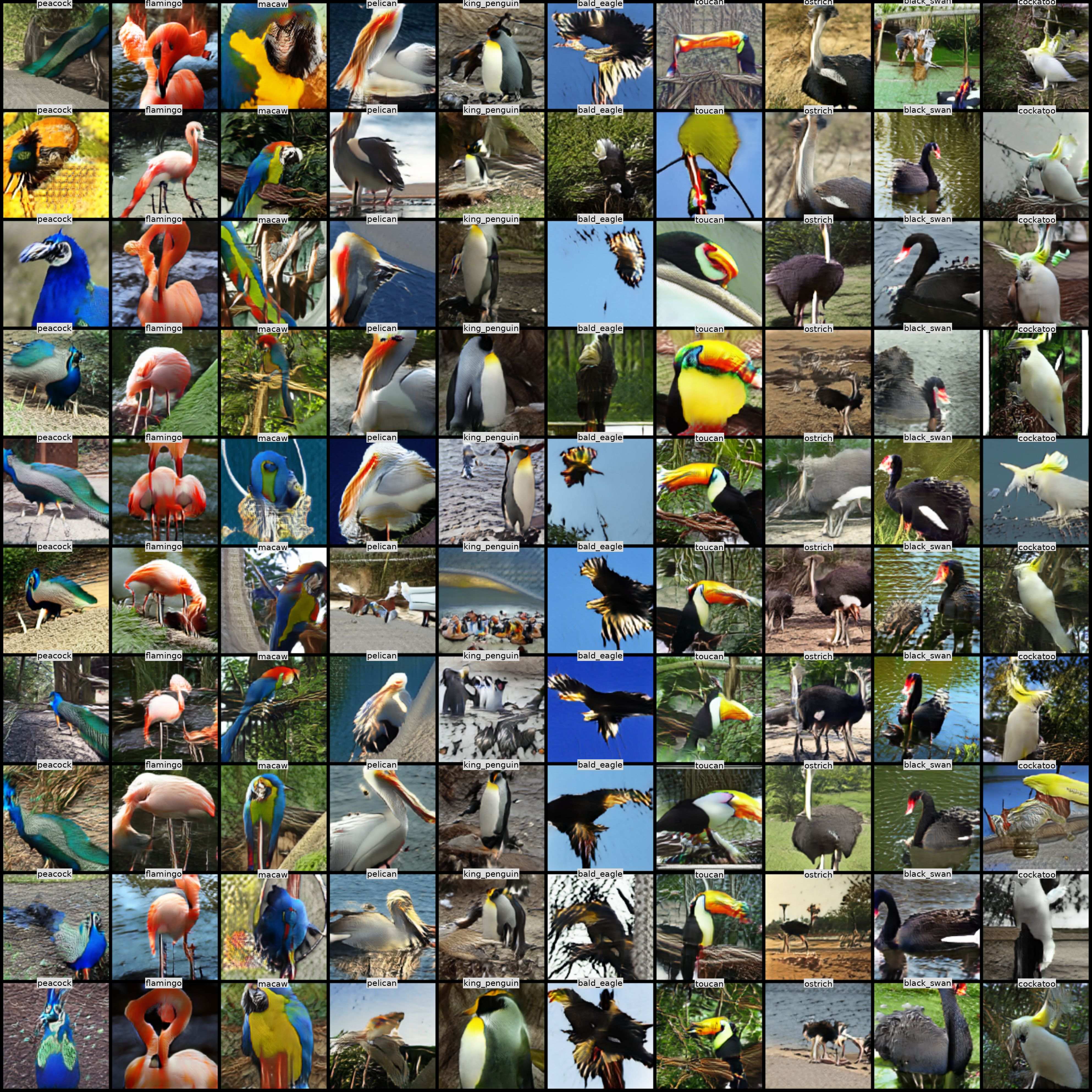}
    \caption{D2M-generated image visualization: ImageSquawk Subset (128$\times$128) with IPC1.}
    \label{fig:ImageSquack}
\end{figure*}

\begin{figure*}
    \centering
    \includegraphics[width=\textwidth]{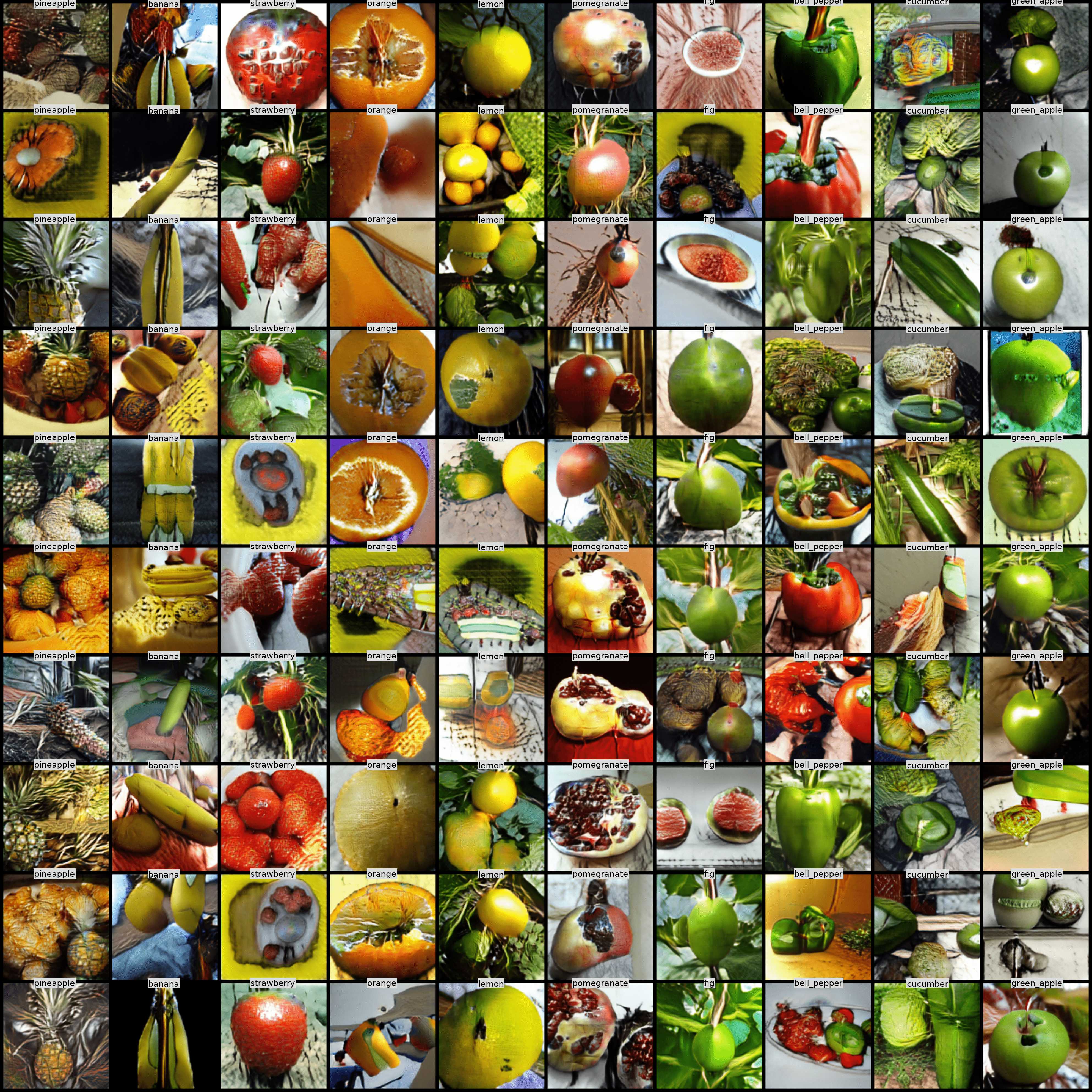}
    \caption{D2M-generated image visualization: ImageFruit Subset (128$\times$128) with IPC1.}
    \label{fig:ImageFruit}
\end{figure*}

\begin{figure*}
    \centering
    \includegraphics[width=\textwidth]{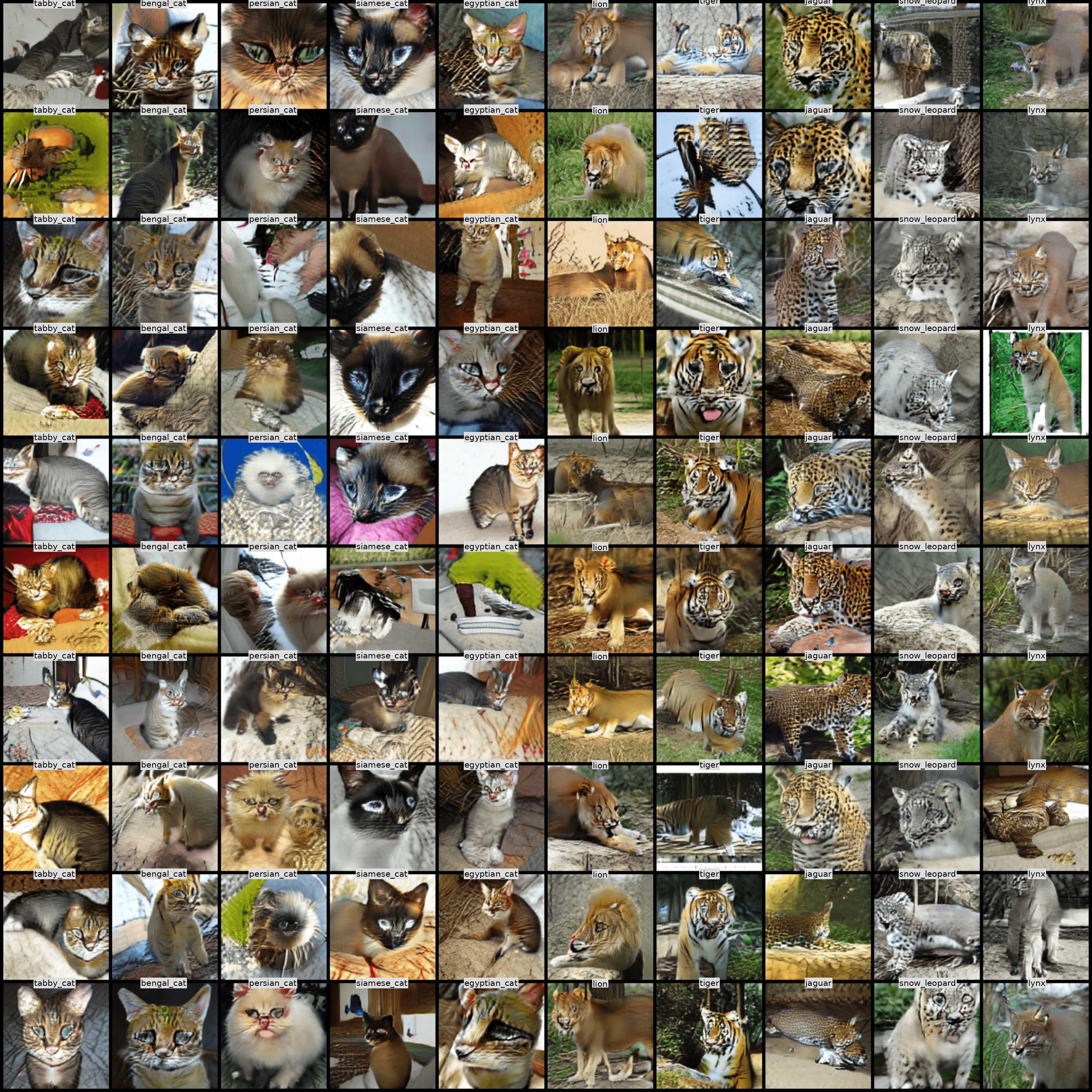}
    \caption{D2M-generated image visualization: ImageMeow Subset (128$\times$128) with IPC1.}
    \label{fig:ImageMeow}
\end{figure*}

\begin{figure*}
    \centering
    \includegraphics[width=\textwidth]{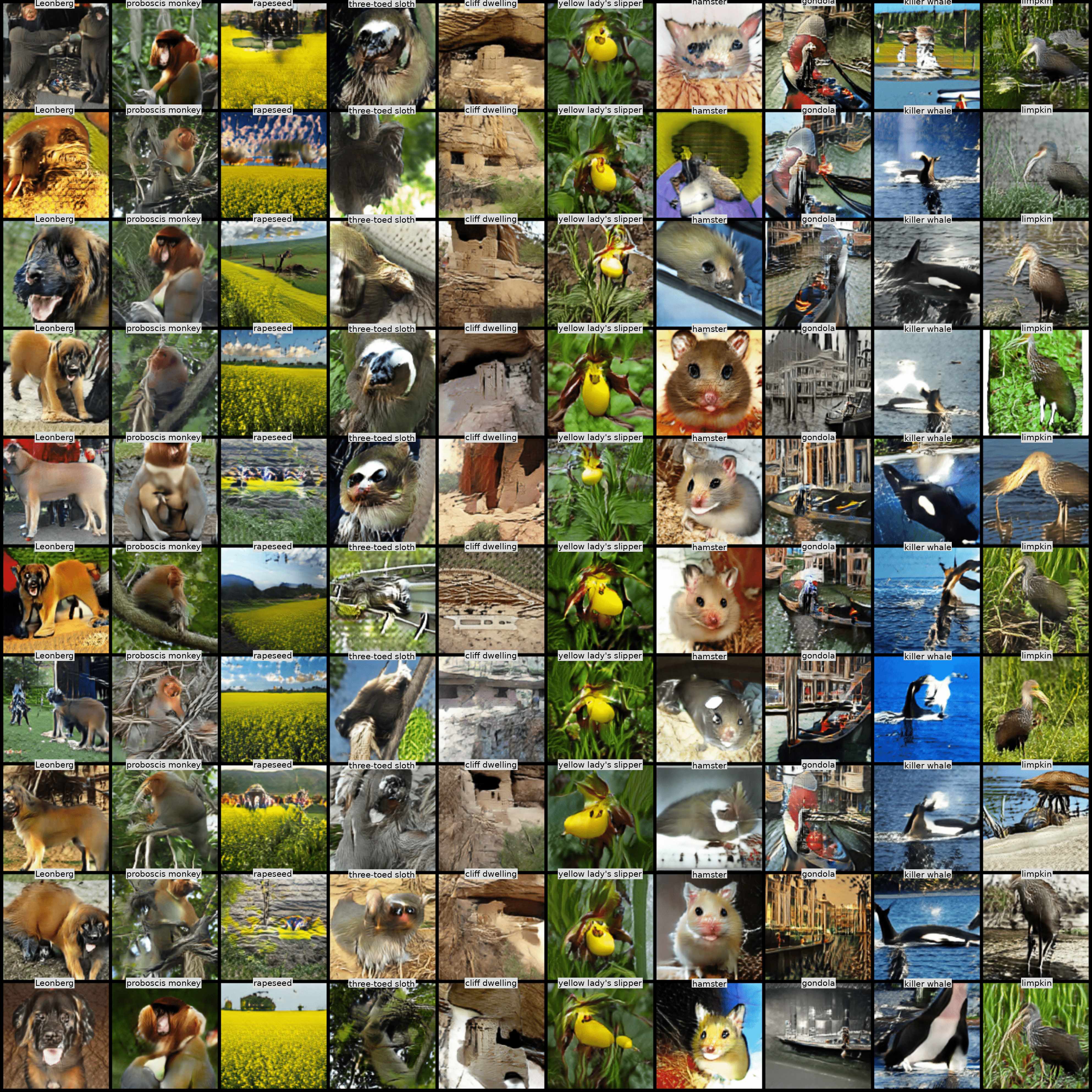}
    \caption{D2M-generated image visualization: ImageNet-A Subset (128$\times$128) with IPC1.}
    \label{fig:ImageNet-A}
\end{figure*}

\begin{figure*}
    \centering
    \includegraphics[width=\textwidth]{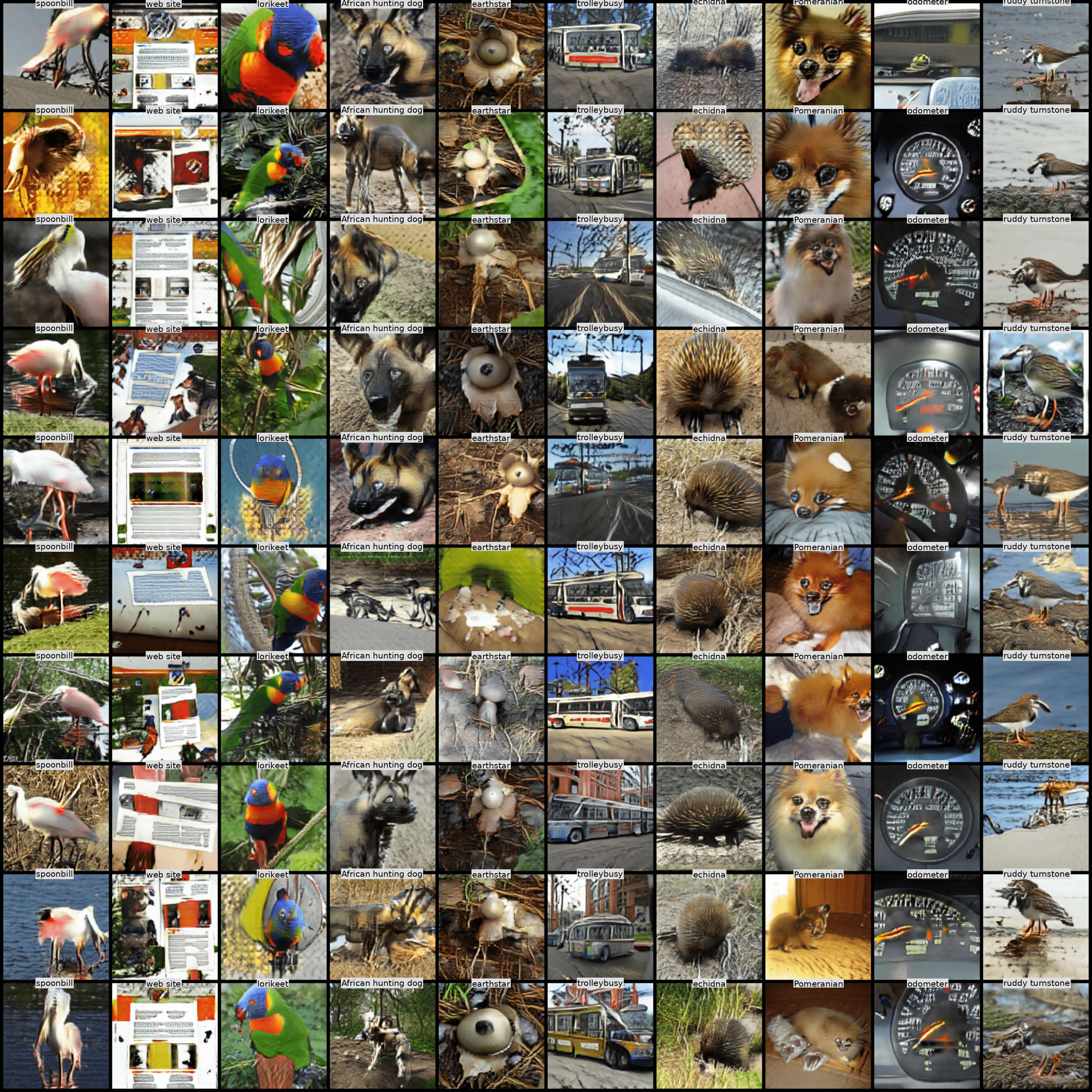}
    \caption{D2M-generated image visualization: ImageNet-B Subset (128$\times$128) with IPC1.}
    \label{fig:ImageNet-B}
\end{figure*}

\begin{figure*}
    \centering
    \includegraphics[width=\textwidth]{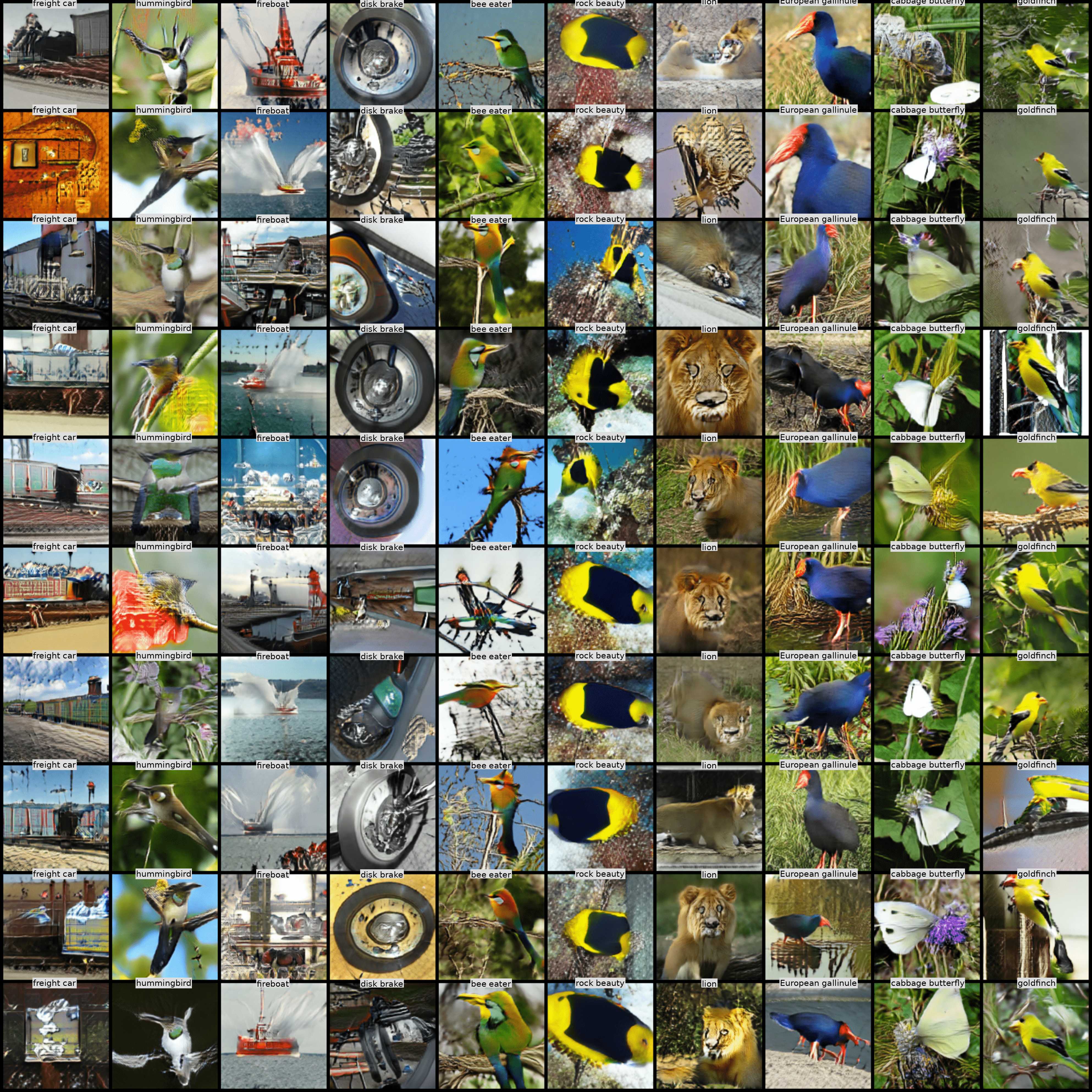}
    \caption{D2M-generated image visualization: ImageNet-C Subset (128$\times$128) with IPC1.}
    \label{fig:ImageNet-C}
\end{figure*}

\begin{figure*}
    \centering
    \includegraphics[width=\textwidth]{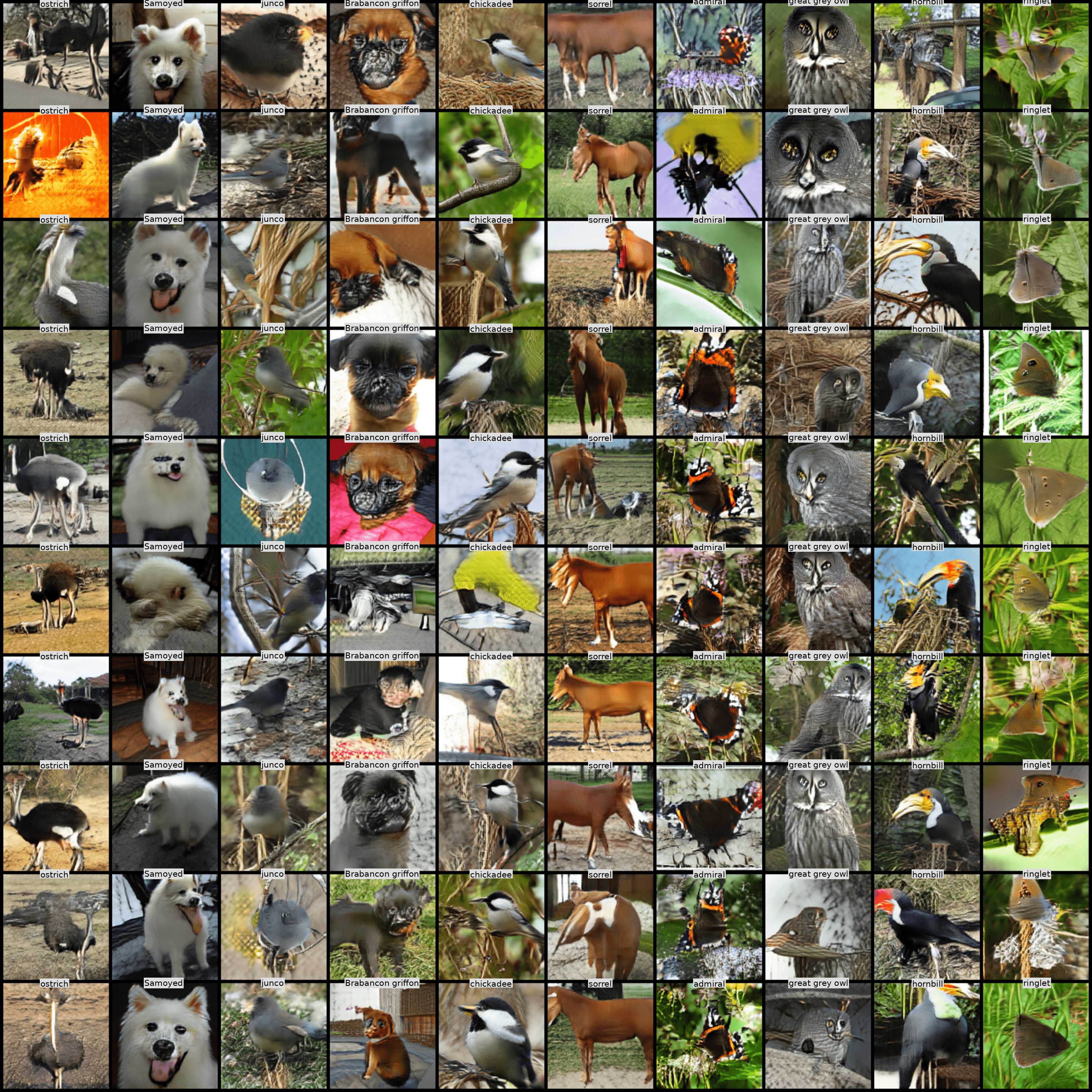}
    \caption{D2M-generated image visualization: ImageNet-D Subset (128$\times$128) with IPC1.}
    \label{fig:ImageNet-D}
\end{figure*}

\begin{figure*}
    \centering
    \includegraphics[width=\textwidth]{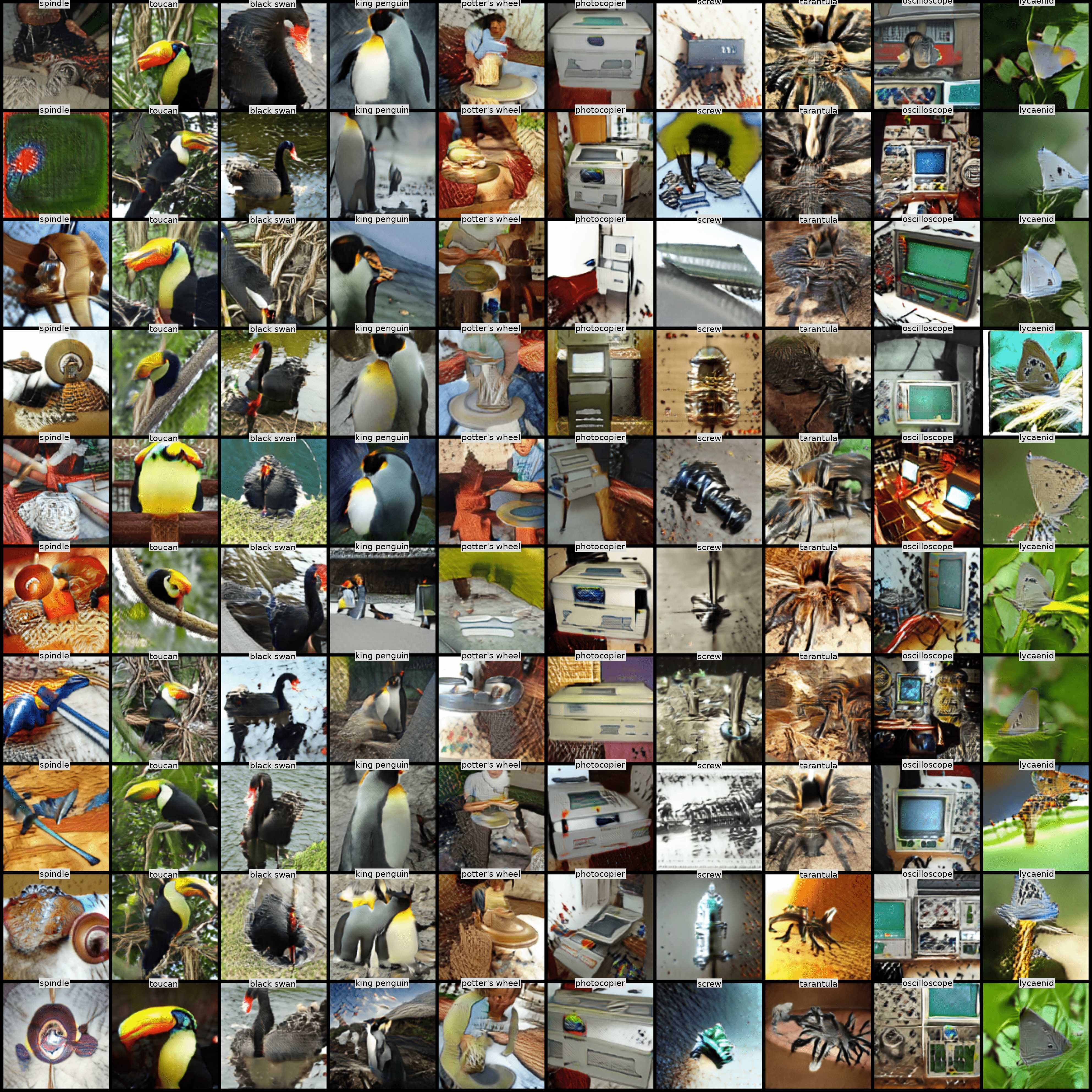}
    \caption{D2M-generated image visualization: ImageNet-E Subset (128$\times$128) with IPC1.}
    \label{fig:ImageNet-E}
\end{figure*}

\textbf{Extended Visualizations.}
We provide additional visualizations of the generated images for CIFAR-10 \& CIFAR-100 in \Cref{fig:CIFAR10-100}, Tiny ImageNet \& ImageNet-1K at resolution (64$\times$64) in \Cref{fig:TinyImagent} and \Cref{fig:ImageNet64a}-\Cref{fig:ImageNet64b}, respectively. In the collection of figures presented in \Cref{fig:ImageSquackHR1}, \Cref{fig:ImageSquackHR2}, \Cref{fig:ImageSquackHR3}, \Cref{fig:ImageSquackHR4}, and \Cref{fig:ImageSquackHR5}, we visually show how the higher resolution of 256$\times$256 pixels enhances object clarity and feature richness while preserving the properties of our matching algorithm. Notable improvements can be observed, such as the enhanced details of multiple heads in the 'eagle' class and the two-facing beaks in the 'toucan' class. In this resolution, we also provide the visualization of the generated images for the medical imaging DermaMNIST dataset in \Cref{fig:derma}. Furthermore, we provide ImageNet-1K at a resolution of 128$\times$128 in \Cref{fig:ImageNet128-PT1}, \Cref{fig:ImageNet128-PT2}, \Cref{fig:ImageNet128-PT3}, \Cref{fig:ImageNet128-PT4},\Cref{fig:ImageNet128-PT5}, and all the associated subsets: ImageNette (\Cref{fig:ImageNette}), ImageWoof (\Cref{fig:ImageWoof}), ImageSquawk (\Cref{fig:ImageSquack}), ImageFruit( \Cref{fig:ImageFruit}), ImageMeow (\Cref{fig:ImageMeow}), ImageNet-A (\Cref{fig:ImageNet-A}), ImageNet-B (\Cref{fig:ImageNet-B}), ImageNet-C (\Cref{fig:ImageNet-C}), ImageNet-D (\Cref{fig:ImageNet-D}), ImageNet-E (\Cref{fig:ImageNet-E}).

\clearpage  

%
%

\end{document}